\documentclass{article}
\usepackage{arxiv}

\usepackage{microtype}
\usepackage{graphicx}
\usepackage{subcaption}
\usepackage{booktabs}

\usepackage{hyperref}

\usepackage{amsmath}
\usepackage{amssymb}
\usepackage{mathtools}
\usepackage{amsthm}
\usepackage{algorithm}
\usepackage{algorithmic}

\usepackage[capitalize,noabbrev]{cleveref}

\title{Trust Regions Sell, But Who’s Buying? Overlap Geometry as an Alternative Trust Region for Policy Optimization}
\titlerunning{Trust Regions Sell, But Who's Buying?}
\date{}
\author{
  Gaurish Trivedi\thanks{Joint first authors. All emails use the domain \texttt{@pilani.bits-pilani.ac.in}.} \\
  Birla Institute of Technology and Science, Pilani\\
  Pilani, Rajasthan (333031) \\
  \texttt{f20220728} 
  \And
  Alakh Sharma\footnotemark[1] \\
  Birla Institute of Technology and Science, Pilani\\
  Pilani, Rajasthan (333031) \\
  \texttt{f20240593} 
  \And
  Kartikey Singh Bhandari \\
  Birla Institute of Technology and Science, Pilani\\
  Pilani, Rajasthan (333031) \\
  \texttt{p20241006} 
  \And
  Yash Sinha\\
  Birla Institute of Technology and Science, Pilani\\
  Pilani, Rajasthan (333031) \\
  \texttt{yash.sinha}
  \And
  Pratik Narang \\
  Birla Institute of Technology and Science, Pilani\\
  Pilani, Rajasthan (333031) \\
  \texttt{pratik.narang} 
  \And
  Dhruv Kumar \\
  Birla Institute of Technology and Science, Pilani\\
  Pilani, Rajasthan (333031) \\
  \texttt{dhruv.kumar} 
  \And
  Jagat Sesh Challa \\
  Birla Institute of Technology and Science, Pilani\\
  Pilani, Rajasthan (333031) \\
  \texttt{jagatsesh} 
}

\begin{document}
\maketitle

\newcommand{\piold}{\pi_{\text{old}}}
\newcommand{\thold}{\theta_{\text{old}}}
\newcommand{\E}{\mathbb{E}}
\newcommand{\Eold}{\E_{s\sim d^{\piold},\,a\sim \piold(\cdot\mid s)}}
\newcommand{\Aold}{A^{\piold}}
\newcommand{\logr}{\Delta_{\theta}(s,a)}
\newcommand{\rth}{r_\theta(s,a)}
\newcommand{\qth}{q_\theta(s,a)}
\newcommand{\pio}{\pi_0}
\newcommand{\Ao}{A_0}
\newcommand{\Eo}{\mathbb{E}_0}

\vspace{2em}
\begin{abstract}
    Standard trust-region methods constrain policy updates via Kullback--Leibler (KL) divergence.
However, KL controls only an average divergence and does not directly prevent rare, large
likelihood-ratio excursions that destabilize training---precisely the failure mode that motivates
heuristics such as PPO's clipping. We propose overlap geometry as an alternative trust region,
constraining distributional overlap via the Bhattacharyya coefficient (closely related to the
Hellinger/R\'enyi-$\tfrac12$ geometry). This objective penalizes separation in the ratio tails, yielding
tighter control over likelihood-ratio excursions without relying on total variation bounds that can
be loose in tail regimes. We derive Bhattacharyya-TRPO (BTRPO) and Bhattacharyya-PPO (BPPO),
enforcing overlap constraints via square-root ratio updates: BPPO clips the square-root ratio
$q=\sqrt{r}$, and BTRPO applies a quadratic Hellinger/Bhattacharyya penalty.
Empirically, overlap-based updates improve robustness and aggregate performance as measured by
RLiable under matched training budgets, suggesting overlap constraints as a practical, principled
alternative to KL for stable policy optimization.

\end{abstract}

\section{Introduction}
Trust-region methods have come to play a pivotal role in modern policy-gradient reinforcement learning , primarily because they mitigate destructive policy updates by regularizing the distance between new and behavior policies, typically employing Kullback-Leibler (KL) divergence as a proxy for local geometry \citep{schulman2015trpo}. Nevertheless, a growing body of literature recognizes that many practical RL pipelines, including widely used proximal methods, still suffer from significant instability \citep{engstrom2020implementationmatters}. This instability is increasingly attributed not to the \emph{average} divergence, but to \emph{rare} likelihood-ratio excursions that dominate gradient variance and lead to \textit{single-batch failures} \citep{schulman2017ppo,garg2021ppoheavytails}. Such phenomena have become particularly salient as reinforcement learning is applied to higher-dimensional action spaces and longer horizons, where policy updates frequently induce heavy-tailed importance ratios, elevating trust-region stability to a primary concern \citep{garg2021ppoheavytails,liu2018horizoncurse}. Our work addresses this challenge within the domain of stable policy optimization, aiming to develop update rules whose geometry is explicitly aligned with the statistical failure modes of likelihood-ratio learning.

\begin{figure*}
  \centering

  \begin{subfigure}[t]{0.32\textwidth}
    \centering
    \includegraphics[width=\linewidth]{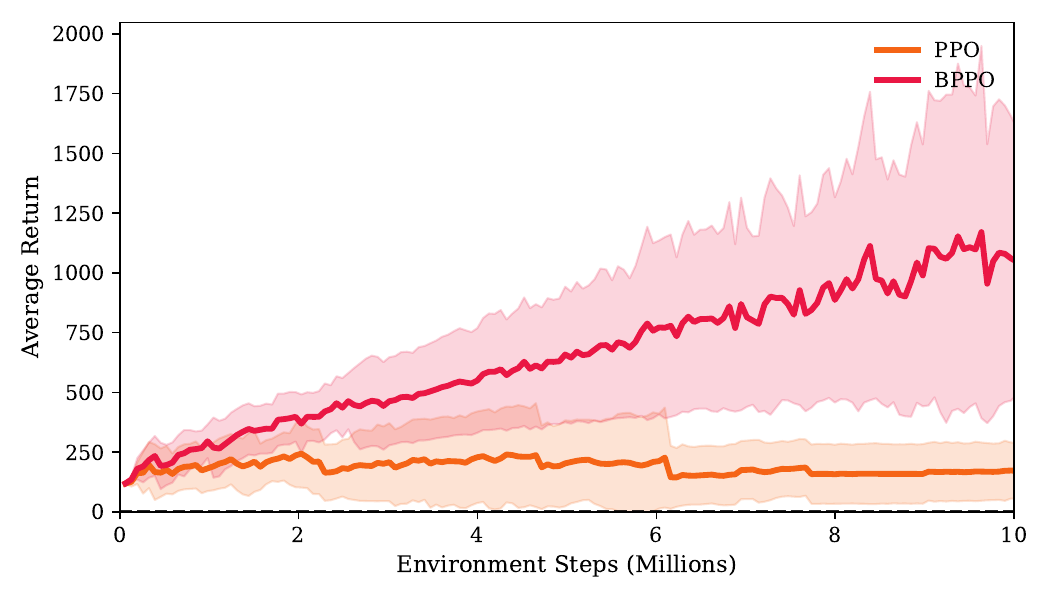}
    \caption{Returns.}
    \label{fig:intro:humanoid:returns}
  \end{subfigure}\hfill
  \begin{subfigure}[t]{0.32\textwidth}
    \centering
    \includegraphics[width=\linewidth]{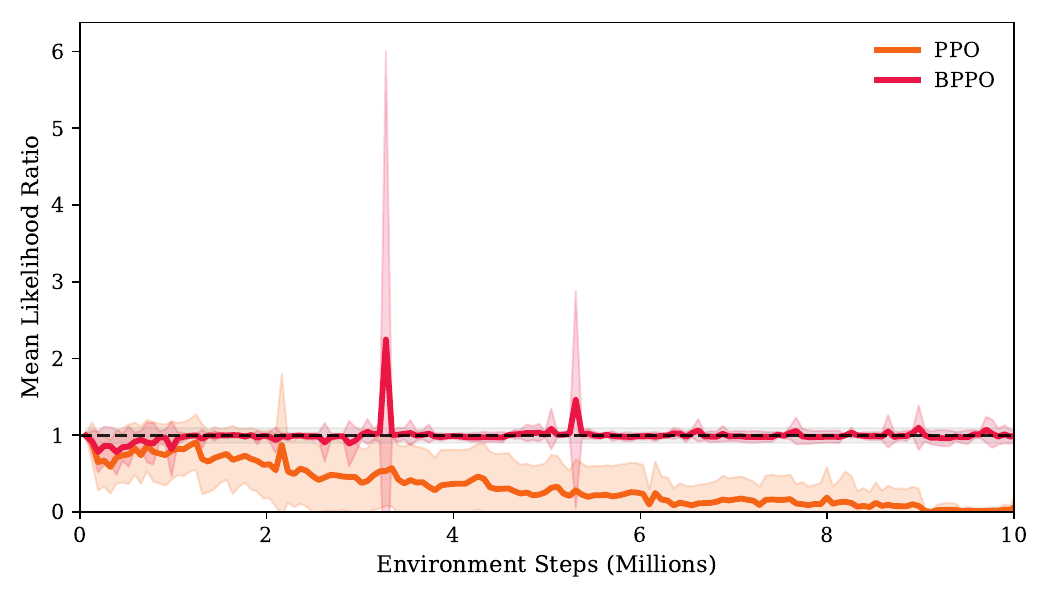}
    \caption{Mean likelihood ratio.}
    \label{fig:intro:humanoid:mean_ratio}
  \end{subfigure}\hfill
  \begin{subfigure}[t]{0.32\textwidth}
    \centering
    \includegraphics[width=\linewidth]{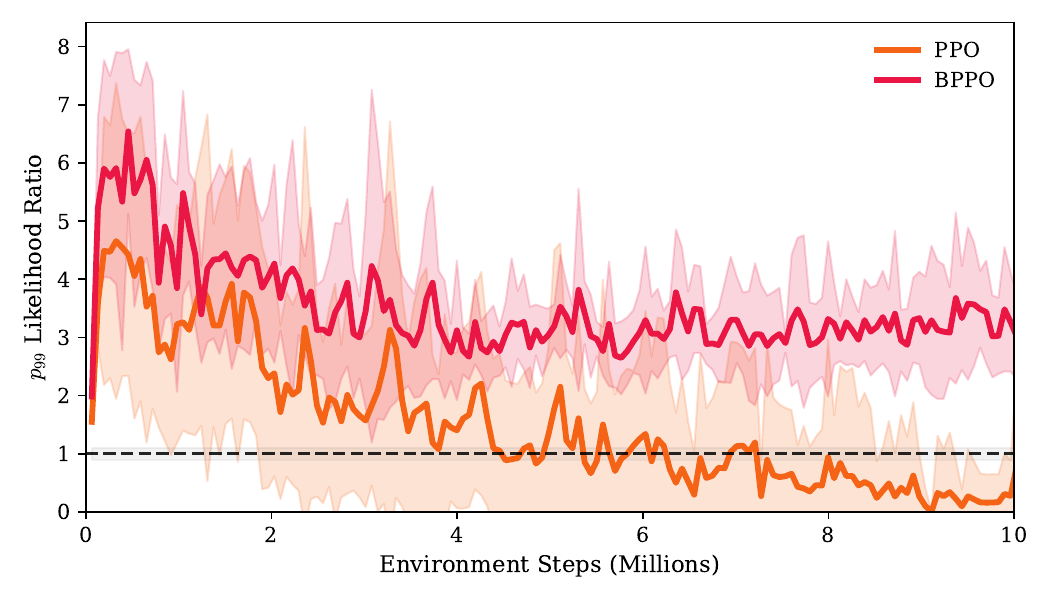}
    \caption{99th percentile likelihood ratio (upper tail).}
    \label{fig:intro:humanoid:p99_ratio}
  \end{subfigure}

  \caption{\textbf{A motivating signature of update shrinkage on \texttt{Humanoid} (9 seeds (0--8), 10M steps).}
  (\textbf{a}) BPPO continues to improve throughout training while PPO plateaus early.
  (\textbf{b--c}) This gap coincides with markedly different likelihood-ratio dynamics:
  PPO’s likelihood-ratio statistics contract over training (both the mean and the upper tail move toward $1$ or below),
  consistent with increasingly constrained effective updates, whereas BPPO maintains near-unity mean ratio and a stable,
  non-trivial upper tail. Shaded regions indicate variability across seeds (as plotted).}
  \label{fig:intro:humanoid:teaser}
\end{figure*}

It is now well established that standard KL-based trust regions constrain only the average divergence and do not directly mitigate the rare but extreme likelihood-ratio spikes (large $r(s,a)=\pi_\theta(a|s)/\pi_{\text{old}}(a|s)$) that frequently destabilize training. This precise failure mode has motivated the development of heuristic safeguards such as PPO clipping \citep{schulman2017ppo,garg2021ppoheavytails}. Our formulation \textit{highlights a critical mismatch}: ratio tails can remain heavy even when the KL divergence is small, yielding unstable gradients and indicating that local Fisher geometry alone does not reliably control tail behavior in importance-weighted regimes. Consequently, existing practices often rely on ad hoc or non-smooth mechanisms, such as ratio clipping, which lack a direct relationship to a principled trust-region framework.

Previous studies have established TRPO as a formal method for policy improvement using KL trust-region constraints motivated by the Fisher information metric \citep{schulman2015trpo}, while PPO introduced a clipped surrogate objective to approximate these constraints through first-order methods \citep{schulman2017ppo}. Subsequent off-policy and distributed variants have incorporated truncation and correction mechanisms to manage importance-weight variance \citep{munos2016retrace,espeholt2018impala,wang2016acer}. While these methods have improved empirical stability, they largely \emph{react} to ratio pathologies via clipping or truncation rather than \emph{encoding tail control} directly within the trust-region geometry. Recent evidence indicates that PPO gradients remain heavy-tailed, suggesting that clipping functions as an implicit robustification rather than a principled constraint \citep{garg2021ppoheavytails}. Thus, a significant research gap exists: there is a need for a trust-region geometry that maintains local Fisher structure for small updates while providing explicit, principled control over likelihood-ratio tails during larger updates \citep{kakade2001naturalpg}.

To address these limitations, this paper proposes a new methodology for stable policy optimization, termed \textit{Tail-Adaptive Geometries}. This novel trust-region construction realigns update geometry to explicitly manage heavy-tailed likelihood ratios through \emph{overlap constraints}. Specifically, we replace KL divergence with \textit{overlap maximization} using the Bhattacharyya coefficient (equivalently Hellinger / R\'enyi-$\tfrac12$ geometry \citep{li2016renyivi}), which penalizes separation in the ratio tails and provides a principled route to tail control \citep{bhattacharyya1943measure,amari2016information}. In contrast to KL, Bhattacharyya/Hellinger overlap defines a normalized notion of policy proximity that emphasizes distributional overlap rather than average log-density shift, preserving local Fisher structure while providing a principled basis for regularizing larger updates. We leverage this in \S\ref{sec:d_derivation}.

Our methodological approach utilizes the classical square-root density parameterization, $\psi_\theta(a|s)=\sqrt{\pi_\theta(a|s)}$, whereby policies are represented on the unit sphere in $L^2$ and distributional overlap is expressed as an inner product \citep{amari2016information}. We define the Bhattacharyya coefficient as $\rho_s(\theta,\theta')=\langle\psi_\theta,\psi_{\theta'}\rangle$, with the squared Hellinger distance $H_s^2=1-\rho_s$, and demonstrate that it \textit{preserves the local Fisher structure} while maintaining its interpretability as overlap \citep{amari2016information}. Subsequently, we derive a first-order surrogate objective based on the \textit{square-root ratio} $q_\theta(s,a)=\sqrt{r_\theta(s,a)}$, leading to the Hellinger-weighted surrogate $L_{\text{Hell}}(\theta)=\mathbb{E}_{\text{old}}[2(q_\theta-1)A_{\text{old}}]$. Finally, we instantiate two practical algorithms: \textit{BPPO}, which employs clipping on $q_\theta$ within a PPO-style objective, and \textit{BTRPO}, which incorporates a Bhattacharyya/Hellinger regularizer proportional to $1-\text{BC}$.

The proposed overlap-based trust regions were evaluated under matched training budgets, comparing BPPO and BTRPO against standard KL-based baselines across a variety of continuous-control benchmarks. To ensure robust reporting, we utilized aggregate performance metrics and uncertainty-aware evaluation techniques, following established best practices such as RLiable to summarize results across multiple tasks and seeds \citep{agarwal2021rliable}. Hyperparameters were swept systematically across all methods, with particular attention paid to high-learning-rate regimes to determine whether overlap constraints provide genuine stability improvements rather than simple performance shifts.

Overall, these results indicate that overlap-based updates significantly improve both robustness and aggregate performance under matched computational budgets, suggesting that overlap constraints provide a \textit{viable and practical alternative to KL divergence} for stable policy optimization. From a theoretical perspective, our analysis clarifies the two operating regimes of overlap geometry: at low learning rates, Bhattacharyya/Hellinger regularization remains locally equivalent to KL, thus preserving favorable Fisher geometry. At larger steps, overlap regularizes updates through direct control of the square-root ratio $q_\theta=\sqrt{r_\theta}$: BPPO deterministically bounds $r_\theta$ via clipping of $q_\theta$, and BTRPO smooths ratio excursions by penalizing deviations of $q_\theta$ from unity. Furthermore, the parameterization $q_\theta=e^{\Delta/2}$ ensures that ratio spikes are damped smoothly and consistently with the underlying geometry.

\paragraph{Research questions.}
The central focus of this study is to address the following questions:
\begin{enumerate}
  \item \textit{Tail-adaptive trust-region geometry:} Can overlap-based constraints serve as a principled replacement for KL trust regions while preserving local Fisher behavior? \citep{amari2016information}
  \item \textit{Tail control and update shrinkage:} Does optimizing distributional overlap provide practical control over likelihood-ratio tail events that drive instability in policy-gradient updates, while avoiding the update shrinkage often induced by KL- or clipping-based trust-region heuristics?
  \item \textit{Algorithmic instantiations:} To what extent can practical PPO and TRPO-style algorithms (BPPO/BTRPO) be derived from overlap geometry while retaining simplicity and robustness?
  \item \textit{Empirical robustness under matched budgets:} Do overlap-based trust regions demonstrate superior aggregate performance across benchmarks when subjected to robust reporting (RLiable \citep{agarwal2021rliable})? 
\end{enumerate}


\paragraph{Empirical motivation. \textit{What actually goes wrong in practice?}} Figure~\ref{fig:intro:humanoid:teaser} illustrates a key failure mode of standard trust regions: update shrinkage. On \texttt{Humanoid} \citep{todorov2012mujoco}, PPO’s learning stagnates as its update statistics, both mean and tail likelihood ratios contract over time. In contrast, BPPO maintains a stable, non-trivial upper tail that enables sustained improvement (Fig.~\ref{fig:intro:humanoid:teaser}b--c). This raises our central question: Can we design a trust-region geometry that controls the tail events driving instability without suppressing the updates necessary for continued progress?

\section{Derivation}
\label{sec:d_derivation}
\subsection{Square-Root Policy Geometry}
\label{subsec:sqrt_geometry}

In standard reinforcement learning approaches, policies are typically treated as points on a probability simplex, often necessitating complex constraints to manage updates safely. We propose a geometric shift by parameterizing stochastic policies via their \emph{square-root density}. Formally, we define this representation as:
\begin{equation}
\label{eq:psi_def_rewrite}
\psi_\theta(a\mid s)\triangleq\sqrt{\pi_\theta(a\mid s)}.
\end{equation}
This transformation fundamentally alters the underlying manifold structure. For each state $s$, the normalization condition of the probability density implies that $\int \psi_\theta(a\mid s)^2\,da = 1$. Consequently, the function $\psi_\theta(\cdot\mid s)$ resides on the unit sphere in the $L^2$ Hilbert space. This mapping is a classical construction in information geometry \citep{amari2016information,kakade2001naturalpg} which provides significant analytical advantages: it maps probability densities to unit vectors, ensuring that overlaps between distributions become inner products and that Euclidean distances on this sphere correspond naturally to overlap-based divergences.

\subsection{Hellinger--Bhattacharyya Geometry and Local Fisher Structure}
\label{subsec:hellinger_fisher_rewrite}

\paragraph{Bhattacharyya coefficient and Hellinger distance.}
To quantify the proximity between policies in this transformed space, we utilize a natural similarity measure known as the \emph{Bhattacharyya coefficient} (BC) \citep{bhattacharyya1943measure}. Unlike the Kullback-Leibler divergence, which is asymmetric and unbounded, the BC provides a normalized measure of overlap. For a fixed state $s$, we define the coefficient as:
\begin{equation}
\label{eq:bc_state_rewrite}
\rho_s(\theta,\theta')
\triangleq
\int \sqrt{\pi_\theta(a\mid s)\,\pi_{\theta'}(a\mid s)}\,da
=
\langle \psi_\theta(\cdot\mid s),\psi_{\theta'}(\cdot\mid s)\rangle.
\end{equation}
This overlap measure is intrinsically linked to the squared Hellinger distance. Under the convention used in this work, the relationship is given by:
\begin{equation}
\label{eq:hell_bc_rewrite}
H_s^2(\theta,\theta')
\triangleq
\frac{1}{2}\int\big(\sqrt{\pi_\theta(a\mid s)}-\sqrt{\pi_{\theta'}(a\mid s)}\big)^2 da
=
1-\rho_s(\theta,\theta'). 
\end{equation}
which reveals that the geometry defined by the BC and Hellinger distance reduces effectively to standard Euclidean geometry restricted to the unit sphere.

\paragraph{Local equivalence to the Fisher metric.}
A critical property of this geometry is its behavior in the local vicinity of the current policy parameters. Let $F_s(\theta)$ denote the Fisher information matrix of $\pi_\theta(\cdot\mid s)$. If we consider a small perturbation $\delta$ around a reference parameter (typically the behavior policy $\theta_{\mathrm{old}}$), the Bhattacharyya coefficient admits the following second-order Taylor expansion:
\begin{equation}
\label{eq:bc_fisher_rewrite}
\rho_s(\theta,\theta+\delta)
=
1-\frac{1}{8}\,\delta^\top F_s(\theta)\,\delta
+o(\|\delta\|^2).
\end{equation}
For comparison, the standard KL divergence admits a similar quadratic form involving the Fisher matrix:
\begin{equation}
\label{eq:kl_fisher_rewrite}
D_{\mathrm{KL}}\!\Big(\pi_{\theta}(\cdot\mid s)\,\Big\|\,\pi_{\theta+\delta}(\cdot\mid s)\Big)
=
\frac{1}{2}\,\delta^\top F_s(\theta)\,\delta
+o(\|\delta\|^2).
\end{equation}
(Both KL directions share the same local Fisher quadratic; in experiments we use the old-to-new form $D_{\mathrm{KL}}(\pi_{\mathrm{old}}\|\pi_\theta)$.)

This similarity confirms that constraining the BC (or equivalently, the Hellinger distance) induces a trust region that is locally equivalent to the Fisher trust region used in Natural Policy Gradient methods, differing only by a constant factor. \textit{Takeaway (RQ1): BC/Hellinger trust regions preserve the same local Fisher quadratic structure as KL, while remaining bounded and symmetric}. However, distinct from KL-based methods, our approach allows us to estimate these overlap terms directly from samples collected under $\pi_{\mathrm{old}}$, avoiding the need for secondary approximations. 
\paragraph{State-averaged overlap.}
In the context of reinforcement learning, we must consider the policy's performance across the entire state space. We aggregate the local overlap measures using the discounted occupancy measure of the behavior policy $\pi_{\mathrm{old}}$, denoted as $d^{\pi_{\mathrm{old}}}(s)$. We define the expected overlap as:
\begin{equation}
\label{eq:bc_avg_rewrite}
\bar{\rho}(\theta,\theta_{\mathrm{old}})
\triangleq
\mathbb{E}_{s\sim d^{\pi_{\mathrm{old}}}}
\big[\rho_s(\theta,\theta_{\mathrm{old}})\big].
\end{equation}

\subsection{A First-Order Surrogate with Square-Root Ratios}
\label{subsec:hellinger_surrogate_rewrite}

We now derive a surrogate objective that leverages this square-root geometry. Let $\pi_{\mathrm{old}}=\pi_{\theta_{\mathrm{old}}}$ represent the behavior policy and $A_{\mathrm{old}}(s,a)$ its associated advantage function. The standard TRPO surrogate objective is defined as \citep{schulman2015trpo,sutton2000policygradient}:
\begin{equation}
\label{eq:trpo_surrogate_rewrite}
L(\theta)
\triangleq
\mathbb{E}_{\mathrm{old}}\!\Big[r_\theta(s,a)\,A_{\mathrm{old}}(s,a)\Big].
\end{equation}
Where \(r_\theta(s,a)\triangleq\frac{\pi_\theta(a\mid s)}{\pi_{\mathrm{old}}(a\mid s)}\) and
assuming that $\pi_\theta(\cdot\mid s)$ is absolutely continuous with respect to $\pi_{\mathrm{old}}(\cdot\mid s)$ to ensure the ratios are well-defined, we introduce the \emph{square-root ratio}, a key component of our method, let \(\Delta_\theta(s,a)\triangleq
\log\pi_\theta(a\mid s)-\log\pi_{\mathrm{old}}(a\mid s)\):
\begin{equation}
\label{eq:sqrt_ratio_def_rewrite}
q_\theta(s,a)\triangleq \sqrt{r_\theta(s,a)}
=
\exp\!\Big(\tfrac{1}{2}\,\Delta_\theta(s,a)\Big).
\end{equation}
We instantiate a bounded likelihood-ratio transform by operating on the square-root ratio
$q_\theta=\exp(\Delta_\theta/2)$ and applying either saturation through clipping
$q_\theta\in[1-\epsilon,1+\epsilon]$ (BPPO) or smooth saturation via the quadratic
Hellinger penalty $(1-q_\theta)^2$ (BTRPO).
Since the standard importance ratio is the square of our new variable, i.e., $r_\theta=q_\theta^2$, we can perform a first-order Taylor expansion around the point of no change, $q_\theta=1$. This yields:
\begin{equation}
\label{eq:taylor_q_rewrite}
q_\theta^2
=
1+2(q_\theta-1) + O\!\big((q_\theta-1)^2\big) 
\end{equation}
Equivalenty, \(r_\theta \approx 1+2(q_\theta-1)\).
Substituting this expansion back into the original TRPO surrogate expression in \eqref{eq:trpo_surrogate_rewrite} gives us:
\begin{equation}
\label{eq:surrogate_expand_rewrite}
L(\theta)
=
\mathbb{E}_{\mathrm{old}}
\Big[\big(1+2(q_\theta-1)+O((q_\theta-1)^2)\big)\,A_{\mathrm{old}}\Big]
=
\underbrace{\mathbb{E}_{\mathrm{old}}[A_{\mathrm{old}}]}_{=\,0}
+
\mathbb{E}_{\mathrm{old}}\!\Big[2(q_\theta-1)\,A_{\mathrm{old}}\Big]
+
\mathbb{E}_{\mathrm{old}}\!\Big[O((q_\theta-1)^2)\,A_{\mathrm{old}}\Big].
\end{equation}
By utilizing the fact that $\mathbb{E}_{a\sim\pi_{\mathrm{old}}(\cdot\mid s)}[A_{\mathrm{old}}(s,a)]=0$ for every state $s$, the constant term vanishes. If we then drop the second-order error terms, we arrive at what we term the \emph{Hellinger-weighted first-order surrogate}:
\begin{equation}
\label{eq:hell_surrogate_rewrite}
L_{\mathrm{Hell}}(\theta)
\triangleq
\mathbb{E}_{\mathrm{old}}
\Big[
2\big(q_\theta(s,a)-1\big)\,A_{\mathrm{old}}(s,a)
\Big].
\end{equation}
It is important to note that since the residual is $r_\theta - (1+2(q_\theta-1)) = (q_\theta-1)^2$, the truncation error of this approximation is directly controlled by the deviations of $q_\theta$ from $1$. We explicitly manage this deviation via regularization in \S\ref{subsec:hellinger_trust_region_rewrite}. \textit{Interpretation (RQ2): Optimizing in $q=\sqrt{r}$ makes tail excursions harder to dominate the update because extreme ratios correspond to quadratically large deviations in $q$, which are directly saturated (BPPO) or penalized (BTRPO). Unlike clipping $r$, saturating $q$ better preserves a near-nominal mean update scale while still damping rare spikes, matching the ratio-statistic trends we measure empirically.}

\paragraph{Implemented form.}
Because the advantage function has zero mean under the behavior policy, i.e., $\mathbb{E}_{\mathrm{old}}[A_{\mathrm{old}}]=0$, the surrogate can be simplified to an equivalent form for implementation:
\begin{equation}
\label{eq:drop_constant_rewrite}
\mathbb{E}_{\mathrm{old}}
\Big[
2\big(q_\theta-1\big)\,A_{\mathrm{old}}
\Big]
=
\mathbb{E}_{\mathrm{old}}
\Big[
2\,q_\theta\,A_{\mathrm{old}}
\Big].
\end{equation}
In practical applications, we additionally center and normalize the advantages within each minibatch. This ensures that the constant term vanishes exactly even in finite samples, significantly reducing variance. \textit{Takeaway (RQ3): BPPO and BTRPO are \emph{direct algorithmic instantiations} of BC/Hellinger trust-region geometry within the standard on-policy optimization template (PPO/TRPO).
In practice, this requires caching $\log \pi_{\theta_{\mathrm{old}}}(a|s)$ and optimizing in terms of $q=\exp(\Delta/2)$ rather than $r$.} We summarize the resulting on-policy optimization loop for BPPO/BTRPO in Algorithm~\ref{alg:bppo_btrpo_compact}.

\paragraph{BTRPO: BC/Hellinger regularization.}
Alternatively, to rigorously control the update size within the Hellinger geometry, we can penalize deviations of the square-root ratio from unity. We define the regularization term as:
\begin{equation}
\label{eq:bc_reg_rewrite}
R_{\mathrm{BC}}(\theta)
\triangleq
\mathbb{E}_{\mathrm{old}}\!\left[(1-q_\theta(s,a))^2\right].
\end{equation}
Expanding this square reveals a direct connection to our geometric metrics:
\begin{equation}
\label{eq:bc_connection_rewrite}
\mathbb{E}_{\mathrm{old}}[(1-q_\theta)^2]
=
\mathbb{E}_{\mathrm{old}}[1-2q_\theta+r_\theta]
=
2\big(1-\mathbb{E}_{\mathrm{old}}[q_\theta]\big),
\end{equation}
where we have used the identity $\mathbb{E}_{\mathrm{old}}[r_\theta]=1$, which holds exactly in expectation. Furthermore, the expectation of the square-root ratio is the Bhattacharyya coefficient itself:
\begin{equation}
\label{eq:q_is_bc_rewrite}
\mathbb{E}_{\mathrm{old}}[q_\theta]
=
\mathbb{E}_{s\sim d^{\pi_{\mathrm{old}}}}
\left[
\int \sqrt{\pi_\theta(a\mid s)\pi_{\mathrm{old}}(a\mid s)}\,da 
\right]
=
\bar{\rho}(\theta,\theta_{\mathrm{old}}),
\end{equation}
Thus, minimizing $R_{\mathrm{BC}}$ is proportional to minimizing $1-\mathrm{BC}$ (or equivalently, the Hellinger distance), providing a geometrically grounded trust-region penalty. Our final penalized objective for maximization is therefore:
\begin{equation}
\label{eq:btrpo_penalized_rewrite}
\max_\theta
\left\{
\mathbb{E}_{\mathrm{old}}\!\left[
    2\,q_\theta(s,a)\,A_{\mathrm{old}}(s,a)
\right]
- \beta\,
\mathbb{E}_{\mathrm{old}}\!\left[
    (1 - q_\theta(s,a))^2
\right]
\right\}
\end{equation}

where $\beta$ is a fixed coefficient in our current implementation.

\paragraph{BPPO: clipped square-root surrogate.}
Building on the success of PPO's clipping mechanism, we introduce a variant called BPPO. This method applies clipping directly to the square-root ratio, rather than the standard ratio:
\begin{equation}
\label{eq:bppo_obj_rewrite}
\triangleq\;
\mathbb{E}_{\mathrm{old}}\Big[
2 \min \big(
q_\theta(s,a)\,A_{\mathrm{old}}(s,a),
\mathrm{clip}\big(
    q_\theta(s,a),
    1-\epsilon,
    1+\epsilon
\big)\,
A_{\mathrm{old}}(s,a)
\big)
\Big]
\end{equation}

Constraining $q_\theta\in[1-\epsilon,1+\epsilon]$ is equivalent to constraining the standard likelihood ratio to $r_\theta=q_\theta^2\in[(1-\epsilon)^2,(1+\epsilon)^2]$. This results in a smooth, overlap-motivated damping of the ratio updates while retaining the robustness characteristic of PPO.

\paragraph{Why square-root ratios reduce tail sensitivity?}
A major theoretical advantage of this approach lies in its handling of importance weights. Under the log-ratio $\Delta_\theta$, standard importance weights scale exponentially as $r_\theta=e^{\Delta_\theta}$. In contrast, our square-root weights scale as $q_\theta=e^{\Delta_\theta/2}$ \citep{liu2018horizoncurse,metelli2018pois,garg2021ppoheavytails}. This implies that large positive-ratio events, which often destabilize training, and grow exponentially more slowly in our framework, providing a smooth, inherent mechanism for heavy-tail suppression that serves as a robust alternative to hard clipping.

\begin{algorithm}[t]
\caption{Square-Root Policy Optimization (BPPO / BTRPO)}
\label{alg:bppo_btrpo_compact}
\small
\begin{algorithmic}[1]
\STATE Initialize policy $\pi_{\theta_0}$ and value $V_{\phi_0}$.
\FOR{$i = 0,1,2,\dots$}
    \STATE Collect rollouts $\mathcal{D}$ with $\pi_{\theta_i}$; store $\log\pi_{\theta_i}(a\mid s)$.
    \STATE Compute advantages $A_{\theta_i}(s,a)$ and targets $\hat{R}$ (e.g., GAE).
    \STATE Compute $\Delta=\log\pi_{\theta}(a\mid s)-\log\pi_{\theta_i}(a\mid s)$ and $q=\exp(\Delta/2)$.
    \IF{\textbf{BPPO}}
        \STATE Optimize $\max_\theta\;\mathbb{E}[\,2\min(qA,\;\text{clip}(q,1\!-\!\epsilon,1\!+\!\epsilon)A)\,]$.
    \ELSE 
        \STATE \textbf{(BTRPO)}
        \STATE Optimize $\max_\theta\;\mathbb{E}[2qA]-\beta\,\mathbb{E}[(1-q)^2]$.
    \ENDIF
    \STATE Update $\phi$ by minimizing $\mathbb{E}[(V_\phi(s)-\hat{R})^2]$ (optionally add entropy).
\ENDFOR
\end{algorithmic}
\end{algorithm}

\subsection{Hellinger-Regularized Trust Region}
\label{subsec:hellinger_trust_region_rewrite}

\paragraph{BC as an on-policy expectation.}
One of the practical strengths of the Bhattacharyya coefficient is that it admits a simple estimator under the behavior policy $\pi_{\mathrm{old}}$, avoiding the need for complex importance sampling or off-policy corrections:
\begin{equation}
\label{eq:bc_as_expectation_rewrite}
\rho_s(\theta,\theta_{\mathrm{old}})
=
\int \sqrt{\pi_\theta(a\mid s)\,\pi_{\mathrm{old}}(a\mid s)}\,da
=
\mathbb{E}_{a\sim \pi_{\mathrm{old}}(\cdot\mid s)}\!\big[q_\theta(s,a)\big].
\end{equation}
When averaging over the state space, this yields a tractable objective:
\begin{equation}
\label{eq:bc_avg_est_rewrite}
\bar{\rho}(\theta,\theta_{\mathrm{old}})
=
\mathbb{E}_{\mathrm{old}}\!\big[q_\theta(s,a)\big].
\end{equation}

\paragraph{Constrained and penalty forms.}
Finally, we can enforce a geometric trust region by explicitly constraining the policy overlap. This can be formulated as a constrained optimization problem:
\begin{equation}
\label{eq:bc_constraint_rewrite}
\max_\theta\:
\left\{L_{\mathrm{Hell}}(\theta)\;\Big|\;\bar{\rho}(\theta,\theta_{\mathrm{old}})\ge 1-\varepsilon \right\},
\end{equation}
or equivalently, by utilizing a penalty (dual) form which is often easier to optimize via first-order methods \citep{schulman2015trpo,achiam2017cpo}:
\begin{equation}
\label{eq:bc_penalty_rewrite}
\max_\theta\:\left\{
L_{\mathrm{Hell}}(\theta)
-\lambda\big(1-\bar{\rho}(\theta,\theta_{\mathrm{old}})\big)\right\},
\end{equation}
where $\lambda>0$ is a hyperparameter that controls the step size within the Hellinger geometry.



\section{Evaluations and experiments}
\label{sec:e_experiments}

\paragraph{Evaluation philosophy.}
In this section, we describe the evaluation framework used to assess performance across continuous-control benchmarks, placing a particular emphasis on robust reporting and matched budgets. To increase the reliability of measures, results are generally aggregated over multiple random seeds and summarized using RLiable-style statistics, specifically IQM and Optimality Gap, to reduce sensitivity to outlier runs \citep{agarwal2021rliable}. Learning curves, representing the mean across seeds with shaded variability, are reported to highlight both asymptotic performance and training stability. 

\subsection{MuJoCo environments}
\label{subsec:mujoco}

We benchmark BPPO and BTRPO on standard MuJoCo continuous-control tasks (\texttt{-v5} suite) \citep{todorov2012mujoco} and compare these against two widely-used on-policy baselines: PPO \citep{schulman2017ppo} and TRPO \citep{schulman2015trpo}.

\begin{figure*}[t]
    \centering
    \includegraphics[width=\textwidth]{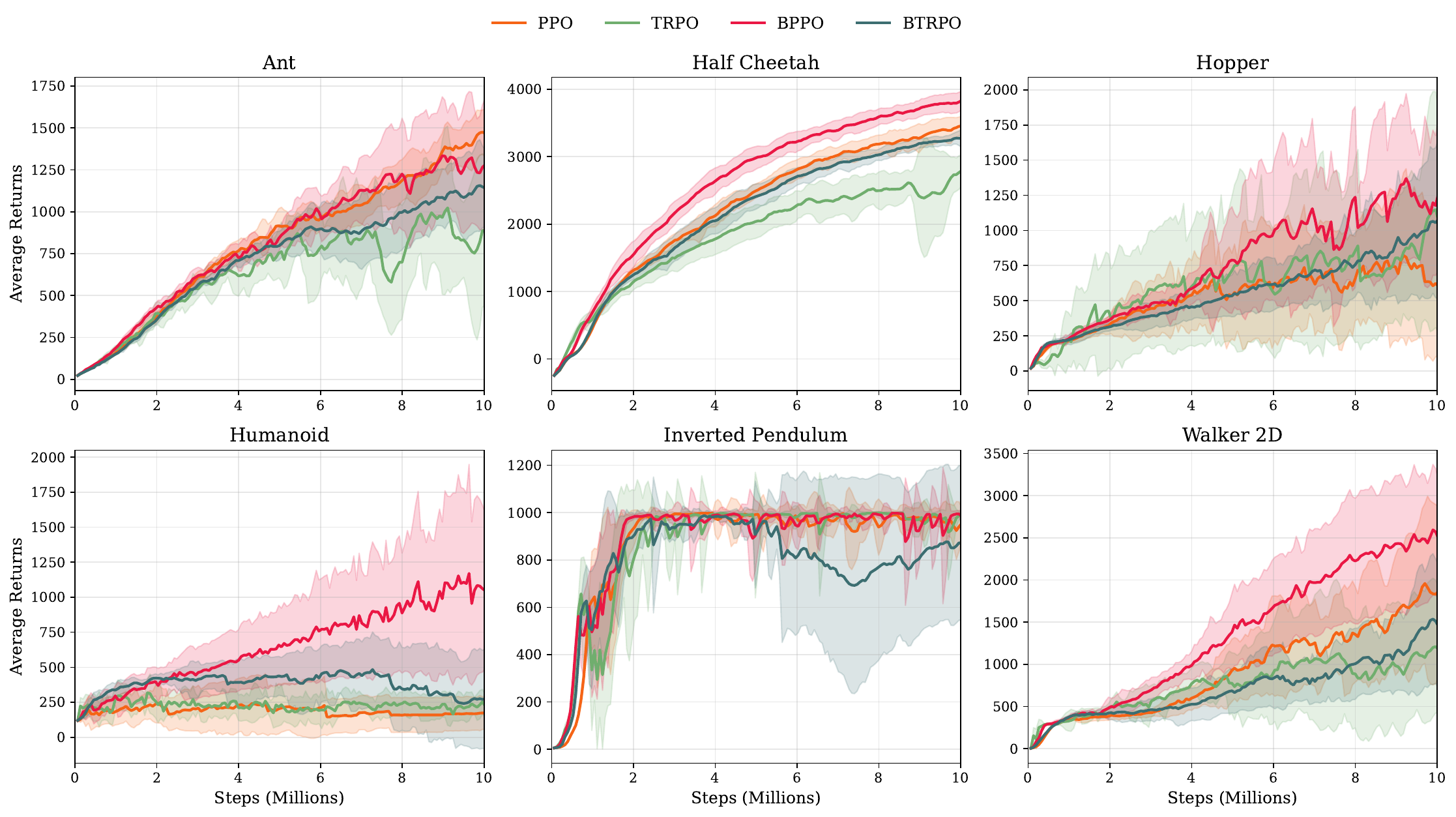}
    \caption{\textbf{MuJoCo learning curves (9 seeds (0--8)).} 
Mean episodic return over training on six MuJoCo tasks; shaded regions indicate $\pm 1$ standard deviation across seeds. 
BPPO generally achieves higher returns and more sustained late-stage improvement on harder locomotion tasks.}
\label{fig:mujoco_all_envs_returns}
\end{figure*}

\textbf{Experimental Protocol.}
Following the protocol established by \citep{agarwal2021rliable}, for each algorithm--task pair we train using \textbf{9 independent random seeds} and report the \textbf{Interquartile Mean (IQM)} of final performance, where higher values indicate better performance. Learning curves in Figure~\ref{fig:mujoco_all_envs_returns} show the average return trajectory over training, with shaded regions indicating across-seed variability.

\textbf{Environments.}
We evaluate performance on six MuJoCo tasks spanning different dynamics and difficulty: \texttt{Ant-v5}, \texttt{HalfCheetah-v5}, \texttt{Hopper-v5}, \texttt{Humanoid-v5}, \texttt{InvertedPendulum-v5}, and \texttt{Walker2d-v5}.

\begin{table}[t]
    \centering
    \caption{\textbf{MuJoCo IQM scores (9 seeds (0--8); higher is better).} 
    We report the Interquartile Mean (IQM) aggregated across 9 seeds for each 
    environment. Best per environment is bolded. The last row reports the mean 
    IQM across the six tasks (an unweighted average over environments).}
    \label{tab:mujoco_iqm}
    \begin{small}
    \begin{sc}
    \begin{tabular}{lcccc}
        \toprule
        Environment & PPO & TRPO & BTRPO & \textbf{BPPO} \\
        \midrule
        Ant-v5               & \textbf{1470.8} & 816.7  & 1169.1 & 1293.9 \\
        HalfCheetah-v5       & 3451.3 & 2819.2 & 3299.2 & \textbf{3856.3} \\
        Hopper-v5            & 617.5  & 913.2  & 928.9  & \textbf{1321.6} \\
        Humanoid-v5          & 147.5  & 204.9  & 110.2  & \textbf{1099.2} \\
        InvertedPendulum-v5  & 977.4  & 992.8  & 979.6  & \textbf{997.0} \\
        Walker2d-v5          & 1765.4 & 1076.7 & 1397.0 & \textbf{2697.3} \\
        \midrule
        \textbf{Mean IQM}    & 1405.0 & 1137.2 & 1314.0 & \textbf{1877.5} \\
        \bottomrule
    \end{tabular}
    \end{sc}
    \end{small}
\end{table}

\textbf{Results and Analysis.}
The results presented in Figure~\ref{fig:mujoco_all_envs_returns} and Table~\ref{tab:mujoco_iqm} indicate that the proposed square-root geometry improves both \emph{learning dynamics} and \emph{robust aggregate performance} on MuJoCo.

\begin{enumerate}
    \item \textbf{BPPO exhibits stronger learning dynamics and achieves the best overall aggregate.} 
    BPPO attains the highest IQM on 5/6 environments and the highest \textbf{Mean IQM} across tasks (Table~\ref{tab:mujoco_iqm}). Importantly, the evidence presented in the learning curves (Figure~\ref{fig:mujoco_all_envs_returns}) suggests that these gains are not limited to final scores: BPPO typically exhibits \emph{faster early learning} and \emph{more sustained late-stage improvement} on the harder locomotion tasks. The performance gap is most pronounced on \texttt{Walker2d-v5} and \texttt{HalfCheetah-v5}, where BPPO maintains a clear advantage as training progresses, and on \texttt{Humanoid-v5}, where BPPO continues improving while PPO/TRPO plateau substantially. These dynamics align with the central contribution of our method: using the square-root ratio $q_\theta=\sqrt{r_\theta}$ damps extreme importance weights while preserving a smooth learning signal.

    \item \textbf{Mechanistic evidence for tail control on \texttt{Humanoid-v5}.} 
    Figure~\ref{fig:intro:humanoid:returns} provides a mechanistic view of the observed performance gap. As training progresses, PPO’s learning curve flattens while BPPO continues to improve; this divergence coincides with a contraction of PPO’s likelihood-ratio statistics, whereas BPPO maintains near-unity mean ratios while retaining a non-trivial upper tail. These findings support the interpretation that overlap-based updates can \emph{preserve effective update magnitude} in high-dimensional control while still \emph{damping} extreme ratio excursions. \textit{This provides direct evidence for RQ2, while Table \ref{tab:mujoco_iqm} establish RQ4 under matched budgets via robust aggregates (Mean IQM).}

    \item \textbf{BTRPO is competitive and exhibits a conservative stability profile.} 
    BTRPO improves over TRPO on several tasks but is more sensitive on \texttt{Humanoid-v5} under a fixed penalty coefficient $\beta$ (Table~\ref{tab:mujoco_iqm}). The learning curves are consistent with a more \emph{conservative} update profile, often showing steady progress without the sharp jumps characteristic of clipping-based updates. Taken together, these results suggest that the Hellinger penalty $\beta(1-q_\theta)^2$ offers a principled trust-region alternative, and indicate that adaptive penalty tuning may further improve robustness.
\end{enumerate}

\subsection{DM Control environments}
\label{subsec:dmcontrol}

We evaluate the proposed geometric surrogates on the DeepMind Control Suite (\texttt{dm\_control}) \citep{Tassa2020dmcontrolSA}, a benchmark characterized by continuous control with complex contact dynamics. We compare BPPO and BTRPO against PPO and TRPO \citep{schulman2017ppo,schulman2015trpo}.

\begin{figure*}[t]
    \centering
    \includegraphics[width=\textwidth]{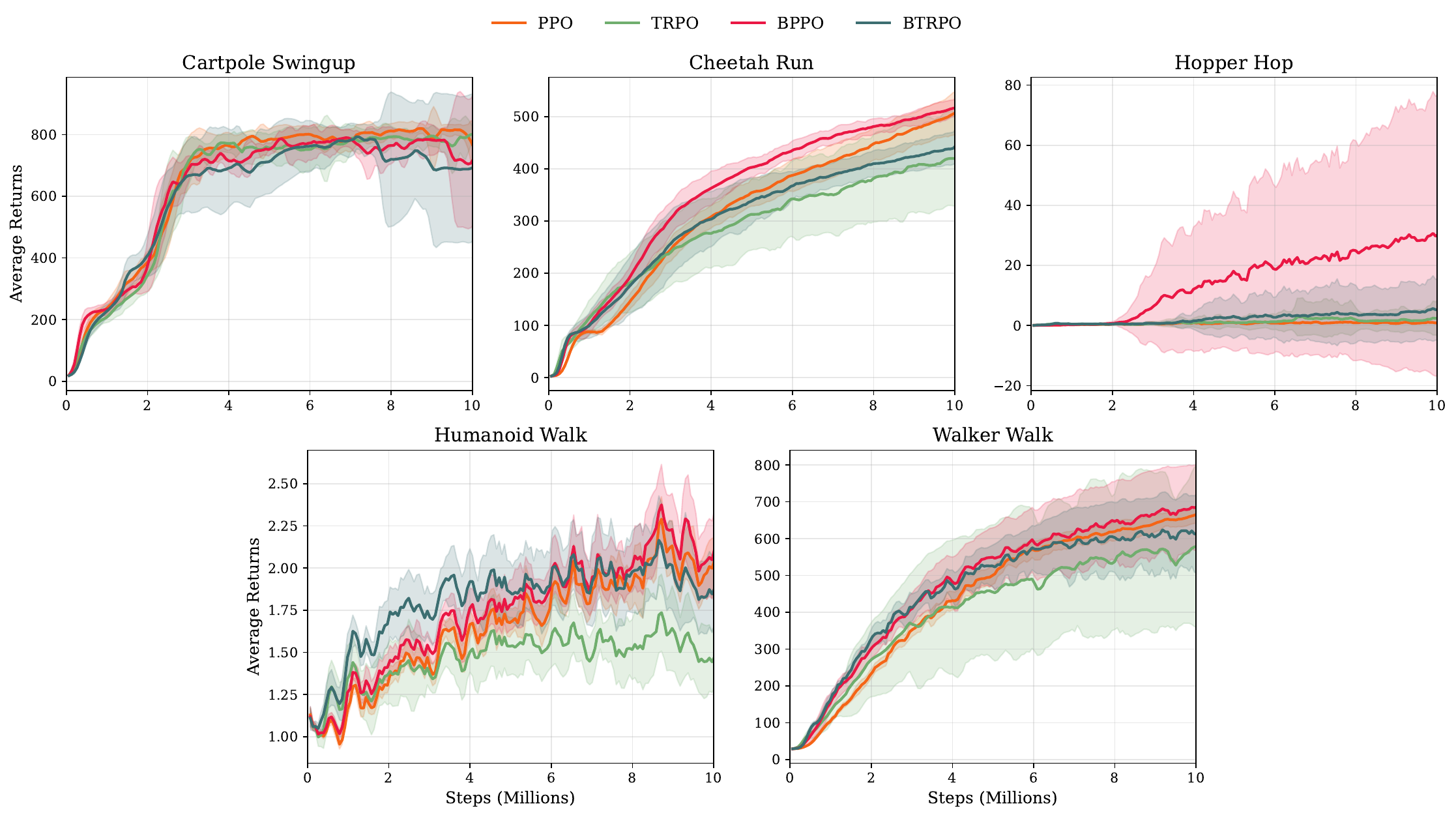}
    \caption{\textbf{DM Control learning curves (9 seeds (0--8)).} 
Average episodic return vs.\ environment steps on five tasks; shaded regions indicate $\pm 1$ standard deviation across seeds. 
BPPO shows stronger learning progress on \texttt{cheetah\_run}, \texttt{hopper\_hop}, and \texttt{walker\_walk}, while PPO remains competitive on \texttt{cartpole\_swingup}.}
    
    \label{fig:dm_all_envs_returns}
\end{figure*}

\textbf{Experimental Protocol.}
We follow the approach of \citep{agarwal2021rliable} and train \textbf{9 seeds} per algorithm--task pair. We report IQM (Table~\ref{tab:dm_control_scores}) and provide learning curves (Figure~\ref{fig:dm_all_envs_returns}) to expose both asymptotic performance and stability.

\textbf{Environments.}
We benchmark five tasks: \texttt{cartpole\_swingup}, \texttt{cheetah\_run}, \texttt{hopper\_hop}, \texttt{walker\_walk}, and \texttt{humanoid\_walk}.

\begin{table}[t]
    \centering
    \caption{\textbf{DM Control (9 seeds (0--8)) IQM scores (higher is better).} 
    Interquartile Mean (IQM) aggregated across 9 seeds for each task. 
    Best per task is bolded. The last row reports the mean IQM across tasks.}
    \label{tab:dm_control_scores}
    \begin{small}
    \begin{sc}
    \begin{tabular}{lcccc}
        \toprule
        Environment & PPO & BTRPO & TRPO & \textbf{BPPO} \\
        \midrule
         cartpole\_swingup & \textbf{806.21} & 799.39 & 804.93 & 778.72 \\
         cheetah\_run      & 513.10 & 441.62 & 429.83 & \textbf{513.22} \\
         hopper\_hop       & 0.91   & 0.69   & 0.57   & \textbf{13.23} \\
         humanoid\_walk    & 1.98   & 1.88   & 1.47   & \textbf{2.10} \\
         walker\_walk      & 656.94 & 626.75 & 550.04 & \textbf{673.23} \\
        \midrule
        \textbf{Mean IQM}  & 395.83 & 374.07 & 357.37 & \textbf{396.10} \\
        \bottomrule
    \end{tabular}
    \end{sc}
    \end{small}
\end{table}

\textbf{Results and Analysis.}
The evidence in Figure~\ref{fig:dm_all_envs_returns} and Table~\ref{tab:dm_control_scores} shows that the proposed square-root geometry yields consistent improvements in both \emph{learning dynamics} and \emph{robust aggregate performance} on DM Control.

\begin{enumerate}
    \item \textbf{BPPO enhances learning dynamics and achieves the best overall aggregate.} 
    BPPO attains the highest IQM on the majority of tasks and achieves the best overall \textbf{Mean IQM} (Table~\ref{tab:dm_control_scores}). Beyond final scores, the learning curves (Figure~\ref{fig:dm_all_envs_returns}) indicate that BPPO typically exhibits \emph{faster early improvement} and \emph{more stable late-stage training} than PPO/TRPO. The most pronounced gain is on \texttt{hopper\_hop}, where BPPO’s curve separates decisively from the baselines, consistent with the method’s motivation of damping extreme importance weights.

    \item \textbf{BTRPO provides a principled trust-region alternative, with a conservative update profile.} 
    BTRPO exceeds TRPO on all five tasks in Table~\ref{tab:dm_control_scores}. The curves in Figure~\ref{fig:dm_all_envs_returns} suggest a more \emph{conservative} update profile, often tracking PPO early but improving more gradually with fewer sharp spikes. Although BTRPO does not dominate PPO in all configurations, its stability-oriented behavior supports the role of Hellinger regularization as a practical alternative to heuristic clipping.
\end{enumerate}

\subsection{Procgen environments}
\label{subsec:procgen}

We evaluate BPPO and BTRPO on the Procgen benchmark \citep{Cobbe2019LeveragingPG}, which measures robustness and generalization in procedurally-generated environments. In contrast to continuous control, Procgen tasks exhibit higher stochasticity, providing a complementary stress test for ratio-based updates.

\textbf{Experimental Protocol.}
We use standard easy-mode configurations and train \textbf{3 seeds} per algorithm for 25M steps, treating Procgen as a complementary breadth benchmark. We report mean episode return across seeds $\pm$ std across seeds.

\textbf{Environments.}
We benchmark four Procgen tasks: \texttt{coinrun}, \texttt{heist}, \texttt{jumper}, and \texttt{ninja}.

\begin{table}[t]
    \centering
    \caption{\textbf{Procgen (easy) aggregate performance (higher is better).} 
    Entries report mean episode return across 3 seeds (0--3)$\pm$ std across seeds 
    (25 eval episodes/seed). Best per environment is bolded (ties bolded).}
    \label{tab:procgen_easy_scores}
    \begin{small}
    \begin{sc}
    \begin{tabular}{lcccc}
        \toprule
        Environment & PPO & TRPO-KL & BTRPO-BC & \textbf{BPPO} \\
        \midrule
        CoinRun & 8.13 $\pm$ 1.29 & \textbf{9.20 $\pm$ 0.40} & 4.87 $\pm$ 4.69 & 8.53 $\pm$ 0.23 \\
        Heist   & \textbf{1.87 $\pm$ 1.40} & 0.67 $\pm$ 0.46 & 0.67 $\pm$ 0.46 & \textbf{1.87 $\pm$ 1.40} \\
        Jumper  & 4.67 $\pm$ 0.23 & 4.80 $\pm$ 1.06 & 5.73 $\pm$ 0.31 & \textbf{5.87 $\pm$ 1.80} \\
        Ninja   & 6.27 $\pm$ 1.40 & \textbf{7.07 $\pm$ 1.01} & 4.27 $\pm$ 0.50 & 6.27 $\pm$ 2.20 \\
        \bottomrule
    \end{tabular}
    \end{sc}
    \end{small}
\end{table}

\textbf{Results and Analysis.}
Table~\ref{tab:procgen_easy_scores} reports aggregate returns on Procgen. Unlike the consistent gains in continuous control, results are more mixed; this variation may be attributed to procedural stochasticity and task-specific reward structure.

\begin{enumerate}
    \item \textbf{BPPO remains competitive under procedural stochasticity.} 
    BPPO improves over PPO on \texttt{coinrun} and achieves the best performance on \texttt{jumper}. This supports the central intuition of the square-root surrogate: moderating large-ratio updates while preserving a smooth learning signal in high-variance settings.

    \item \textbf{BTRPO-BC shows higher sensitivity to fixed penalty strength on Procgen.} 
    In the current configuration, BTRPO-BC exhibits higher variance and lower mean performance on certain tasks (Table~\ref{tab:procgen_easy_scores}). This suggests that a non-adaptive penalty coefficient $\beta$ may require more careful tuning in Procgen, where reward sparsity can amplify the exploration--stability trade-off.
\end{enumerate}

\label{sec:exps}


\section{Limitations}
Our approach has several limitations. First, the square-root surrogate is motivated by a first-order expansion around $q=1$; when updates drift far from this regime (e.g., due to large learning rates or long inner-loop optimization), approximation error can grow and empirical performance may degrade. Second, stability depends on ratio-control hyperparameters (e.g., the BPPO clipping range $\epsilon$ and the BTRPO penalty coefficient $\beta$). In particular, using a fixed, non-adaptive $\beta$ can be sensitive across environments and training budgets, motivating adaptive or dual-tuned penalties as future work. Third, the method assumes sufficient support overlap between $\pi_{\text{old}}$ and $\pi_{\theta}$ so that likelihood ratios and square-root ratios are well-defined; if policy support changes sharply, ratio-based surrogates (including ours) can become unstable or overly conservative. Finally, while we evaluate across standard continuous-control benchmarks and Procgen, Procgen results are mixed, suggesting that the benefits of overlap geometry may interact with exploration dynamics and reward sparsity in task-dependent ways, and warrant broader evaluation on additional domains and larger-scale sweeps.

\label{sec:i_limitations}

\section{Conclusion}
We propose an overlap-based trust-region perspective for
policy optimization by reparameterizing updates in square-
root space and controlling tail-sensitive deviations via
Bhattacharyya/Hellinger geometry. This yields a simple
first-order surrogate with two practical algorithms: BPPO,
which clips the square-root ratio, and BTRPO, which ap-
plies a quadratic Hellinger-style penalty. Across standard
continuous-control benchmarks under matched budgets,
these overlap-constrained updates improve stability and
overall performance, with BPPO achieving the strongest
aggregate results while maintaining robust update behavior.
In addition, likelihood-ratio diagnostics show that overlap
constraints reduce extreme tail events without collapsing
the mean update magnitude. A promising direction is
adaptive tuning of the Hellinger penalty to further
strengthen BTRPO’s robustness without sacrificing its
conservative stability profile.

\label{sec:h_conclusion}

\section*{Impact Statement}

This research introduces technical improvements to on-policy reinforcement learning intended to enhance training stability and sample efficiency by utilizing overlap-based policy updates, specifically through Hellinger/Bhattacharyya geometries. These methodological advances are designed to mitigate common training failures, such as policy collapse and extreme likelihood-ratio excursions, which may ultimately reduce the energy consumption and computational overhead required for agent training in simulations. As this work focuses on fundamental optimization and learning principles, it does not provide capabilities tailored to any specific application domain. The broader downstream implications of this study are consistent with those of the wider reinforcement learning field, including potential risks associated with the misuse or hazardous deployment of trained policies in real-world environments. Such concerns should be addressed through the implementation of domain-specific safety protocols, rigorous evaluation, and continuous monitoring prior to any practical deployment.
\label{sec:f_impact}

\printbibliography[title={References}]

@inproceedings{schulman2015trpo,
  title={Trust region policy optimization},
  author={Schulman, John and Levine, Sergey and Abbeel, Pieter and Jordan, Michael and Moritz, Philipp},
  booktitle={International conference on machine learning},
  pages={1889--1897},
  year={2015},
  organization={PMLR}
}

@article{schulman2017ppo,
  title={Proximal policy optimization algorithms},
  author={Schulman, John and Wolski, Filip and Dhariwal, Prafulla and Radford, Alec and Klimov, Oleg},
  journal={arXiv preprint arXiv:1707.06347},
  year={2017}
}

@article{wang2016acer,
  title={Sample efficient actor-critic with experience replay},
  author={Wang, Ziyu and Bapst, Victor and Heess, Nicolas and Mnih, Volodymyr and Munos, Remi and Kavukcuoglu, Koray and De Freitas, Nando},
  journal={arXiv preprint arXiv:1611.01224},
  year={2016}
}

@article{belousov2018fdivergence,
  title={f-Divergence constrained policy improvement},
  author={Belousov, Boris and Peters, Jan},
  journal={arXiv preprint arXiv:1801.00056},
  year={2017}
}

@inproceedings{zhang2018wasserstein,
  title={Policy optimization as wasserstein gradient flows},
  author={Zhang, Ruiyi and Chen, Changyou and Li, Chunyuan and Carin, Lawrence},
  booktitle={International Conference on machine learning},
  pages={5737--5746},
  year={2018},
  organization={PMLR}
}

@article{greensmith2004variance,
  title={Variance reduction techniques for gradient estimates in reinforcement learning},
  author={Greensmith, Evan and Bartlett, Peter L and Baxter, Jonathan},
  journal={Journal of Machine Learning Research},
  volume={5},
  number={Nov},
  pages={1471--1530},
  year={2004}
}

@article{munos2016retrace,
  title={Safe and efficient off-policy reinforcement learning},
  author={Munos, R{\'e}mi and Stepleton, Tom and Harutyunyan, Anna and Bellemare, Marc},
  journal={Advances in neural information processing systems},
  volume={29},
  year={2016}
}

@inproceedings{espeholt2018impala,
  title={Impala: Scalable distributed deep-rl with importance weighted actor-learner architectures},
  author={Espeholt, Lasse and Soyer, Hubert and Munos, Remi and Simonyan, Karen and Mnih, Vlad and Ward, Tom and Doron, Yotam and Firoiu, Vlad and Harley, Tim and Dunning, Iain and others},
  booktitle={International conference on machine learning},
  pages={1407--1416},
  year={2018},
  organization={PMLR}
}

@article{peng2019awr,
  title={Advantage-weighted regression: Simple and scalable off-policy reinforcement learning},
  author={Peng, Xue Bin and Kumar, Aviral and Zhang, Grace and Levine, Sergey},
  journal={arXiv preprint arXiv:1910.00177},
  year={2019}
}

@article{wu2019brac,
  title={Behavior regularized offline reinforcement learning},
  author={Wu, Yifan and Tucker, George and Nachum, Ofir},
  journal={arXiv preprint arXiv:1911.11361},
  year={2019}
}

@inproceedings{garg2021ppoheavytails,
  title={On proximal policy optimization’s heavy-tailed gradients},
  author={Garg, Saurabh and Zhanson, Joshua and Parisotto, Emilio and Prasad, Adarsh and Kolter, Zico and Lipton, Zachary and Balakrishnan, Sivaraman and Salakhutdinov, Ruslan and Ravikumar, Pradeep},
  booktitle={International Conference on Machine Learning},
  pages={3610--3619},
  year={2021},
  organization={PMLR}
}

@article{sutton2000policygradient,
  title={Policy gradient methods for reinforcement learning with function approximation},
  author={Sutton, Richard S and McAllester, David and Singh, Satinder and Mansour, Yishay},
  journal={Advances in neural information processing systems},
  volume={12},
  year={1999}
}

@inproceedings{
nachum2018trustpcl,
title={Trust-{PCL}: An Off-Policy Trust Region Method for Continuous Control},
author={Ofir Nachum and Mohammad Norouzi and Kelvin Xu and Dale Schuurmans},
booktitle={International Conference on Learning Representations},
year={2018},
url={https://openreview.net/forum?id=HyrCWeWCb},
}

@article{becker2025troll,
  title={TROLL: Trust Regions improve Reinforcement Learning for Large Language Models},
  author={Becker, Philipp and Freymuth, Niklas and Thilges, Serge and Otto, Fabian and Neumann, Gerhard},
  journal={arXiv preprint arXiv:2510.03817},
  year={2025}
}

@article{wang2018tailadaptive,
  title={Variational inference with tail-adaptive f-divergence},
  author={Wang, Dilin and Liu, Hao and Liu, Qiang},
  journal={Advances in Neural Information Processing Systems},
  volume={31},
  year={2018}
}

@article{nair2020awac,
  title={Awac: Accelerating online reinforcement learning with offline datasets},
  author={Nair, Ashvin and Gupta, Abhishek and Dalal, Murtaza and Levine, Sergey},
  journal={arXiv preprint arXiv:2006.09359},
  year={2020}
}

@article{williams1992reinforce,
  title={Simple statistical gradient-following algorithms for connectionist reinforcement learning},
  author={Williams, Ronald J},
  journal={Machine learning},
  volume={8},
  number={3},
  pages={229--256},
  year={1992},
  publisher={Springer}
}

@article{bhattacharyya1943measure,
  title={On a measure of divergence between two statistical populations defined by their probability distribution},
  author={Bhattacharyya, Anil},
  journal={Bulletin of the Calcutta Mathematical Society},
  volume={35},
  pages={99--110},
  year={1943}
}

@book{amari2016information,
  title={Information geometry and its applications},
  author={Amari, Shun-ichi},
  volume={194},
  year={2016},
  publisher={Springer}
}

@article{agarwal2021rliable,
  title={Deep reinforcement learning at the edge of the statistical precipice},
  author={Agarwal, Rishabh and Schwarzer, Max and Castro, Pablo Samuel and Courville, Aaron C and Bellemare, Marc},
  journal={Advances in neural information processing systems},
  volume={34},
  pages={29304--29320},
  year={2021}
}

@article{kakade2001naturalpg,
  title={A natural policy gradient},
  author={Kakade, Sham M},
  journal={Advances in neural information processing systems},
  volume={14},
  year={2001}
}

@article{li2016renyivi,
  title={R{\'e}nyi divergence variational inference},
  author={Li, Yingzhen and Turner, Richard E},
  journal={Advances in neural information processing systems},
  volume={29},
  year={2016}
}

@inproceedings{engstrom2020implementationmatters,
  title={Implementation matters in deep rl: A case study on ppo and trpo},
  author={Engstrom, Logan and Ilyas, Andrew and Santurkar, Shibani and Tsipras, Dimitris and Janoos, Firdaus and Rudolph, Larry and Madry, Aleksander},
  booktitle={International conference on learning representations},
  year={2019}
}

@inproceedings{achiam2017cpo,
  title={Constrained policy optimization},
  author={Achiam, Joshua and Held, David and Tamar, Aviv and Abbeel, Pieter},
  booktitle={International conference on machine learning},
  pages={22--31},
  year={2017},
  organization={PMLR}
}

@article{liu2018horizoncurse,
  title={Breaking the curse of horizon: Infinite-horizon off-policy estimation},
  author={Liu, Qiang and Li, Lihong and Tang, Ziyang and Zhou, Dengyong},
  journal={Advances in neural information processing systems},
  volume={31},
  year={2018}
}

@article{metelli2018pois,
  title={Policy optimization via importance sampling},
  author={Metelli, Alberto Maria and Papini, Matteo and Faccio, Francesco and Restelli, Marcello},
  journal={Advances in Neural Information Processing Systems},
  volume={31},
  year={2018}
}

@inproceedings{todorov2012mujoco,
  title={MuJoCo: A physics engine for model-based control},
  author={Todorov, Emanuel and Erez, Tom and Tassa, Yuval},
  booktitle={2012 IEEE/RSJ International Conference on Intelligent Robots and Systems},
  pages={5026--5033},
  year={2012},
  organization={IEEE},
  doi={10.1109/IROS.2012.6386109}
}

@article{lin2002divergence,
  title={Divergence measures based on the Shannon entropy},
  author={Lin, Jianhua},
  journal={IEEE Transactions on Information theory},
  volume={37},
  number={1},
  pages={145--151},
  year={2002},
  publisher={IEEE}
}

@Inbook{Pearson1992,
author="Pearson, Karl",
title="On the Criterion that a Given System of Deviations from the Probable in the Case of a Correlated System of Variables is Such that it Can be Reasonably Supposed to have Arisen from Random Sampling",
bookTitle="Breakthroughs in Statistics: Methodology and Distribution",
year="1992",
publisher="Springer New York",
address="New York, NY",
pages="11--28",
isbn="978-1-4612-4380-9",
doi="10.1007/978-1-4612-4380-9_2",
url="https://doi.org/10.1007/978-1-4612-4380-9_2"
}

@book{jeffreys1998theory,
  title={The theory of probability},
  author={Jeffreys, Harold},
  year={1998},
  publisher={OuP Oxford}
}

@article{cuturi2013sinkhorn,
  title={Sinkhorn distances: Lightspeed computation of optimal transport},
  author={Cuturi, Marco},
  journal={Advances in neural information processing systems},
  volume={26},
  year={2013}
}

@article{sinkhorn1967concerning,
  title={Concerning nonnegative matrices and doubly stochastic matrices},
  author={Sinkhorn, Richard and Knopp, Paul},
  journal={Pacific Journal of Mathematics},
  volume={21},
  number={2},
  pages={343--348},
  year={1967},
  publisher={Mathematical Sciences Publishers}
}

@inproceedings{feydy2019interpolating,
  title={Interpolating between optimal transport and mmd using sinkhorn divergences},
  author={Feydy, Jean and S{\'e}journ{\'e}, Thibault and Vialard, Fran{\c{c}}ois-Xavier and Amari, Shun-ichi and Trouv{\'e}, Alain and Peyr{\'e}, Gabriel},
  booktitle={The 22nd international conference on artificial intelligence and statistics},
  pages={2681--2690},
  year={2019},
  organization={PMLR}
}

@article{pfau2025wasserstein,
  title={Wasserstein Policy Optimization},
  author={Pfau, David and Davies, Ian and Borsa, Diana and Araujo, Joao GM and Tracey, Brendan and van Hasselt, Hado},
  journal={arXiv preprint arXiv:2505.00663},
  year={2025}
}

@inproceedings{Cobbe2019LeveragingPG,
  title={Leveraging Procedural Generation to Benchmark Reinforcement Learning},
  author={Karl Cobbe and Christopher Hesse and Jacob Hilton and John Schulman},
  booktitle={International Conference on Machine Learning},
  year={2019},
  url={https://api.semanticscholar.org/CorpusID:208547624}
}

@article{Tassa2020dmcontrolSA,
  title={dm\_control: Software and Tasks for Continuous Control},
  author={Yuval Tassa and Saran Tunyasuvunakool and Alistair Muldal and Yotam Doron and Siqi Liu and Steven Bohez and Josh Merel and Tom Erez and Timothy P. Lillicrap and Nicolas Manfred Otto Heess},
  journal={Softw. Impacts},
  year={2020},
  volume={6},
  pages={100022},
  url={https://api.semanticscholar.org/CorpusID:219980295}
}

\newpage
\appendix

\section{Related Works}

\paragraph{KL-based trust regions and proximal policy optimization.}Trust-region policy optimization has established itself as a canonical framework for stabilizing policy-gradient updates by regulating the deviation between successive policies. The most prominent example of this is TRPO \citep{schulman2015trpo}, which utilizes a KL-divergence constraint. PPO approximates this through a clipped surrogate objective to mitigate likelihood-ratio excursions, a technique that has seen broad adoption due to its empirical success \citep{schulman2017ppo}. Extensive research has also been conducted into practical extensions that preserve the relative-entropy perspective, such as Trust-PCL \citep{nachum2018trustpcl}. More recently, scholars have revisited trust-region enforcement in the context of RLHF and sequence-level optimization, proposing more principled projections while maintaining the KL control signal \citep{becker2025troll}. \textit{In contrast to these KL-centric approaches, the present work replaces the standard geometry with an overlap-centric framework based on the Bhattacharyya coefficient \citep{bhattacharyya1943measure}. Specifically, our update is formulated directly within square-root density geometry (Hellinger/Bhattacharyya), ensuring that tail events typically manifesting as ratio spikes are managed intrinsically through the geometry rather than via external constraints or post-hoc heuristics.}

\paragraph{Divergence-generalized and information-geometric policy optimization.}A substantial body of research has sought to generalize trust regions beyond the KL divergence by exploring broader families of divergences and geometric interpretations. Many commonly used alternatives like Jensen-Shannon \citep{lin2002divergence}, Pearson $\chi^2$ \citep{Pearson1992}, and symmetric/Jeffreys variants \citep{jeffreys1998theory} fall naturally within,
or are closely related to, this information-theoretic divergence landscape. Notably, $f$-divergence constrained policy improvement provides a formal framework for monotone updates, allowing for a trade-off between mode-seeking and mass-covering behaviors \citep{belousov2018fdivergence}. Related efforts have investigated alternative metric structures, such as Wasserstein-gradient-flow perspectives that substitute local KL geometry for transport-based dynamics \citep{zhang2018wasserstein}. From a practical standpoint, entropic optimal transport (OT) yields scalable approximations via Sinkhorn-style algorithms \citep{cuturi2013sinkhorn,sinkhorn1967concerning,feydy2019interpolating}. In adjacent fields of probabilistic inference, \textit{tail-adaptive} constructions have been proposed to mitigate instability arising from heavy-tailed importance weights, highlighting the practical limitations of standard ratio-based objectives when estimation is dominated by tails \citep{wang2018tailadaptive,metelli2018pois}.\textit{Our contribution is not merely to swap divergences, but to operationalize an overlap-maximization principle using an estimator and surrogate objective expressed via the square-root ratio. This results in an inherently tail-damped policy-update mechanism (utilizing $\sqrt{r}$ rather than $r$) and establishes a direct link between the trust region and the Bhattacharyya/Hellinger geometry \citep{bhattacharyya1943measure,amari2016information}.}

\paragraph{Controlling likelihood-ratio tails via importance sampling and robustification.}It is well recognized that likelihood ratios and importance weights represent a central source of variance in policy-gradient and off-policy learning. Classical analyses have emphasized the importance of variance reduction, baselines, and control variates \citep{williams1992reinforce,sutton2000policygradient,greensmith2004variance}. In practice, RL systems frequently rely on truncation or bias-correction strategies to manage weight explosions, such as Retrace-style corrections \citep{munos2016retrace} or actor-critic variants like ACER \citep{wang2016acer}. A further area of research involves distributed architectures that require explicit correction mechanisms, such as V-trace, to manage off-policy drift \citep{espeholt2018impala}. Recent evidence has emerged showing that even on-policy PPO gradients can exhibit heavy-tailed behavior, suggesting that common heuristics like clipping function as implicit robust-statistics interventions \citep{garg2021ppoheavytails}.\textit{The present approach departs from this literature by enforcing tail control at the level of the trust-region geometry itself. Rather than applying post-hoc truncation to ratios or gradients, optimizing overlap via the Bhattacharyya coefficient and square-root ratios ensures that the influence of rare spikes is structurally attenuated while maintaining a principled trust-region interpretation.}

\paragraph{Overlap and behavior-regularized updates under distribution shift.}A separate but closely related line of work addresses the challenges of support mismatch and distribution shift, particularly in offline RL, by regularizing policy updates toward the behavior distribution. Much of the current literature on behavior-regularized approaches focuses on using divergence penalties to prevent extrapolation error \citep{wu2019brac}. Similarly, supervised-style schemes like advantage-weighted regression place a high priority on conservative updates that remain within regions of adequate data coverage \citep{peng2019awr,nair2020awac}. While these methods are frequently motivated by the need for overlap and conservatism, they typically implement these constraints using KL-style regularization or implicit weighting heuristics.\textit{Our work shares this motivation but distinguishes itself by establishing overlap as the primary trust-region quantity through the Bhattacharyya coefficient. This yields an update rule where the constraint targets overlap directly, providing explicit control over ratio spikes rather than indirectly encouraging conservatism through KL penalties.}

\newpage

\section{Bipedal Walk}
\subsection{Baselines}
\begin{figure*}[ht]
    \centering
    \includegraphics[width=\textwidth]{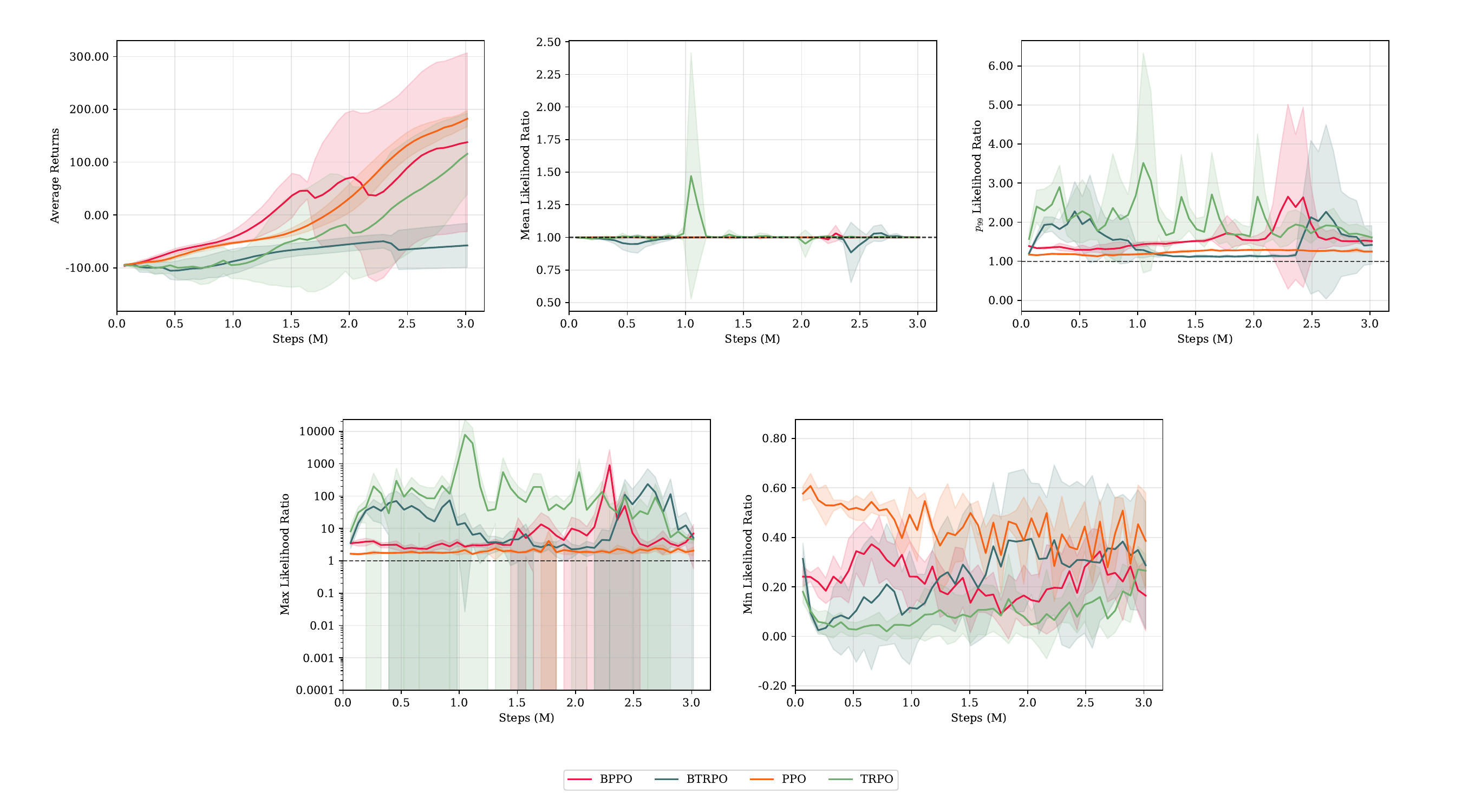}
    \caption{\textsc{BipedalWalker-v3} baseline learning curves (4 seeds).}
\label{fig:bipedal_walk_baseline}
\end{figure*}

\noindent\textbf{Learning dynamics.}
Figure \ref{fig:bipedal_walk_baseline} shows that BPPO and PPO exhibit the strongest improvement trends on \textsc{BipedalWalker-v3}, while TRPO improves more slowly and BTRPO is noticeably more conservative under the same budget.
A useful mechanistic interpretation comes from the likelihood-ratio diagnostics: across all methods, the mean likelihood ratio stays close to $1$ (no systematic drift), but the \emph{upper tail} differs substantially.

\noindent\textbf{Ratio tails and stability.}
TRPO tends to produce larger $p_{99}$ and max-ratio events than the clipping-based methods, reflecting rarer but more extreme importance-weight excursions.
BPPO keeps the upper tail more controlled while still enabling continued improvement, consistent with the square-root surrogate dampening sensitivity to large positive-ratio events.
The min-ratio curve highlights the strongest downweighting events; extremely small minima correspond to samples that become far less likely under the updated policy.

\subsection{Baselines with entropy bonus}
\begin{figure*}[ht]
    \centering
    \includegraphics[width=\textwidth]{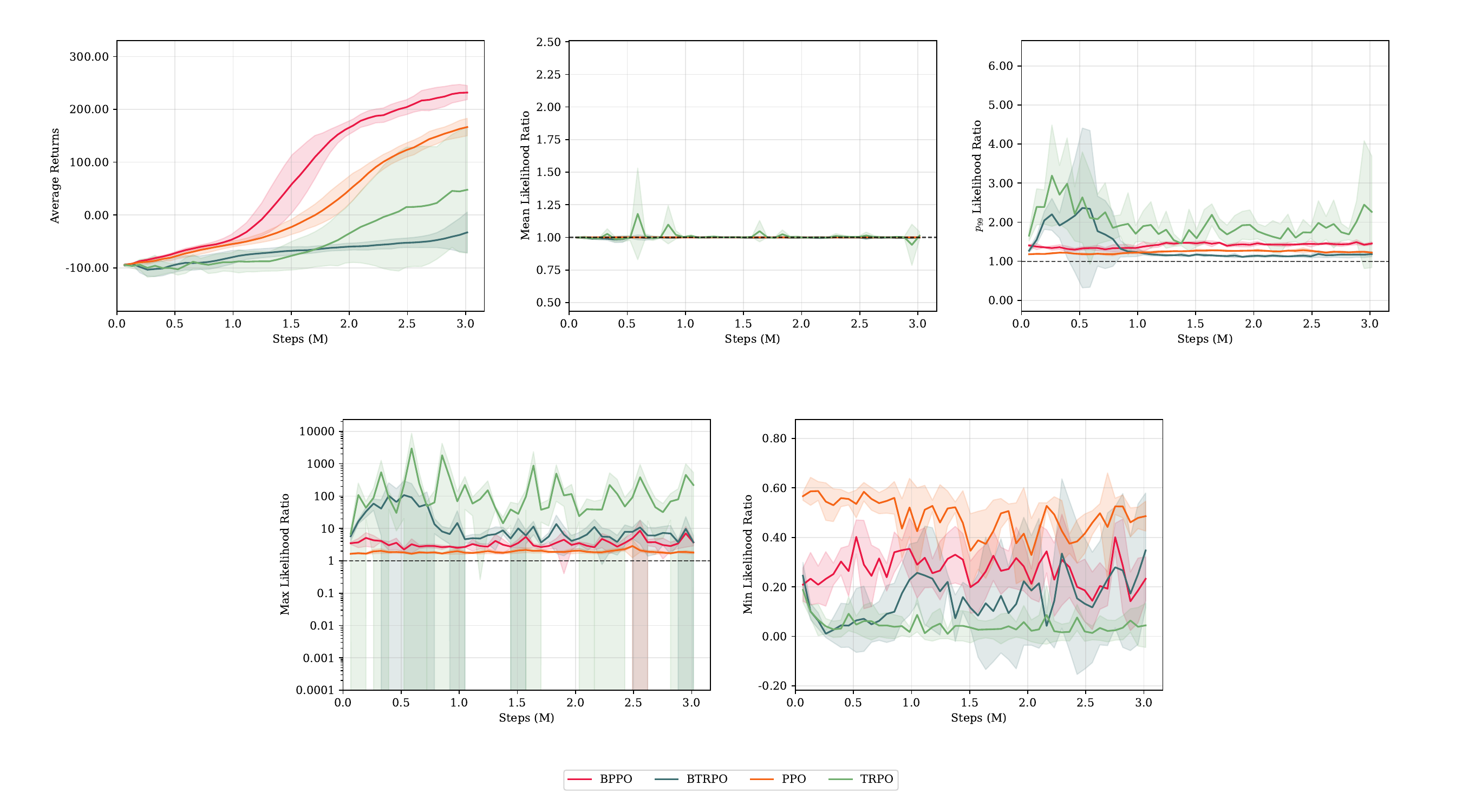}
    \caption{\textsc{BipedalWalker-v3} baseline with entropy bonus learning curves (4 seeds).}
\label{fig:bipedal_walk_baseline_ent}
\end{figure*}

\noindent\textbf{Entropy helps BPPO most here.}
Figure \ref{fig:bipedal_walk_baseline_ent} shows that adding a small entropy bonus improves BPPO’s robustness and learning speed, and typically leads to stronger late-stage performance.
Entropy encourages broader exploration early, which can increase ratio variability, but the BPPO update remains stable because its effective weighting scales with $\sqrt{r}$ rather than $r$.

\noindent\textbf{Tail behavior under entropy.}
With entropy, TRPO’s upper-tail spikes remain pronounced, whereas BPPO and PPO keep the upper tail (especially $p_{99}$) in a more moderate range.
This supports the motivation that controlling tail events matters even when the mean ratio looks benign.

\subsection{Regularized terms}
\begin{figure*}[ht]
    \centering
    \includegraphics[width=\textwidth]{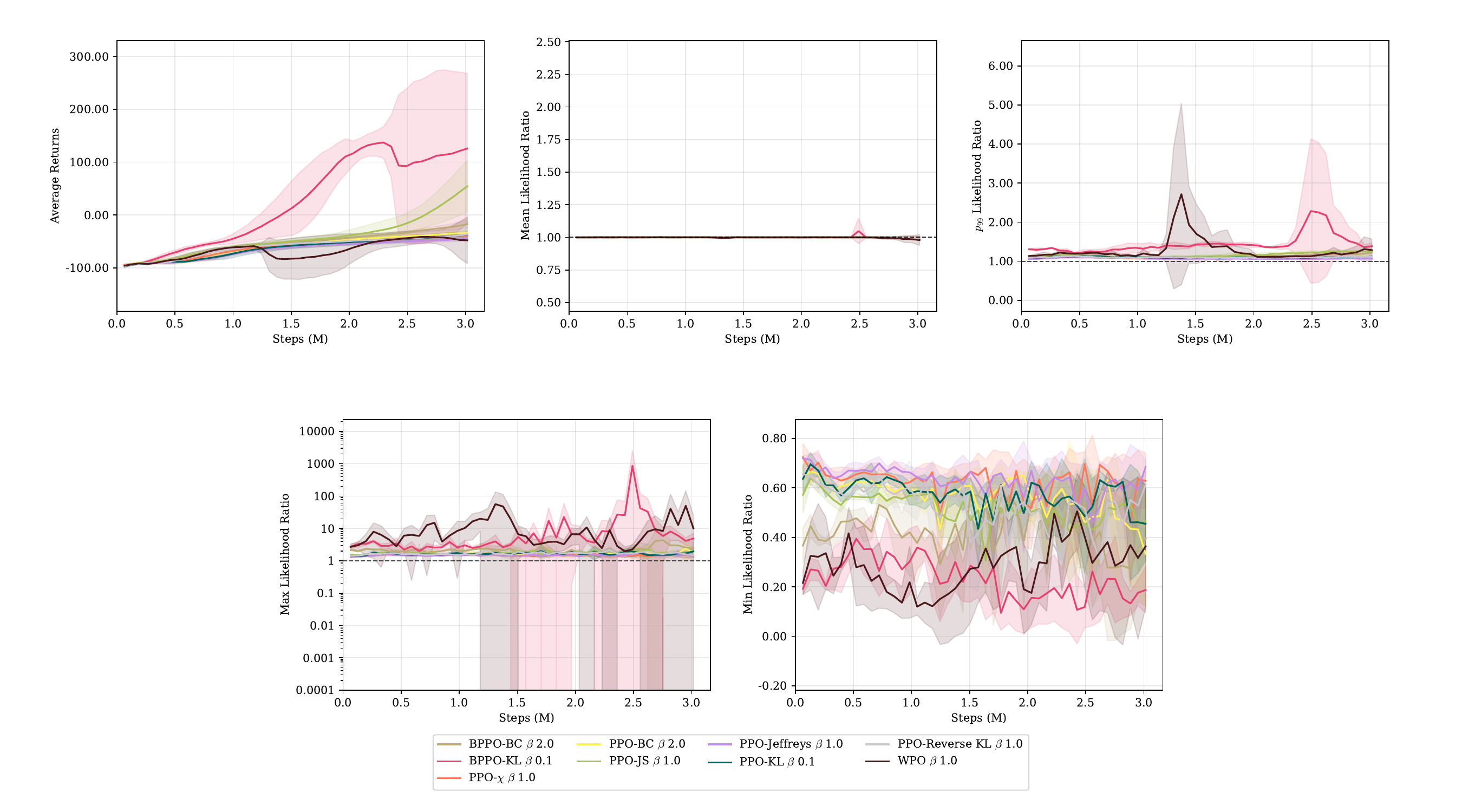}
    \caption{\textsc{BipedalWalker-v3} regularized baseline learning curves (4 seeds).}
\label{fig:bipedal_walk_regularized}
\end{figure*}

\noindent\textbf{Regularization can over-constrain learning.}
Figure \ref{fig:bipedal_walk_regularized} compares multiple regularizers (BC/Hellinger-style penalties, KL variants, $f$-divergence penalties, and WPO \cite{pfau2025wasserstein}/VPG-like baselines).
A recurring qualitative pattern is that several regularizers produce “safe-looking” ratio statistics (smaller max spikes, higher minimum ratios), yet the return curve flattens early.
This is the typical over-regularization regime: the update becomes too conservative to make coordinated locomotion improvements.

\noindent\textbf{Practical takeaway.}
For this task, the best-performing behavior is not “minimize tails at all costs,” but rather “moderate tails while maintaining effective update magnitude.”
BPPO’s square-root clipping empirically appears to strike that balance better than many explicit penalty choices.

\subsection{Regularized terms with entropy bonus}
\begin{figure*}[ht]  
    \centering
    \includegraphics[width=\textwidth]{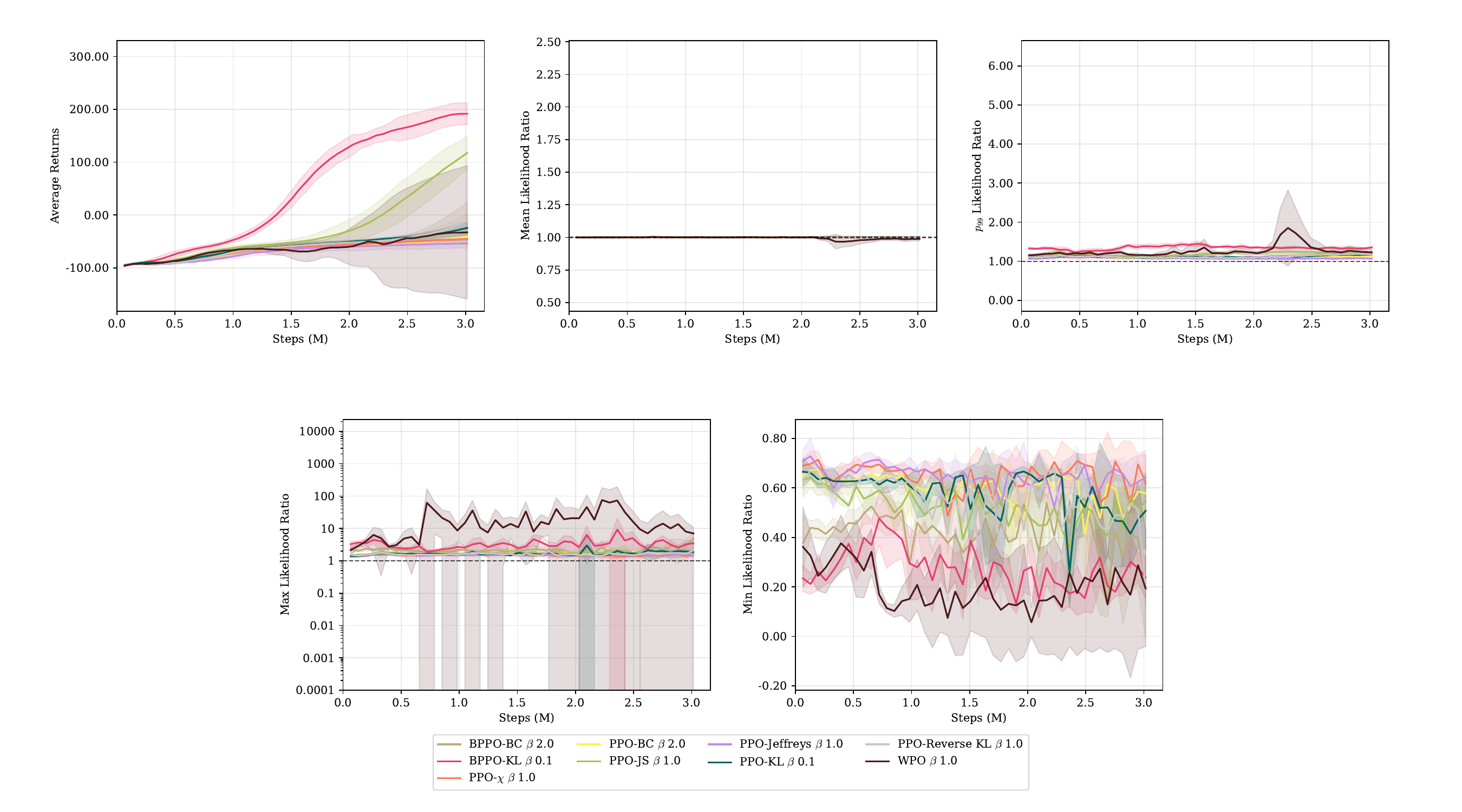}
    \caption{\textsc{BipedalWalker-v3} regularized baseline with entropy bonus learning curves (4 seeds).}
\label{fig:bipedal_walk_regularized_ent}
\end{figure*}

\noindent\textbf{Entropy partially offsets over-constraint.}
Figure \ref{fig:bipedal_walk_regularized_ent} shows that entropy can partially counteract conservative regularizers by encouraging exploration, but overly strong penalties can still suppress progress.
The ratio plots are again informative: some regularizers keep ratios extremely close to 1 (both mean and tails), which indicates very small effective policy change and correlates with weak learning.

\begin{figure*}[ht]
\centering

\setlength{\fboxrule}{0.4pt}
\setlength{\fboxsep}{1.5pt}
\setlength{\tabcolsep}{2pt}  
\renewcommand{\arraystretch}{1} 

\newcommand{\frameimg}[2]{\fbox{\includegraphics[width=#1]{#2}}}

\begin{tabular}{ccccc}
\frameimg{0.17\textwidth}{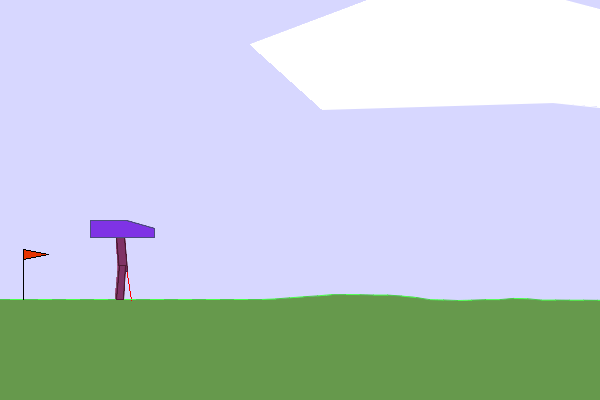} &
\frameimg{0.17\textwidth}{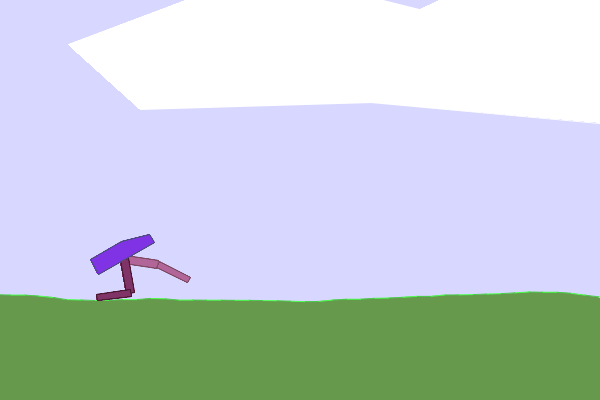} &
\frameimg{0.17\textwidth}{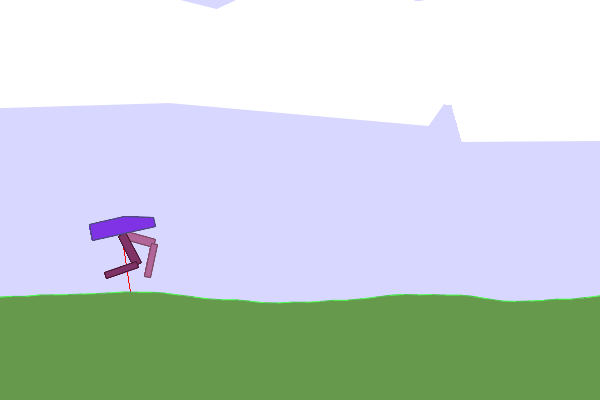} &
\frameimg{0.17\textwidth}{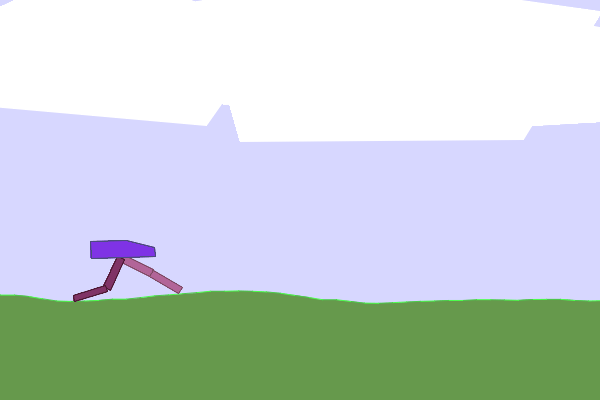} &
\frameimg{0.17\textwidth}{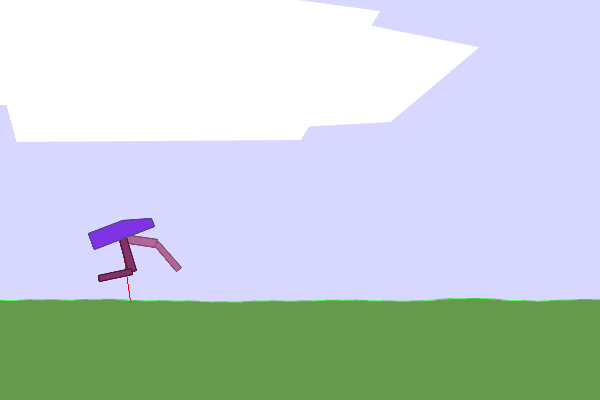} \\[4pt]
\end{tabular}
\caption{BPPO with \(\text{entropy} = 0.01\) and \(\beta = 0\) \textsc{BipedalWalker-v3} inference.}
\label{fig:bipedal_walk_inference}
\end{figure*}

\subsection{IQM Scores}
\noindent\textbf{Robust summary.}
Table \ref{tab:iqm_bipedalwalker} confirms the plot-level trends.
The best robust score is achieved by \textbf{BPPO with entropy $0.01$} (IQM $193.63$), while many explicitly regularized variants (especially BC-penalized ones) achieve negative IQM.
Notably, PPO is positive (IQM $\approx 100$), but trails BPPO+entropy, suggesting that square-root clipping plus mild exploration provides the strongest overall robustness on this environment.

\begin{table}[ht]
\centering
\caption{IQM scores on \textsc{BipedalWalker-v3} (4 seeds). Best result in bold.}
\label{tab:iqm_bipedalwalker}
\begin{tabular}{lcccc}
\toprule
Algorithm & Entropy & $\beta$ & IQM Score & Seeds \\
\midrule
BPPO & 0.0  & 0.0 & 93.86 & 4 \\
BPPO & 0.01 & 0.0 & \textbf{193.63} & 4 \\

BPPO-BC & 0.0  & 2.0 & -34.37 & 4 \\
BPPO-BC & 0.01 & 2.0 & -46.37 & 4 \\

BPPO-KL & 0.0  & 0.1 & 149.56 & 4 \\
BPPO-KL & 0.01 & 0.1 & 148.20 & 4 \\

BTRPO & 0.0  & 2.0 & -56.09 & 4 \\
BTRPO & 0.01 & 2.0 & -51.99 & 4 \\

PPO & 0.0  & 0.0 & 100.91 & 4 \\
PPO & 0.01 & 0.0 & 99.19 & 4 \\

PPO-$\chi$ & 0.0  & 1.0 & -46.46 & 4 \\
PPO-$\chi$ & 0.01 & 1.0 & -51.62 & 4 \\

PPO-BC & 0.0  & 2.0 & -41.26 & 4 \\
PPO-BC & 0.01 & 2.0 & -51.87 & 4 \\

PPO-JS & 0.0  & 1.0 & -10.44 & 4 \\
PPO-JS & 0.01 & 1.0 & 26.94 & 4 \\

PPO-Jeffreys & 0.0  & 1.0 & -49.78 & 4 \\
PPO-Jeffreys & 0.01 & 1.0 & -54.82 & 4 \\

PPO-KL & 0.0  & 1.0 & -45.94 & 4 \\
PPO-KL & 0.01 & 1.0 & -41.12 & 4 \\

PPO-Reverse KL & 0.0  & 1.0 & -45.22 & 4 \\
PPO-Reverse KL & 0.01 & 1.0 & -43.17 & 4 \\

TRPO & 0.0  & 0.1 & -5.70 & 4 \\
TRPO & 0.01 & 0.1 & -15.01 & 4 \\

WPO & 0.0  & 1.0 & -54.56 & 4 \\
WPO & 0.01 & 1.0 & -78.62 & 4 \\
\bottomrule
\end{tabular}
\end{table}

\newpage

\section{Mountain Car Continuous}
\subsection{Baselines}
\begin{figure*}[ht]
    \centering
    \includegraphics[width=\textwidth]{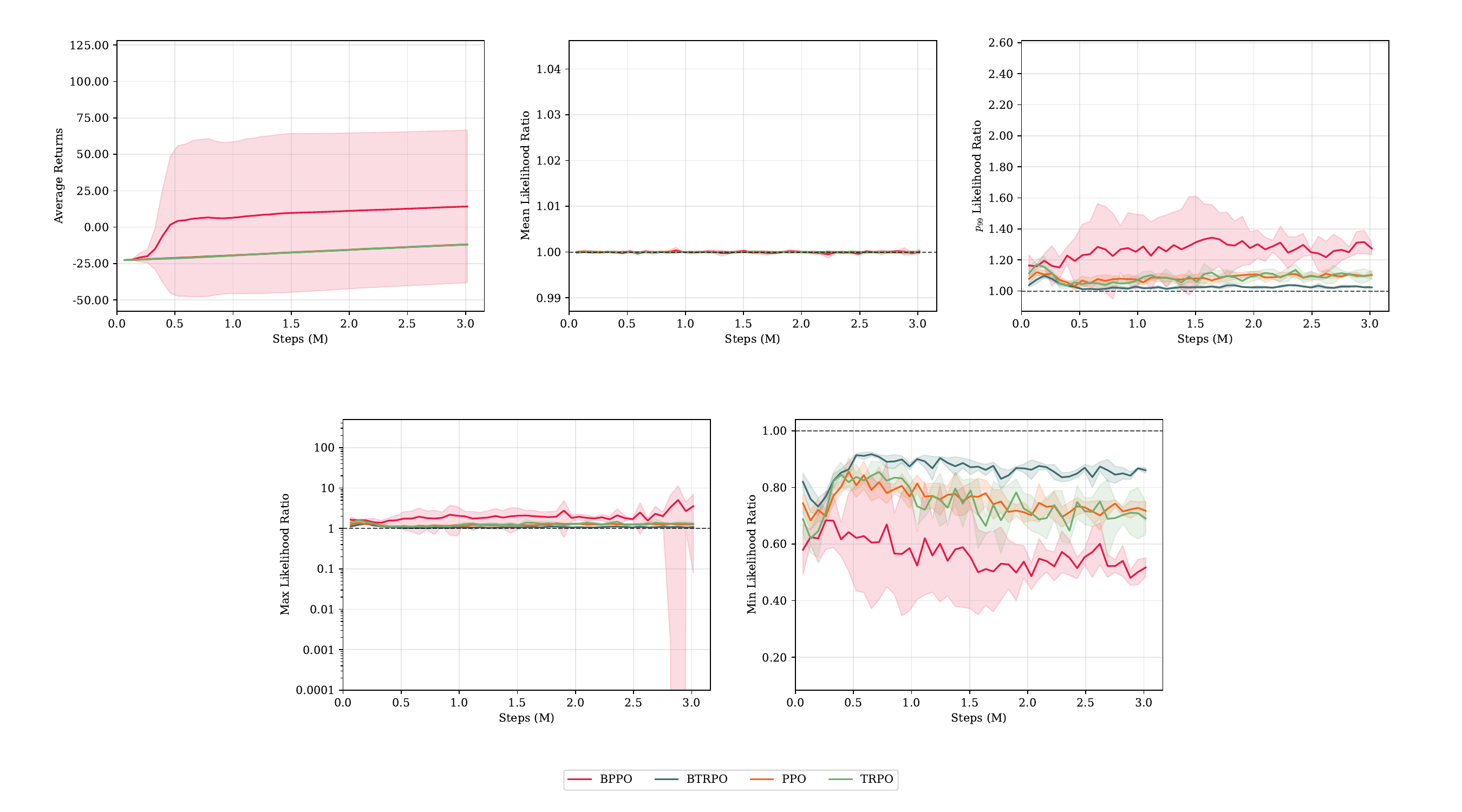}
    \caption{\textsc{MountainCarContinuous-v0} baseline learning curves (4 seeds).}
\label{fig:mountain_car_baseline}
\end{figure*}

\noindent\textbf{Learning dynamics.}
Figure \ref{fig:mountain_car_baseline} shows that MountainCarContinuous is highly sensitive under the matched-budget setup.
BPPO is the only baseline that consistently escapes the near-failure regime early, while PPO/TRPO/BTRPO remain close to a low plateau for much of training.

\noindent\textbf{What the ratio plots suggest.}
Across methods, the mean likelihood ratio stays close to 1, so the performance gap is not explained by systematic drift.
Instead, the difference is in how methods handle rare but important policy changes: BPPO retains a slightly more active upper tail (visible in $p_{99}$) without inducing catastrophic max-ratio spikes.

\subsection{Baselines with entropy bonus}
\begin{figure*}[ht]
    \centering
    \includegraphics[width=\textwidth]{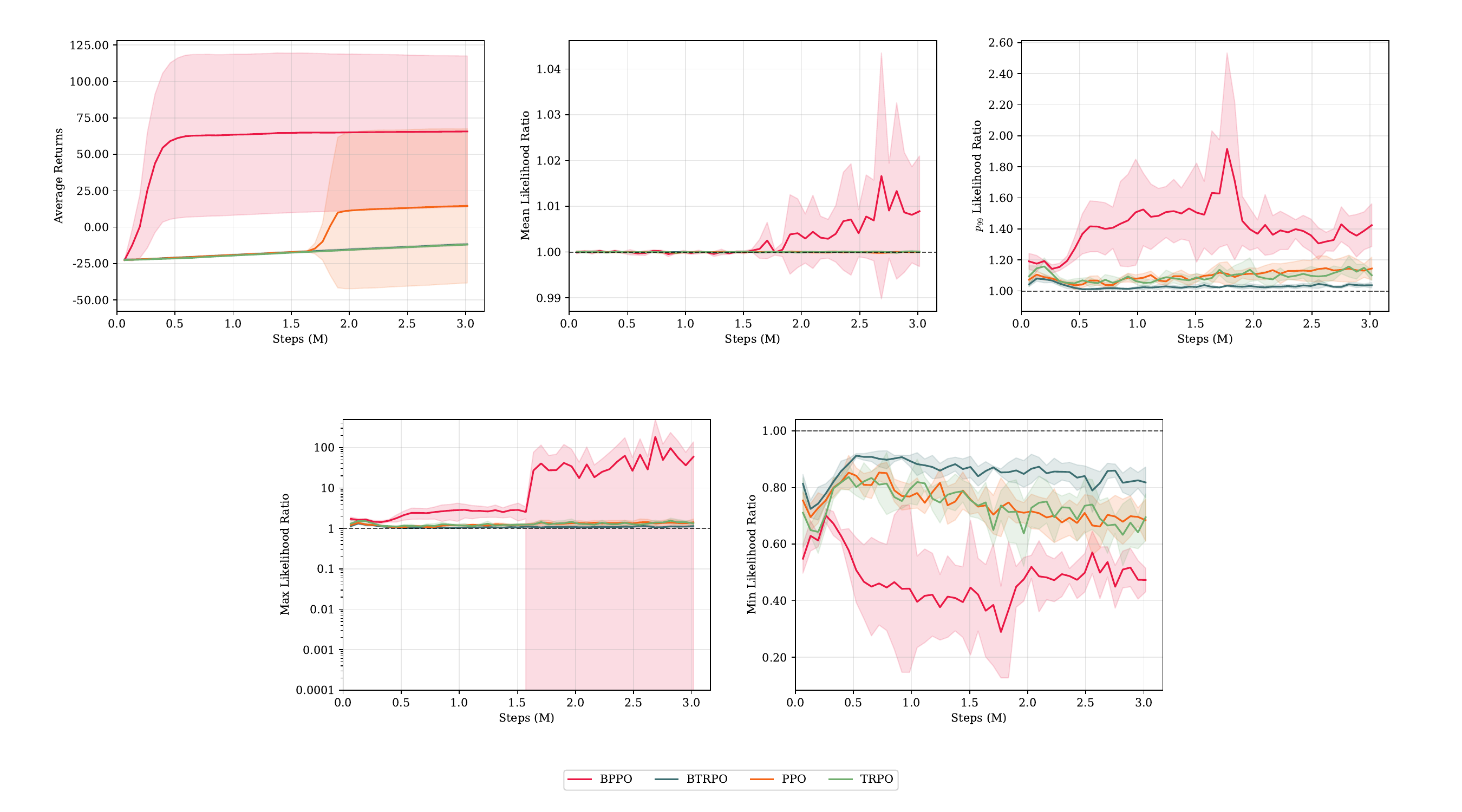}
    \caption{\textsc{MountainCarContinuous-v0} baseline with entropy bonus learning curves (4 seeds).}
\label{fig:mountain_car_baseline_ent}
\end{figure*}

\noindent\textbf{Effect of entropy.}
\noindent\textbf{Entropy is essential here.}
Figure \ref{fig:mountain_car_baseline_ent} shows that adding entropy dramatically improves BPPO, yielding fast progress to high returns.
Other methods benefit far less: PPO improves later (or modestly), while TRPO/BTRPO remain weak.

\noindent\textbf{Tail behavior under entropy.}
Entropy increases exploration and can increase ratio variance.
BPPO may show larger upper-tail activity late in training, but because BPPO’s surrogate uses $q=\sqrt{r}$, the impact of large positive-ratio events is structurally damped compared to optimizing with $r$ directly.

\subsection{Regularized terms}
\begin{figure*}[ht]
    \centering
    \includegraphics[width=\textwidth]{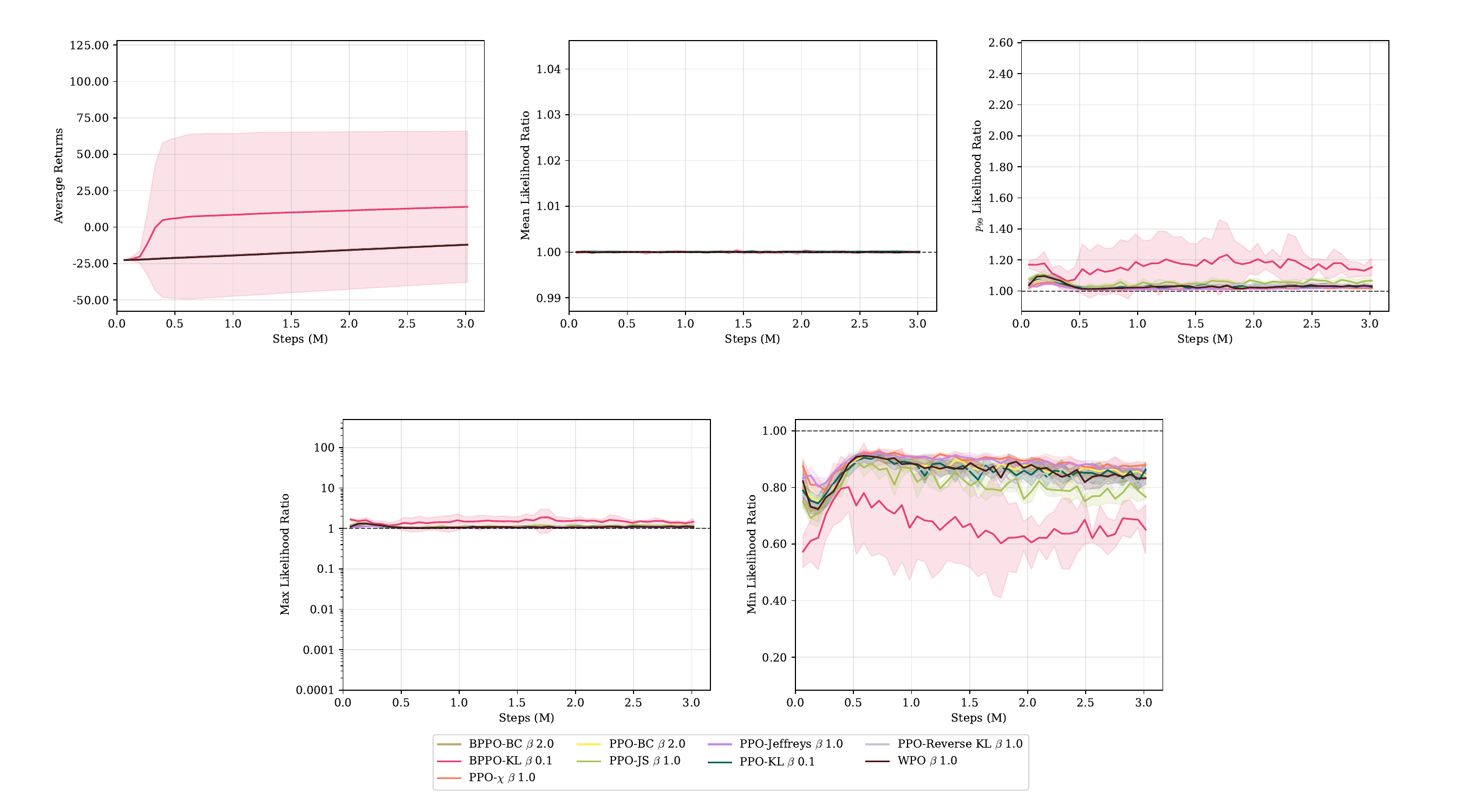}
    \caption{\textsc{MountainCarContinuous-v0} regularized baseline learning curves (4 seeds).}
\label{fig:mountain_car_regularized}
\end{figure*}

\noindent\textbf{Regularization often prevents escape.}
Figure \ref{fig:mountain_car_regularized} shows that many explicit penalty variants fail to escape the failure regime, even when their ratio tails appear benign.
This indicates that the main challenge is not only stability, but also maintaining enough effective update magnitude and exploration to discover the right control strategy.

\subsection{Regularized terms with entropy bonus}
\noindent\textbf{Regularization + entropy.}
Figure \ref{fig:mountain_car_regularized} shows that entropy can help some regularized variants, but BPPO remains the clearest winner.
Many penalty choices still underperform, supporting the conclusion that over-constraint is a dominant failure mode on this task.

\begin{figure*}[ht]  
    \centering
    \includegraphics[width=\textwidth]{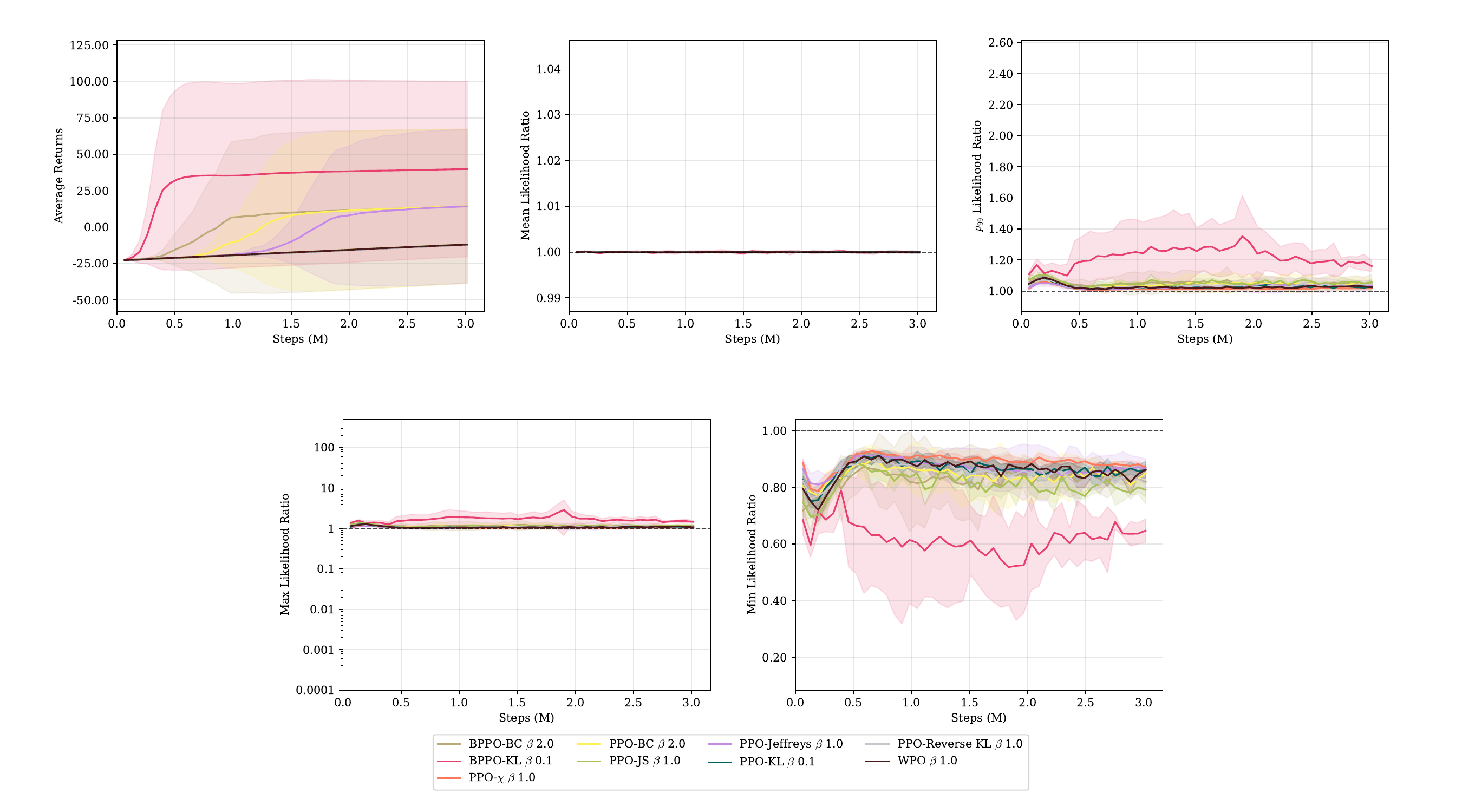}
    \caption{\textsc{MountainCarContinuous-v0} regularized baseline with entropy bonus learning curves (4 seeds).}
\label{fig:mountain_car_regularized_ent}
\end{figure*}

\begin{figure*}[ht]
\centering

\setlength{\fboxrule}{0.4pt}
\setlength{\fboxsep}{1.5pt}
\setlength{\tabcolsep}{2pt}  
\renewcommand{\arraystretch}{1} 

\newcommand{\frameimg}[2]{\fbox{\includegraphics[width=#1]{#2}}}

\begin{tabular}{ccccc}
\frameimg{0.17\textwidth}{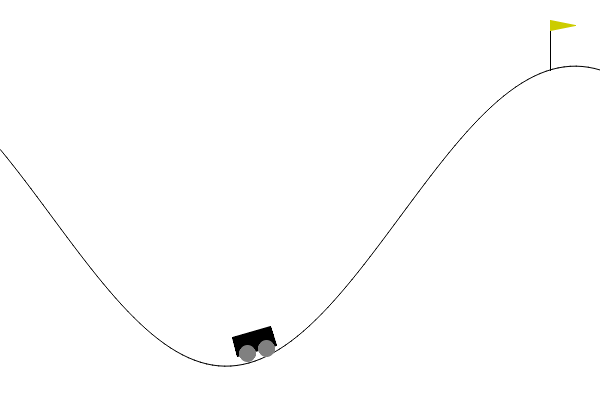} &
\frameimg{0.17\textwidth}{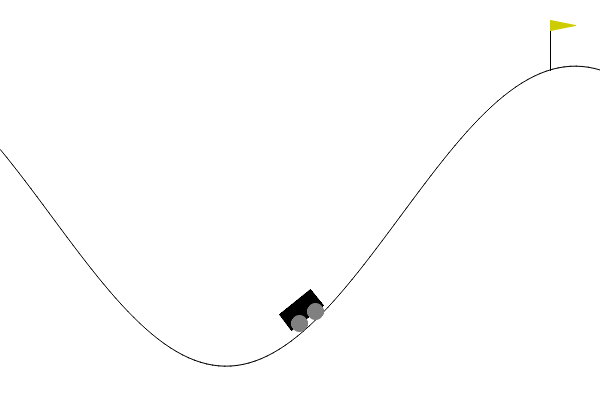} &
\frameimg{0.17\textwidth}{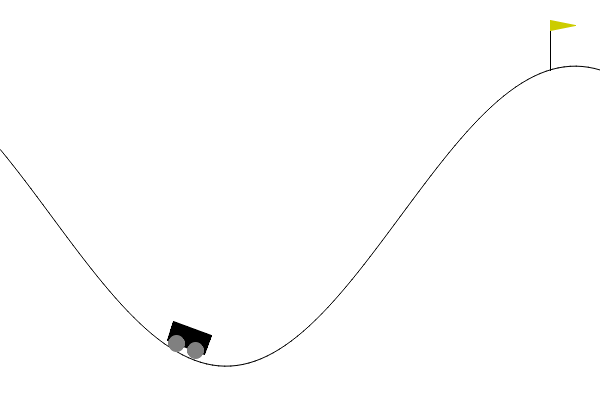} &
\frameimg{0.17\textwidth}{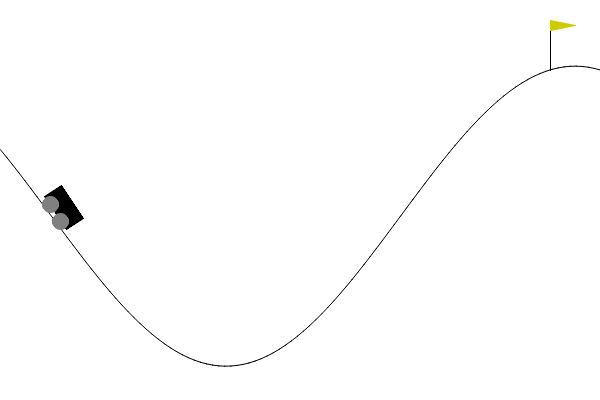} &
\frameimg{0.17\textwidth}{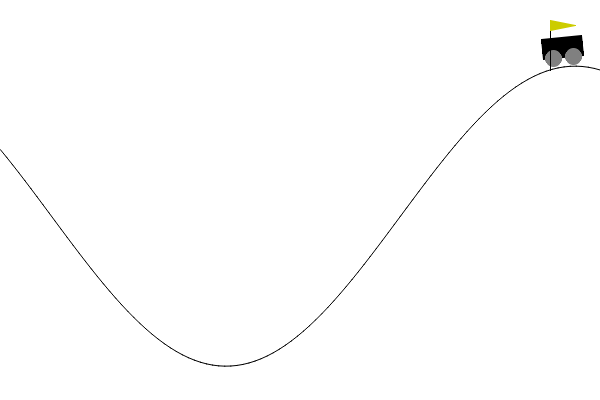} \\[4pt]
\end{tabular}
\caption{BPPO with \(\text{entropy} = 0.01\) and \(\beta = 0\) \textsc{MountainCarContinuous-v0} inference.}
\label{fig:mountain_car_inference}
\end{figure*}

\subsection{IQM Scores}
\noindent\textbf{Robust summary (Table~5).}
Table \ref{tab:iqm_mountaincarcontinuous} shows a stark separation: \textbf{BPPO with entropy $0.01$} achieves the best IQM ($91.77$), while most other configurations remain near the near-failure floor (IQM $\approx -14$).
The only other configuration that meaningfully improves is BPPO-KL with entropy (IQM $38.98$).
Overall, MountainCarContinuous appears to be an exploration-limited regime where square-root clipping plus mild entropy provides the right balance of stability and progress.

\begin{table}[t]
\centering
\caption{IQM scores across algorithms on \textsc{MountainCarContinuous-v0} (4 seeds). Best result in bold.}
\label{tab:iqm_mountaincarcontinuous}
\begin{tabular}{lcccc}
\toprule
Algorithm & Entropy & $\beta$ & IQM Score & Seeds \\
\midrule
BPPO & 0.0  & 0.0 & -14.05 & 4 \\
BPPO & 0.01 & 0.0 & \textbf{91.77} & 4 \\

BPPO-BC & 0.0  & 2.0 & -14.30 & 4 \\
BPPO-BC & 0.01 & 2.0 & -14.14 & 4 \\

BPPO-KL & 0.0  & 0.1 & -14.13 & 4 \\
BPPO-KL & 0.01 & 0.1 & 38.98  & 4 \\

BTRPO & 0.0  & 2.0 & -14.16 & 4 \\
BTRPO & 0.01 & 2.0 & -14.19 & 4 \\

PPO & 0.0  & 0.0 & -14.07 & 4 \\
PPO & 0.01 & 0.0 & -13.94 & 4 \\

PPO-$\chi$ & 0.0  & 1.0 & -14.30 & 4 \\
PPO-$\chi$ & 0.01 & 1.0 & -14.26 & 4 \\

PPO-BC & 0.0  & 2.0 & -14.08 & 4 \\
PPO-BC & 0.01 & 2.0 & -14.24 & 4 \\

PPO-JS & 0.0  & 1.0 & -14.12 & 4 \\
PPO-JS & 0.01 & 1.0 & -14.29 & 4 \\

PPO-Jeffreys & 0.0  & 1.0 & -14.12 & 4 \\
PPO-Jeffreys & 0.01 & 1.0 & -14.25 & 4 \\

PPO-KL & 0.0  & 1.0 & -14.22 & 4 \\
PPO-KL & 0.01 & 1.0 & -14.32 & 4 \\

PPO-Reverse KL & 0.0  & 1.0 & -14.00 & 4 \\
PPO-Reverse KL & 0.01 & 1.0 & -14.33 & 4 \\

TRPO & 0.0  & 0.1 & -14.08 & 4 \\
TRPO & 0.01 & 0.1 & -14.14 & 4 \\

WPO & 0.0  & 1.0 & -14.30 & 4 \\
WPO & 0.01 & 1.0 & -14.16 & 4 \\
\bottomrule
\end{tabular}
\end{table}

\newpage

\section{Lunar Lander}
\subsection{Baselines}
\begin{figure*}[ht]
    \centering
    \includegraphics[width=\textwidth]{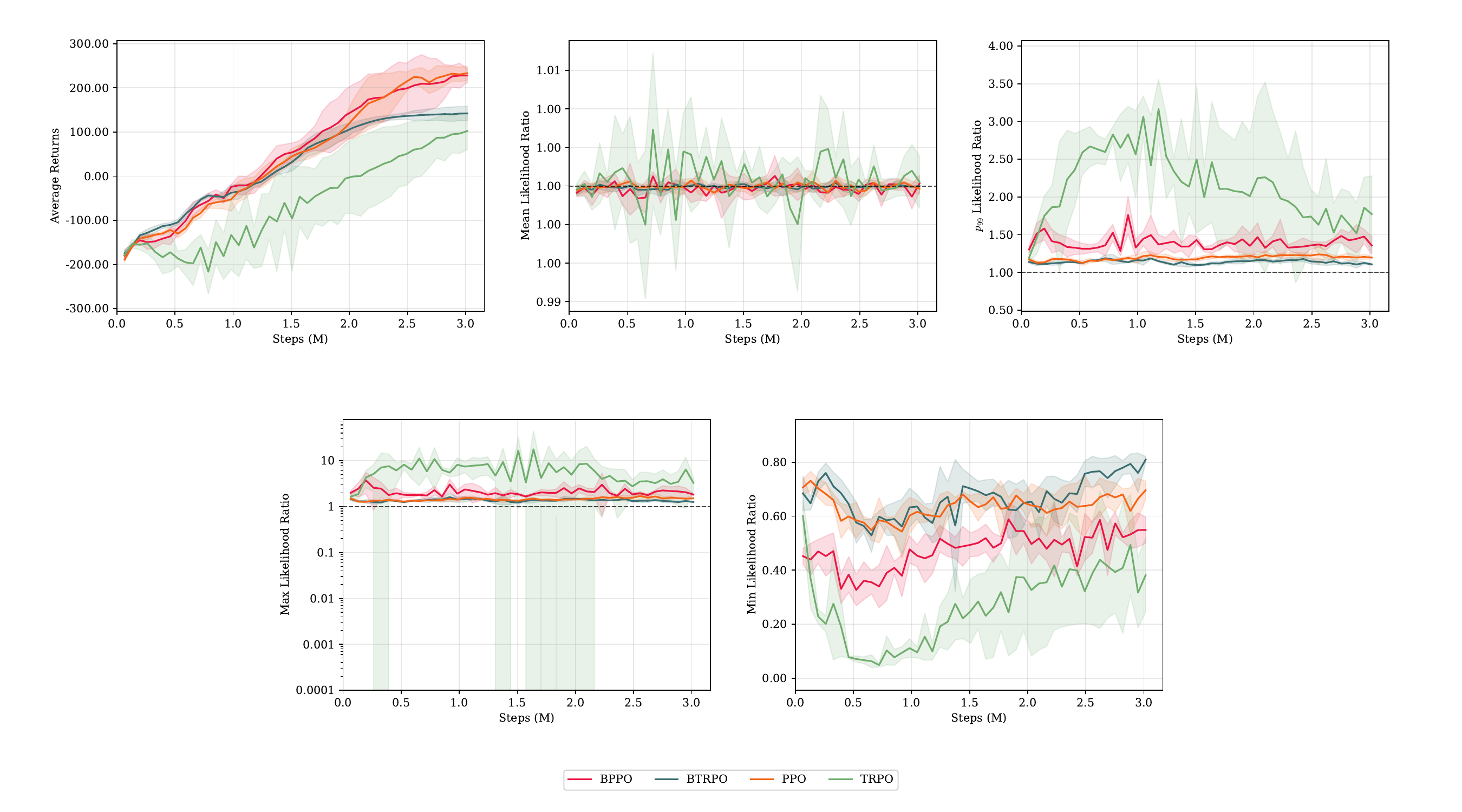}
    \caption{\textsc{LunarLander-v3} baseline learning curves (4 seeds).}
\label{fig:lunar_lander_baseline}
\end{figure*}

\noindent\textbf{Learning dynamics.}
Figure \ref{fig:lunar_lander_baseline} indicate that LunarLander is comparatively easier than locomotion: multiple methods reach strong performance under the same budget.
BPPO and PPO generally learn reliably, while TRPO lags and shows more variability.

\noindent\textbf{Ratio interpretation.}
The ratio plots still reveal qualitative differences:
TRPO exhibits larger upper-tail summaries (notably $p_{99}$ and max), indicating more extreme importance-weight events.
BPPO and PPO typically keep the mean ratio close to 1 and moderate the upper tail, which correlates with smoother learning.

\subsection{Baselines with entropy bonus}
\noindent\textbf{Entropy effects are task-dependent.}
Entropy can improve exploration early, but on LunarLander it does not uniformly help all methods.
In particular, BPPO often performs best without entropy, suggesting that once a good landing policy is discovered, additional exploration pressure can slightly degrade robustness.

\begin{figure*}[ht]
    \centering
    \includegraphics[width=\textwidth]{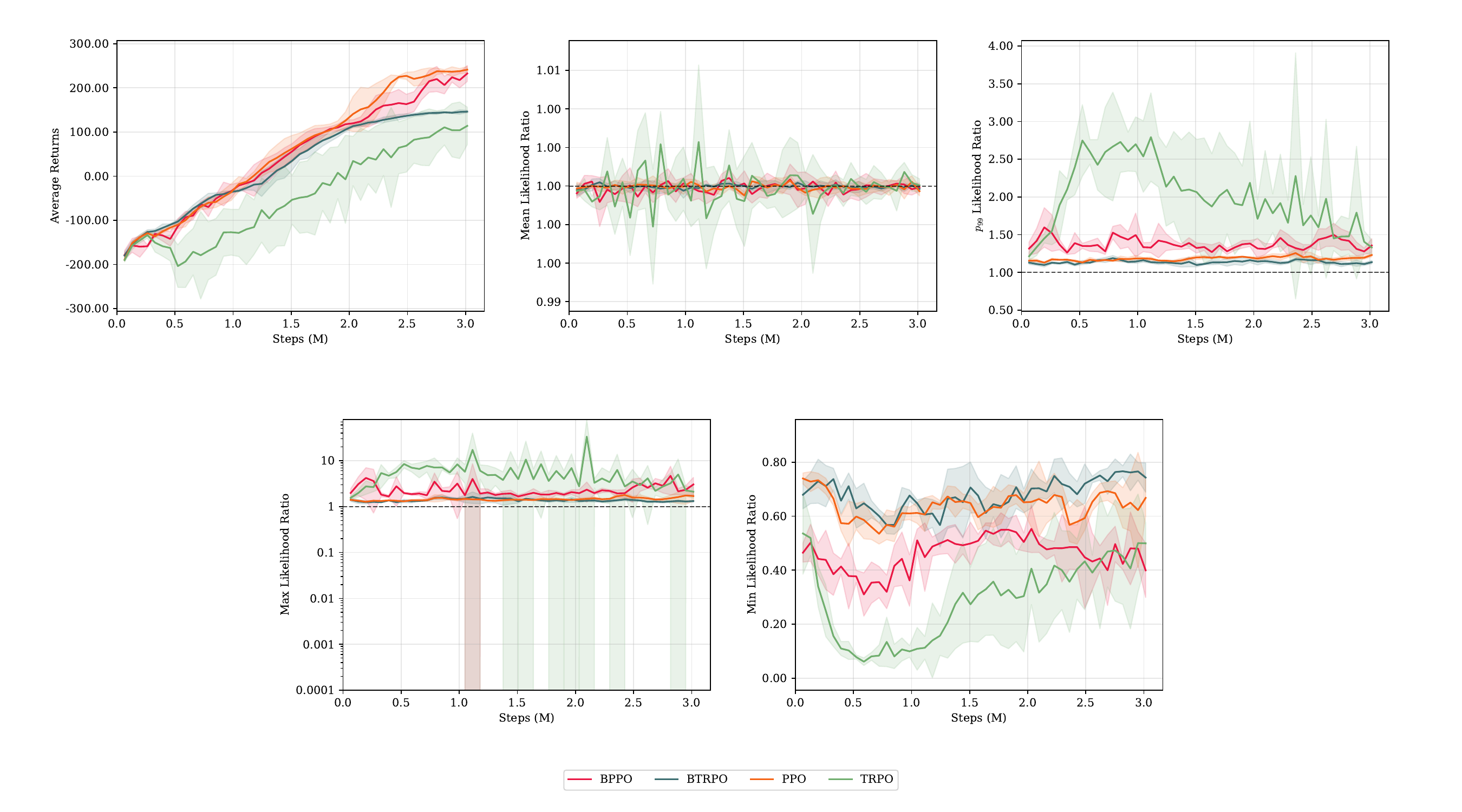}
    \caption{\textsc{LunarLander-v3} baseline with entropy bonus learning curves (4 seeds).}
\label{fig:lunar_lander_baseline_ent}
\end{figure*}

\subsection{Regularized terms}
\begin{figure*}[ht]
    \centering
    \includegraphics[width=\textwidth]{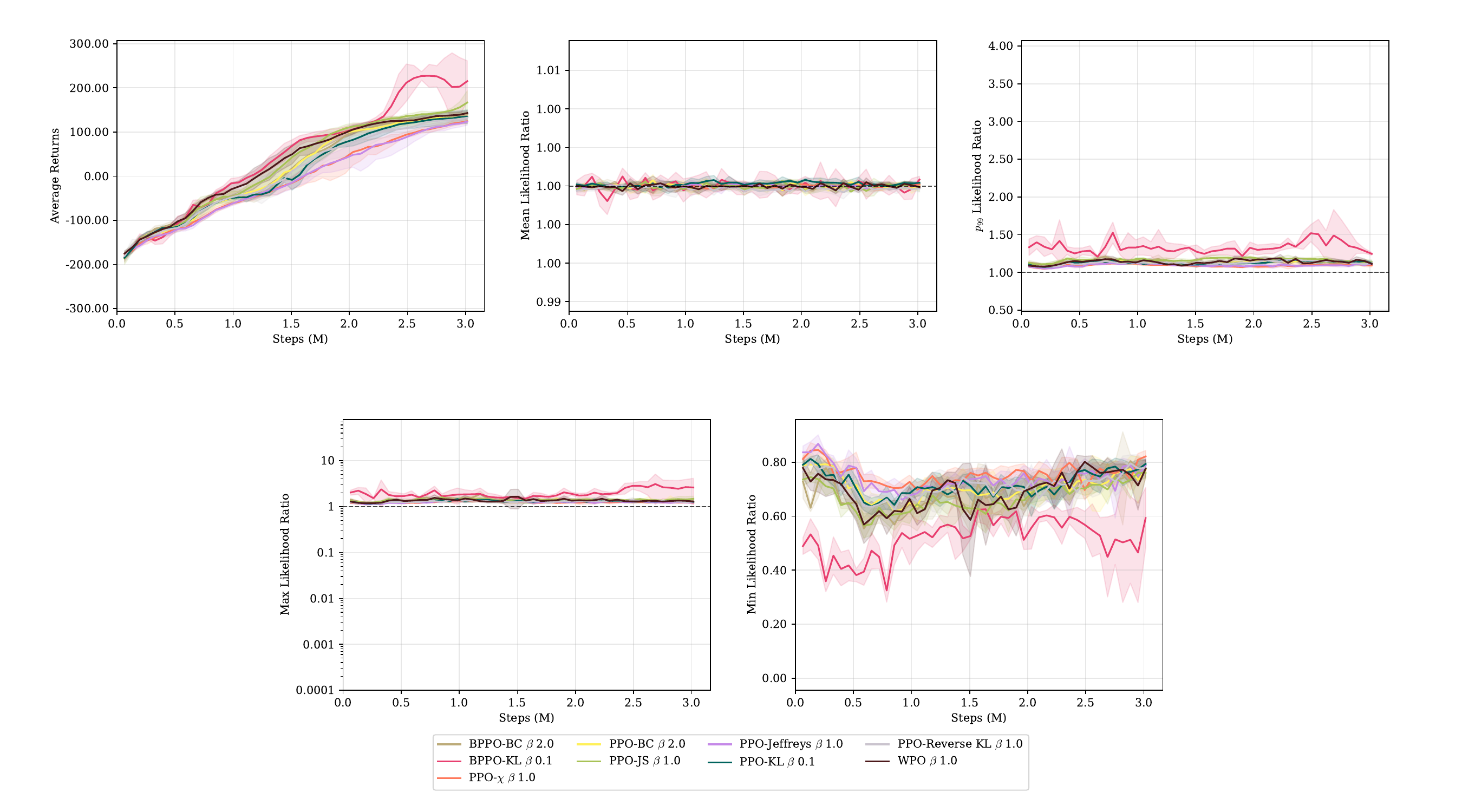}
    \caption{\textsc{LunarLander-v3} regularized baseline learning curves (4 seeds).}
\label{fig:lunar_lander_regularized}
\end{figure*}

\noindent\textbf{Many regularizers are viable here.}
Figure \ref{fig:lunar_lander_regularized} show that multiple regularized variants yield similar learning curves and only modest differences in tail statistics.
This suggests that LunarLander is less “tail-limited” under the tested settings, and that stability is achievable with several trust-region choices.

\subsection{Regularized terms with entropy bonus}
\noindent\textbf{Regularization + entropy.}
With entropy, regularization tends to compress tails (especially max ratio), but the return differences remain moderate.
Overall, the primary takeaway is that LunarLander does not strongly discriminate between geometries; many choices are competitive.

\begin{figure*}[ht]  
    \centering
    \includegraphics[width=\textwidth]{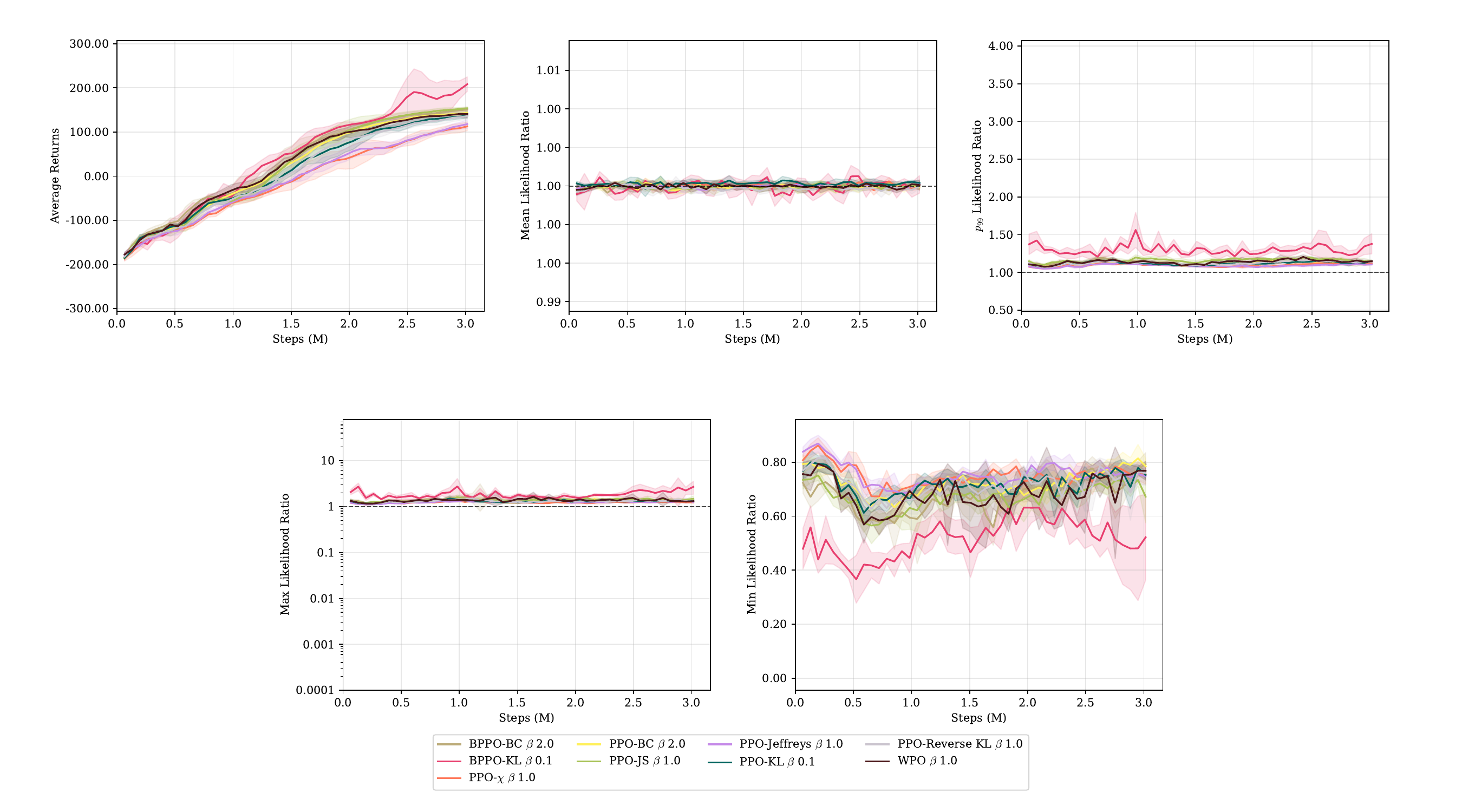}
    \caption{\textsc{LunarLander-v3} regularized baseline with entropy bonus learning curves (4 seeds).}
\label{fig:lunar_lander_regularized_ent}
\end{figure*}

\begin{figure*}[ht]
\centering

\setlength{\fboxrule}{0.4pt}
\setlength{\fboxsep}{1.5pt}
\setlength{\tabcolsep}{2pt}  
\renewcommand{\arraystretch}{1} 

\newcommand{\frameimg}[2]{\fbox{\includegraphics[width=#1]{#2}}}

\begin{tabular}{ccccc}
\frameimg{0.17\textwidth}{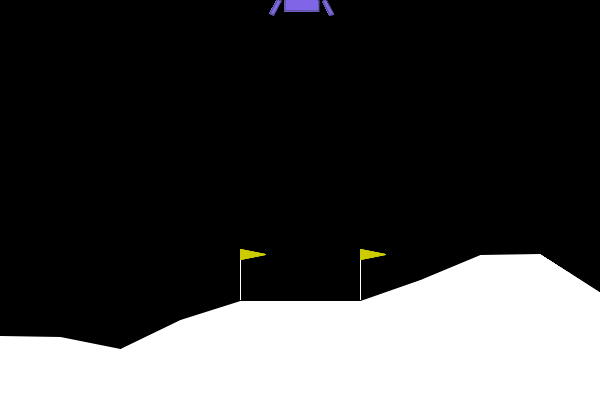} &
\frameimg{0.17\textwidth}{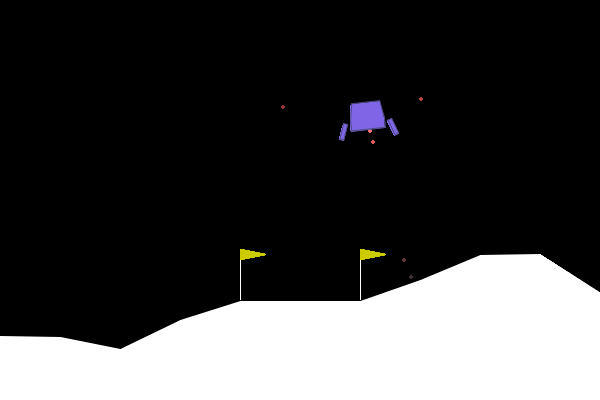} &
\frameimg{0.17\textwidth}{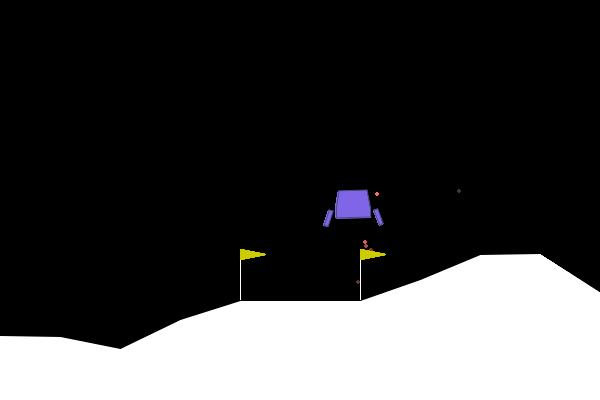} &
\frameimg{0.17\textwidth}{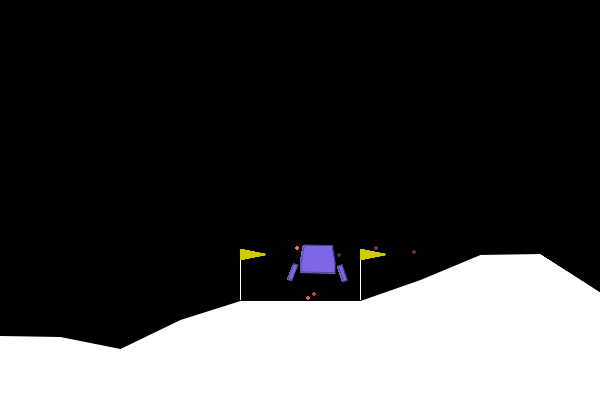} &
\frameimg{0.17\textwidth}{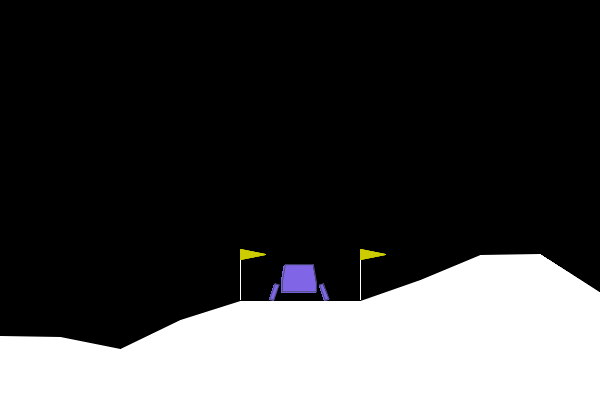} \\[4pt]
\end{tabular}
\caption{BPPO with \(\text{entropy} = 0.0\) and \(\beta = 0\) \textsc{LunarLander-v3} inference.}
\label{fig:lunar_lander_inference}
\end{figure*}

\subsection{IQM Scores}
\noindent\textbf{Robust summary.}
Table \ref{tab:iqm_lunarlander} shows \textbf{BPPO without entropy} achieves the best IQM ($186.72$), while \textbf{PPO with entropy $0.01$} is very close ($185.00$).
BPPO-KL without entropy is also strong ($166.01$), and several BC-penalized variants remain competitive but below the top performers.
This supports the view that overlap-based updates are competitive on LunarLander, but the main benefit is robustness rather than a dramatic performance separation.

\begin{table}[t]
\centering
\caption{IQM scores across algorithms on \textsc{LunarLander-v3} (4 seeds). Best result in bold.}
\label{tab:iqm_lunarlander}
\begin{tabular}{lcccc}
\toprule
Algorithm & Entropy & $\beta$ & IQM Score & Seeds \\
\midrule
BPPO & 0.0  & 0.0 & \textbf{186.72} & 4 \\
BPPO & 0.01 & 0.0 & 162.82 & 4 \\

BPPO-BC & 0.0  & 2.0 & 116.64 & 4 \\
BPPO-BC & 0.01 & 2.0 & 125.03 & 4 \\

BPPO-KL & 0.0  & 0.1 & 166.01 & 4 \\
BPPO-KL & 0.01 & 0.1 & 149.15 & 4 \\

BTRPO & 0.0  & 2.0 & 126.32 & 4 \\
BTRPO & 0.01 & 2.0 & 125.47 & 4 \\

PPO & 0.0  & 0.0 & 174.80 & 4 \\
PPO & 0.01 & 0.0 & 185.00 & 4 \\

PPO-$\chi$ & 0.0  & 1.0 & 80.63 & 4 \\
PPO-$\chi$ & 0.01 & 1.0 & 74.13 & 4 \\

PPO-BC & 0.0  & 2.0 & 115.07 & 4 \\
PPO-BC & 0.01 & 2.0 & 120.76 & 4 \\

PPO-JS & 0.0  & 1.0 & 127.31 & 4 \\
PPO-JS & 0.01 & 1.0 & 132.25 & 4 \\

PPO-Jeffreys & 0.0  & 1.0 & 81.16 & 4 \\
PPO-Jeffreys & 0.01 & 1.0 & 75.95 & 4 \\

PPO-KL & 0.0  & 1.0 & 108.63 & 4 \\
PPO-KL & 0.01 & 1.0 & 104.31 & 4 \\

PPO-Reverse KL & 0.0  & 1.0 & 102.18 & 4 \\
PPO-Reverse KL & 0.01 & 1.0 & 102.92 & 4 \\

TRPO & 0.0  & 0.1 & 37.85 & 4 \\
TRPO & 0.01 & 0.1 & 72.12 & 4 \\

WPO & 0.0  & 1.0 & 118.14 & 4 \\
WPO & 0.01 & 1.0 & 119.18 & 4 \\
\bottomrule
\end{tabular}
\end{table}

\newpage

\section{Frozen Lake}
\subsection{Baselines}
\begin{figure*}[t]
    \centering
    \includegraphics[width=\textwidth]{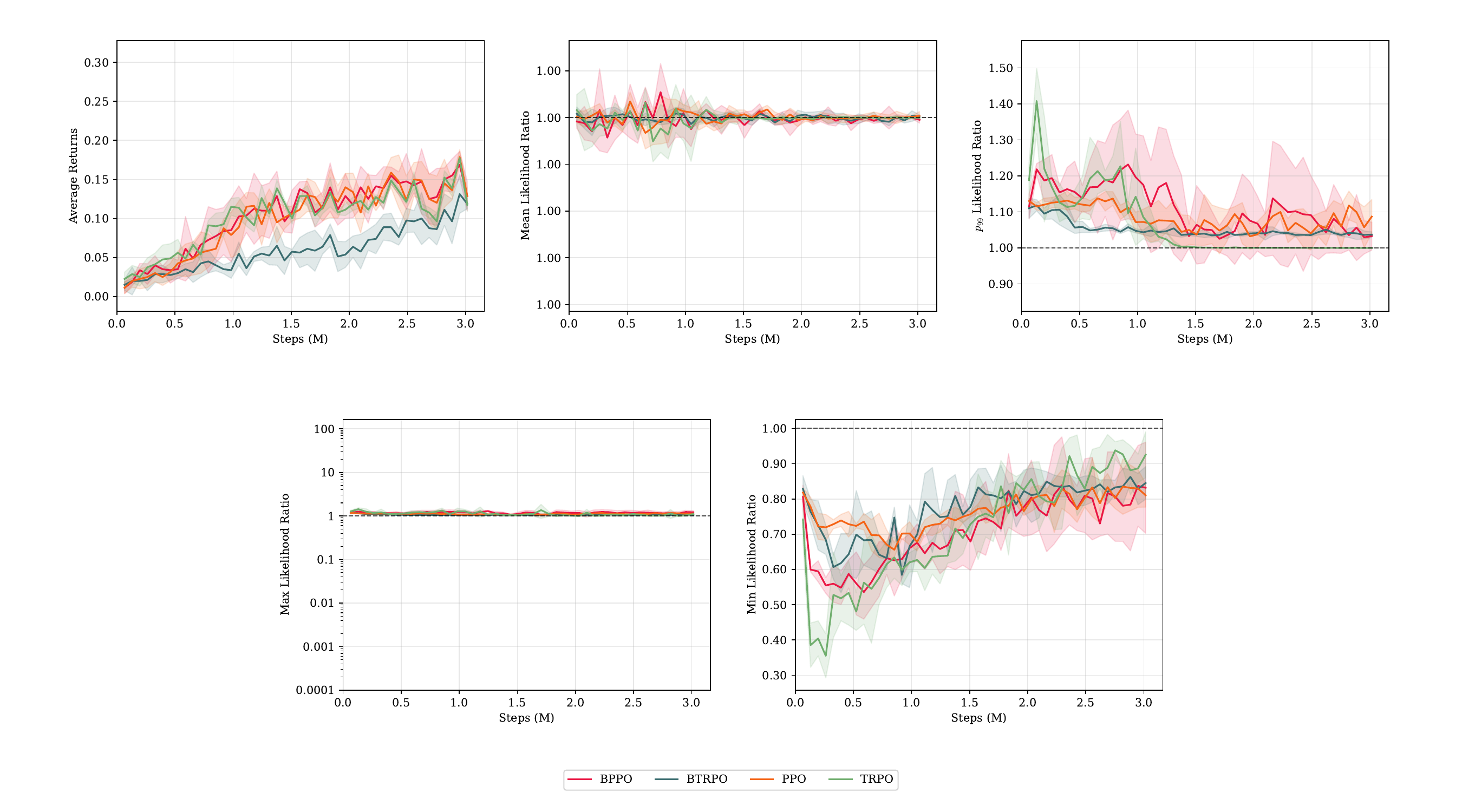}
    \caption{\textsc{FrozenLake-v1} baseline learning curves (4 seeds).}
\label{fig:frozen_lake_baseline}
\end{figure*}

\noindent\textbf{Learning dynamics.}
Figure \ref{fig:frozen_lake_baseline} shows that FrozenLake exhibits small return scales and relatively mild likelihood-ratio behavior.
Most methods improve gradually with small gaps between curves, and training is not dominated by heavy-tail events under this setup.

\noindent\textbf{Ratio interpretation.}
Mean ratios remain close to 1 and the upper-tail summaries are modest, indicating that tail control is not a primary bottleneck here.
As a result, differences tend to reflect optimization variance rather than strong geometric effects.

\subsection{Baselines with entropy bonus}
\begin{figure*}[t]
    \centering
    \includegraphics[width=\textwidth]{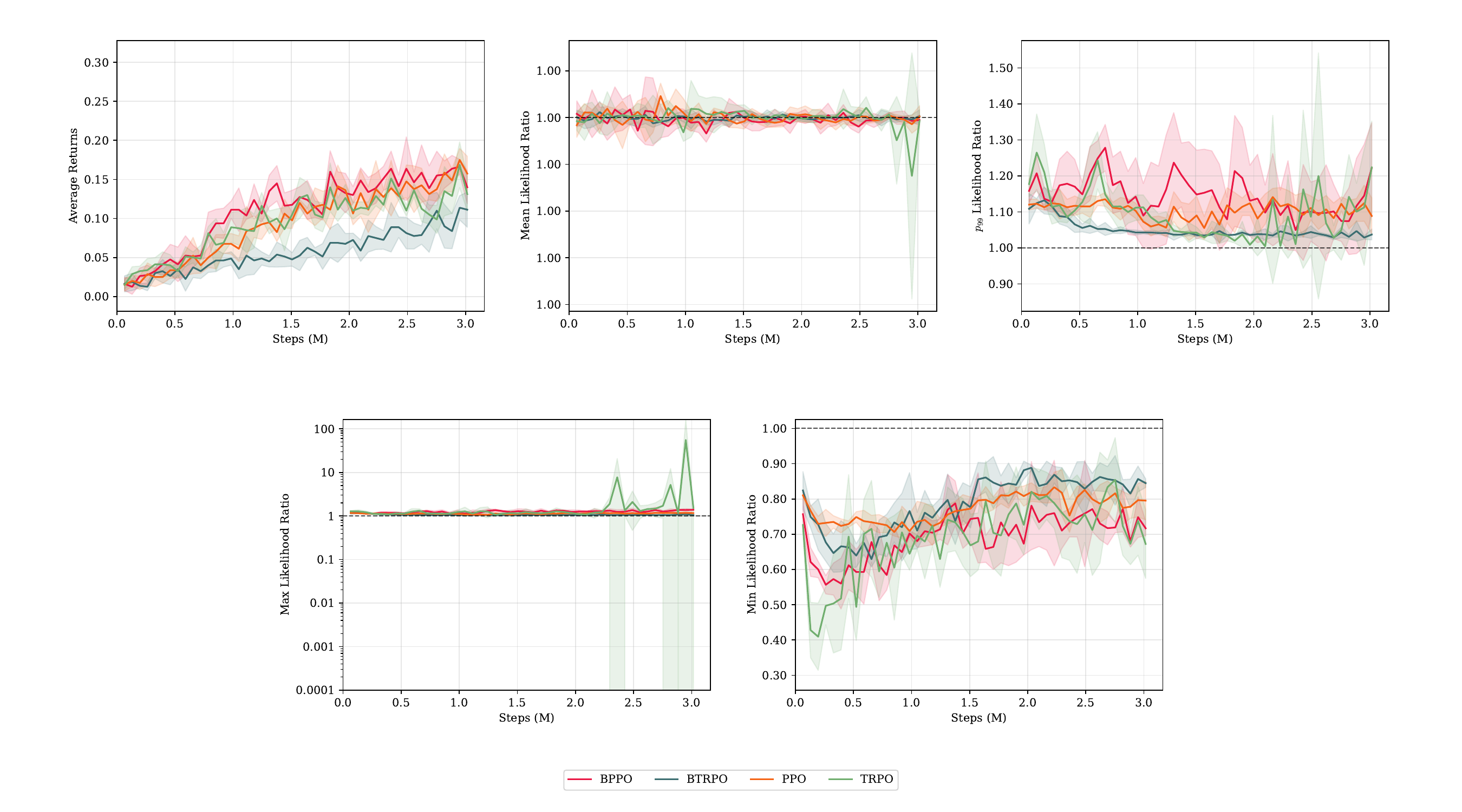}
    \caption{\textsc{FrozenLake-v1} baseline with entropy bonus learning curves (4 seeds).}
\label{fig:frozen_lake_baseline_ent}
\end{figure*}

\noindent\textbf{Entropy helps exploration modestly.}
Figure \ref{fig:frozen_lake_baseline_ent} shows entropy can slightly improve exploration and returns, but does not substantially change the ordering.
Ratio tails remain small overall, reinforcing that this is not a heavy-tail stress test.

\subsection{Regularized terms}
\begin{figure*}[t]
    \centering
    \includegraphics[width=\textwidth]{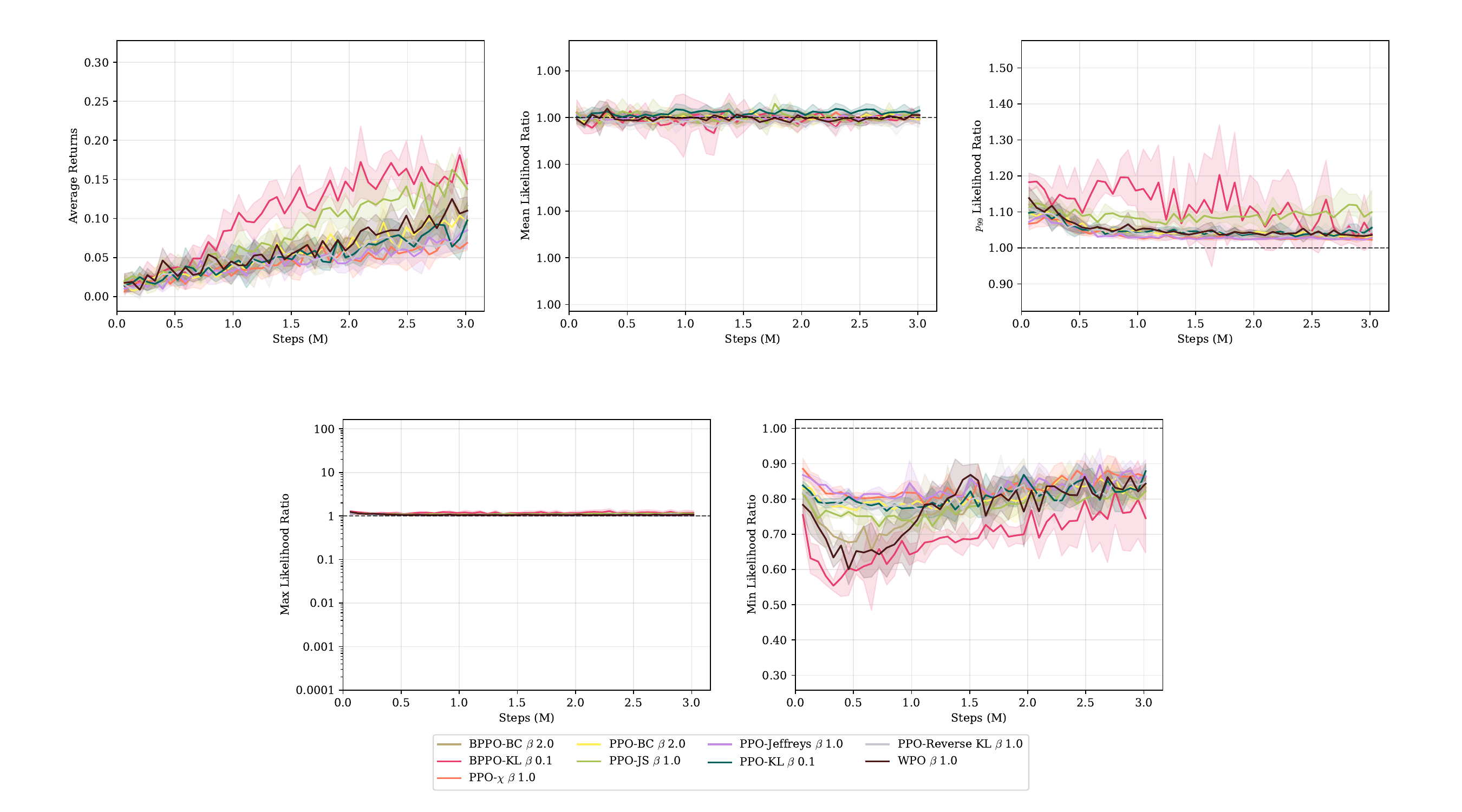}
    \caption{\textsc{FrozenLake-v1} regularized baseline learning curves (4 seeds).}
\label{fig:frozen_lake_regularized}
\end{figure*}

\noindent\textbf{Regularization changes little.}
Figure \ref{fig:frozen_lake_regularized} indicates that most regularizers behave similarly on this environment.
Because updates remain small, many divergences are effectively interchangeable.

\subsection{Regularized terms with entropy bonus}
\noindent\textbf{Regularization + entropy.}
Figure \ref{fig:cartpole_regularized_ent} shows mild differences in tails but no major performance separation.
This aligns with the local-geometry story: when policy changes are small, overlap and KL-style controls behave similarly in practice.

\begin{figure*}[t]  
    \centering
    \includegraphics[width=\textwidth]{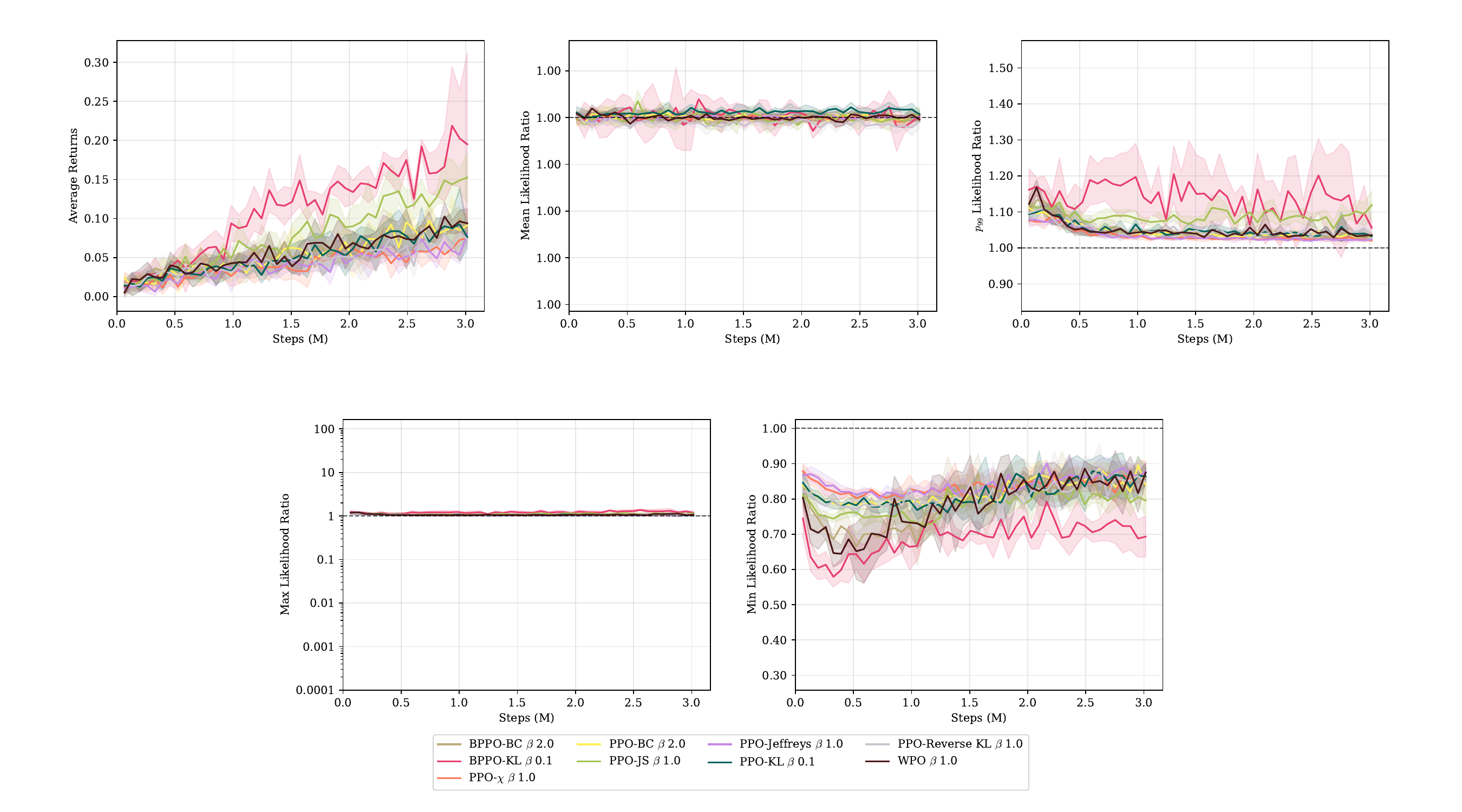}
    \caption{\textsc{FrozenLake-v1} regularized baseline with entropy bonus learning curves (4 seeds).}
\label{fig:frozen_lake_regularized_ent}
\end{figure*}

\begin{figure*}[ht]
\centering

\setlength{\fboxrule}{0.4pt}
\setlength{\fboxsep}{1.5pt}
\setlength{\tabcolsep}{2pt}  
\renewcommand{\arraystretch}{1} 

\newcommand{\frameimg}[2]{\fbox{\includegraphics[width=#1]{#2}}}

\begin{tabular}{ccccc}
\frameimg{0.17\textwidth}{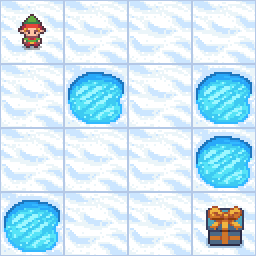} &
\frameimg{0.17\textwidth}{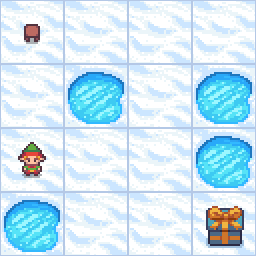} &
\frameimg{0.17\textwidth}{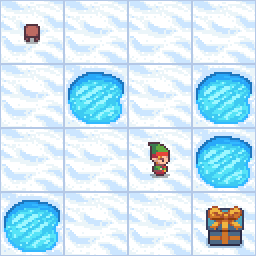} &
\frameimg{0.17\textwidth}{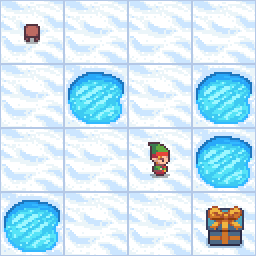} &
\frameimg{0.17\textwidth}{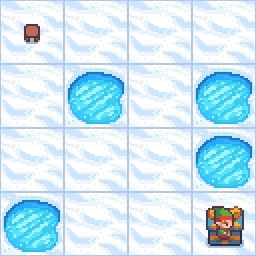} \\[4pt]
\end{tabular}
\caption{BPPO with \(\text{entropy} = 0.0\) and \(\beta = 0\) \textsc{FrozenLake-v1} inference.}
\label{fig:frozen_lake_inference}
\end{figure*}

\subsection{IQM Scores}
\noindent\textbf{Robust summary.}
Table \ref{tab:iqm_frozenlake} shows very tight clustering.
The best IQM is achieved by BPPO-KL without entropy (IQM $0.16$), but several variants are close (e.g., BPPO with entropy $0.01$ at $0.15$, PPO variants around $0.13$--$0.14$).
Overall, FrozenLake provides a sanity-check regime where geometry choice has limited effect.

\begin{table}[t]
\centering
\caption{IQM scores across algorithms on \textsc{FrozenLake-v1} (4 seeds). Best result in bold.}
\label{tab:iqm_frozenlake}
\begin{tabular}{lcccc}
\toprule
Algorithm & Entropy & $\beta$ & IQM Score & Seeds \\
\midrule
BPPO & 0.0  & 0.0 & 0.13 & 4 \\
BPPO & 0.01 & 0.0 & 0.15 & 4 \\

BPPO-BC & 0.0  & 2.0 & 0.08 & 4 \\
BPPO-BC & 0.01 & 2.0 & 0.07 & 4 \\

BPPO-KL & 0.0  & 0.1 & \textbf{0.16} & 4 \\
BPPO-KL & 0.01 & 0.1 & 0.15 & 4 \\

BTRPO & 0.0  & 2.0 & 0.09 & 4 \\
BTRPO & 0.01 & 2.0 & 0.08 & 4 \\

PPO & 0.0  & 0.0 & 0.13 & 4 \\
PPO & 0.01 & 0.0 & 0.14 & 4 \\

PPO-$\chi$ & 0.0  & 1.0 & 0.06 & 4 \\
PPO-$\chi$ & 0.01 & 1.0 & 0.06 & 4 \\

PPO-BC & 0.0  & 2.0 & 0.08 & 4 \\
PPO-BC & 0.01 & 2.0 & 0.08 & 4 \\

PPO-JS & 0.0  & 1.0 & 0.13 & 4 \\
PPO-JS & 0.01 & 1.0 & 0.12 & 4 \\

PPO-Jeffreys & 0.0  & 1.0 & 0.06 & 4 \\
PPO-Jeffreys & 0.01 & 1.0 & 0.05 & 4 \\

PPO-KL & 0.0  & 1.0 & 0.07 & 4 \\
PPO-KL & 0.01 & 1.0 & 0.07 & 4 \\

PPO-Reverse KL & 0.0  & 1.0 & 0.06 & 4 \\
PPO-Reverse KL & 0.01 & 1.0 & 0.06 & 4 \\

TRPO & 0.0  & 0.1 & 0.13 & 4 \\
TRPO & 0.01 & 0.1 & 0.12 & 4 \\

WPO & 0.0  & 1.0 & 0.09 & 4 \\
WPO & 0.01 & 1.0 & 0.08 & 4 \\
\bottomrule
\end{tabular}
\end{table}

\newpage

\section{Cartpole}
\subsection{Baselines}
\noindent\textbf{Learning dynamics.}
Figure~20 shows that \textsc{CartPole-v1} is close to a ceiling task in this setup: multiple methods achieve high returns quickly, and differences are mostly in learning speed and variability rather than final performance.

\noindent\textbf{Ratio interpretation.}
Even in an easy regime, ratio diagnostics can reveal stability differences: TRPO tends to show larger tail excursions (max and sometimes $p_{99}$) than clipping-based methods, but these excursions do not necessarily translate into lower final returns here because the task saturates.

\begin{figure*}[ht]
    \centering
    \includegraphics[width=\textwidth]{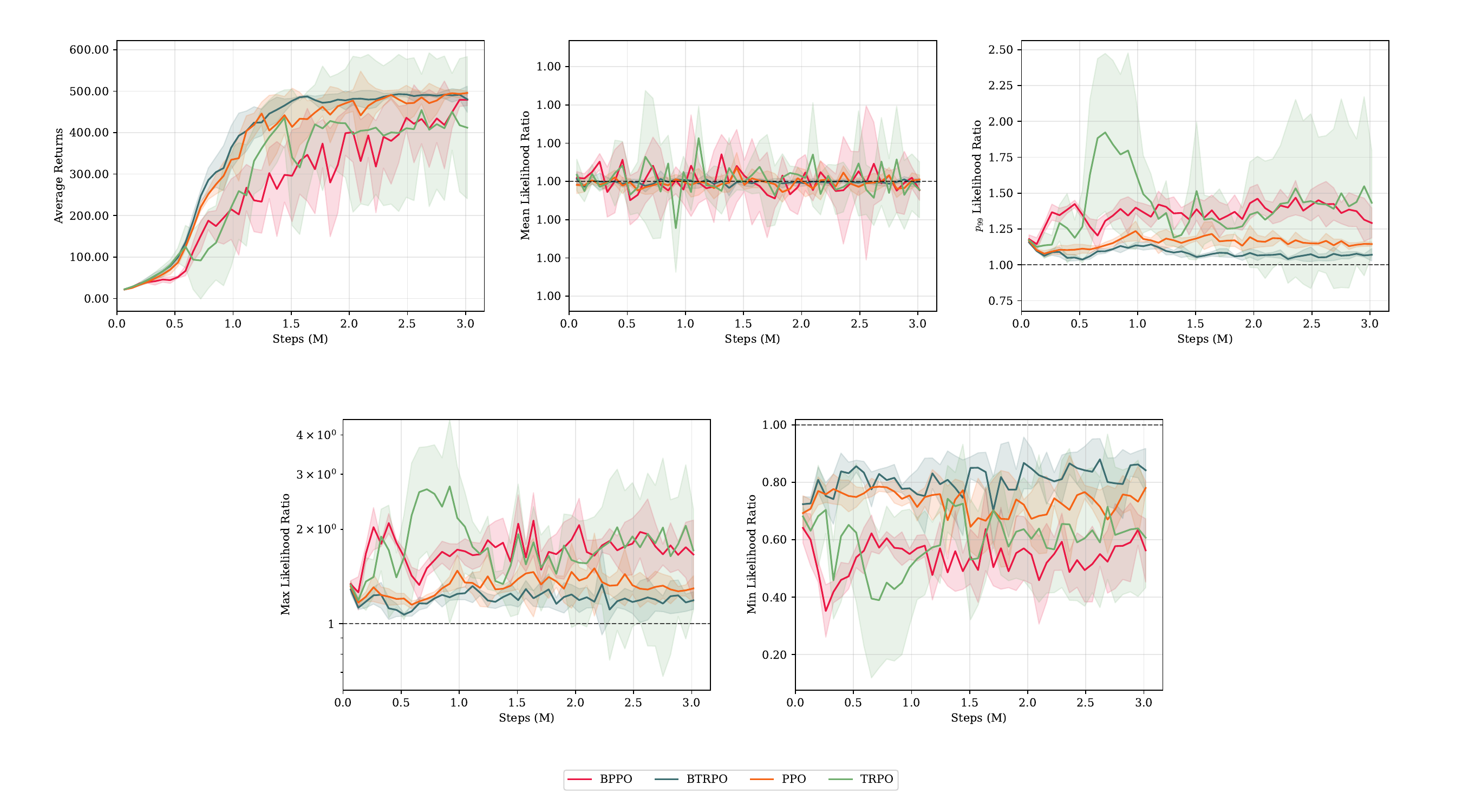}
    \caption{\textsc{CartPole-v1} baseline learning curves (4 seeds).}
\label{fig:cartpole_baseline}
\end{figure*}

\subsection{Baselines with entropy bonus}
\noindent\textbf{Entropy affects ordering.}
Figure~21 shows entropy can speed up early improvement for some methods, but it can also add variance.
Clipping-based methods retain stable ratios and remain near-ceiling; conservative penalties may reduce variance but are not essential on this task.

\begin{figure*}[ht]
    \centering
    \includegraphics[width=\textwidth]{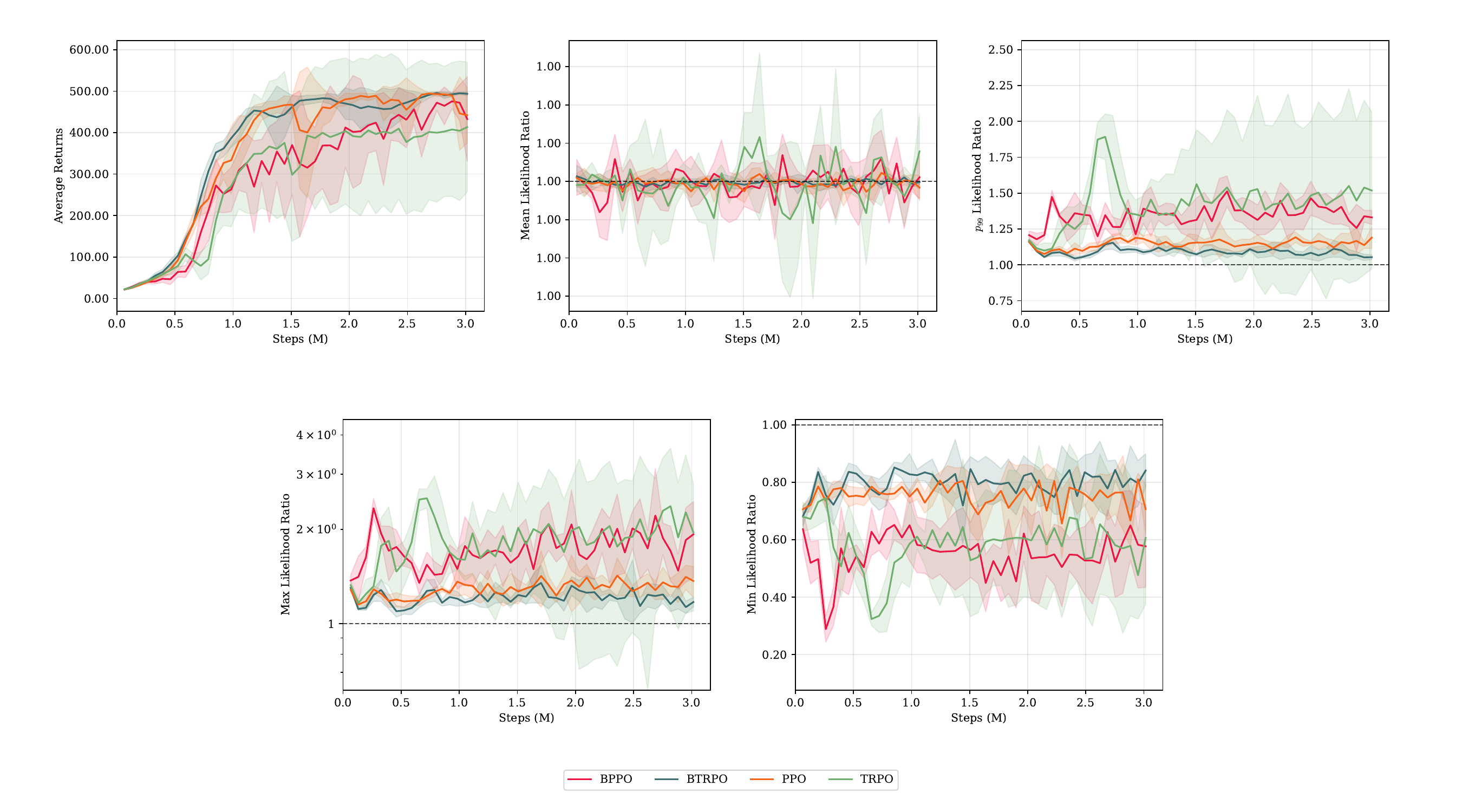}
    \caption{\textsc{CartPole-v1} baseline with entropy bonus learning curves (4 seeds).}
\label{fig:cartpole_baseline_ent}
\end{figure*}

\subsection{Regularized terms}
\noindent\textbf{Many choices work.}
Figure~22 shows that a wide range of regularizers still reach near-ceiling performance.
The main visible role of regularization is to compress ratio tails without materially improving returns.

\begin{figure*}[ht]
    \centering
    \includegraphics[width=\textwidth]{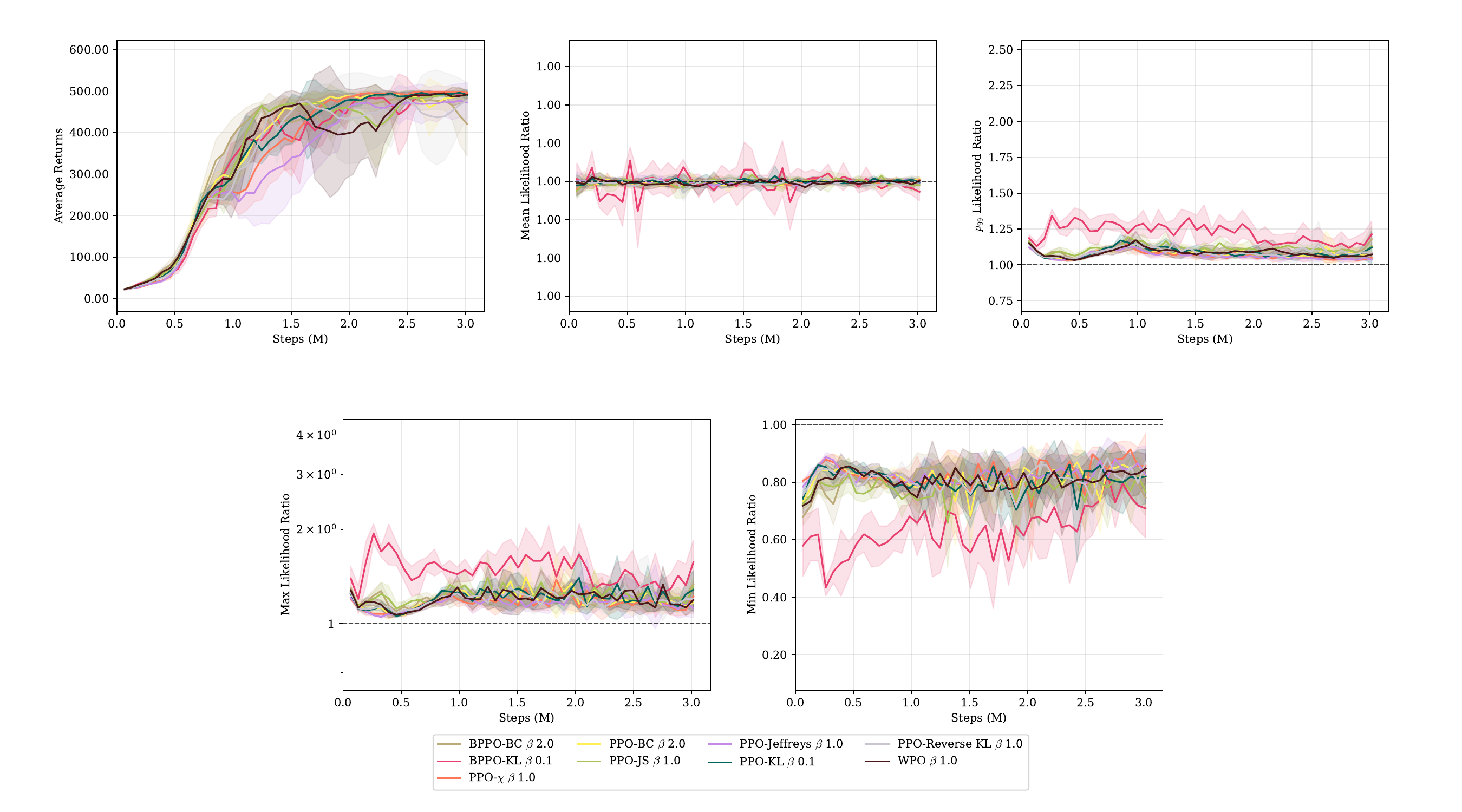}
    \caption{\textsc{CartPole-v1} regularized baseline learning curves (4 seeds).}
\label{fig:cartpole_regularized}
\end{figure*}

\subsection{Regularized terms with entropy bonus}
\noindent\textbf{Regularization + entropy.}
Figure~23 indicates regularizers reduce max-ratio spikes, but since performance is near-saturated, these changes primarily affect update statistics rather than final return.

\begin{figure*}[ht]  
    \centering
    \includegraphics[width=\textwidth]{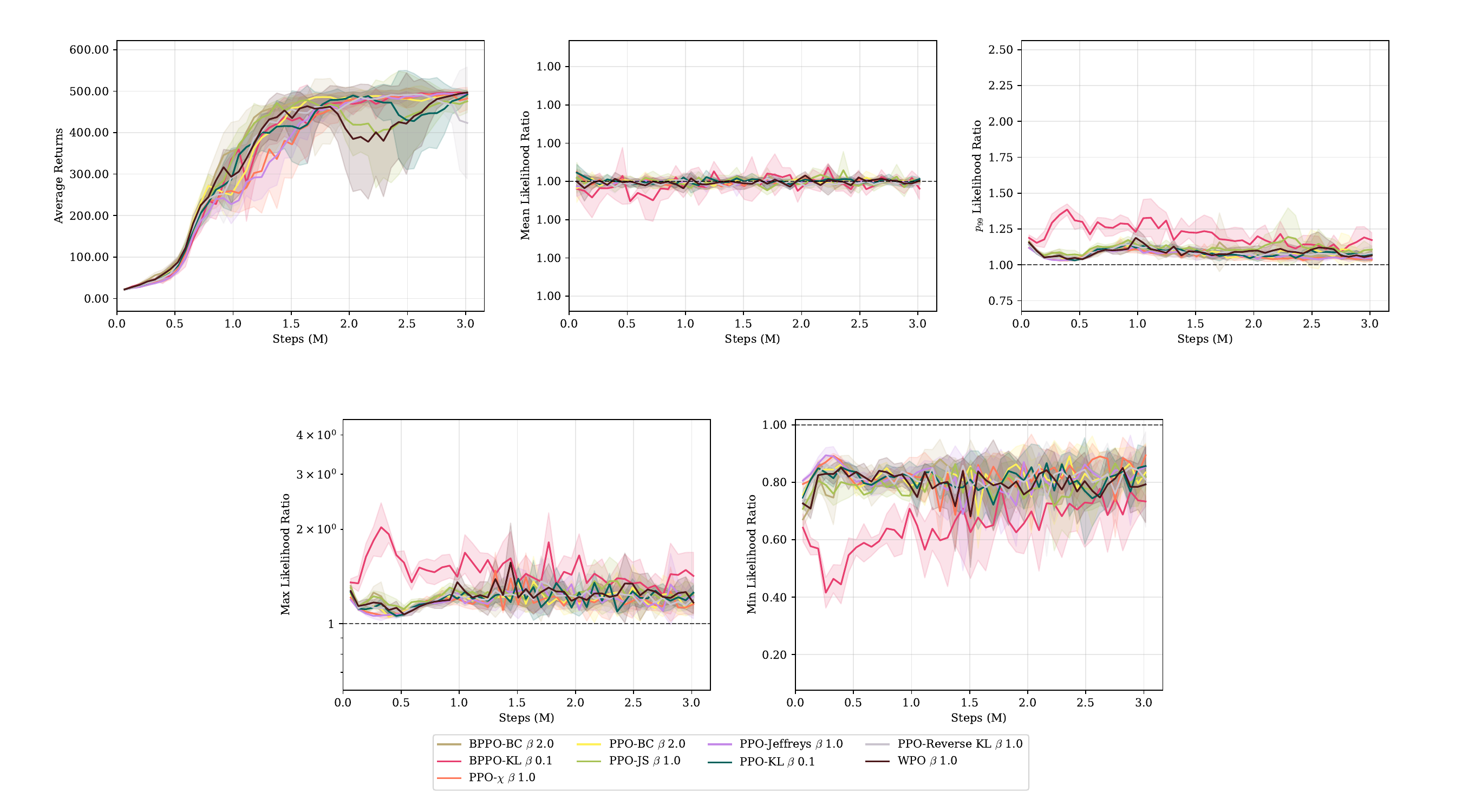}
    \caption{\textsc{CartPole-v1} regularized baseline with entropy bonus learning curves (4 seeds).}
\label{fig:cartpole_regularized_ent}
\end{figure*}

\begin{figure*}[ht]
\centering

\setlength{\fboxrule}{0.4pt}
\setlength{\fboxsep}{1.5pt}
\setlength{\tabcolsep}{2pt}  
\renewcommand{\arraystretch}{1} 

\newcommand{\frameimg}[2]{\fbox{\includegraphics[width=#1]{#2}}}

\begin{tabular}{ccccc}
\frameimg{0.17\textwidth}{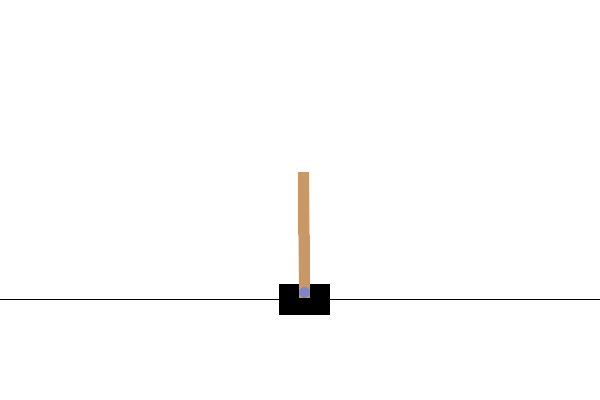} &
\frameimg{0.17\textwidth}{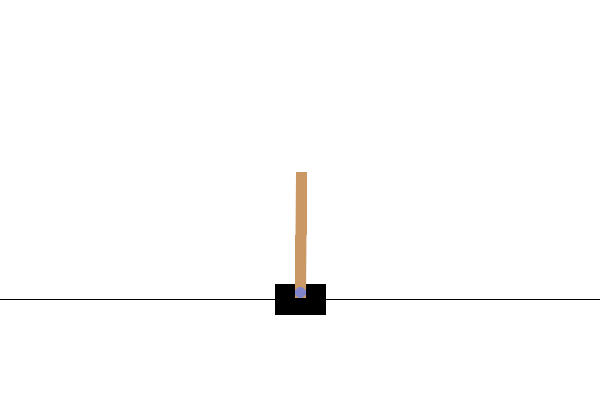} &
\frameimg{0.17\textwidth}{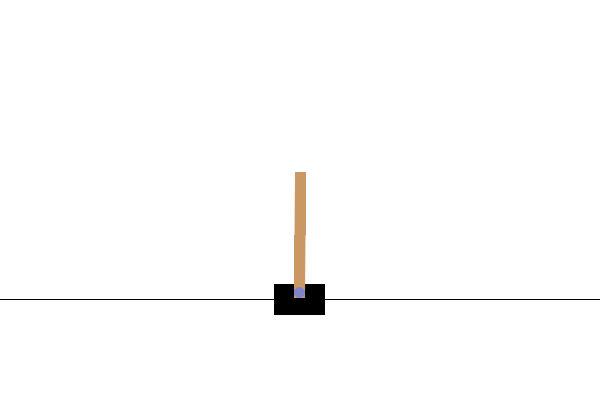} &
\frameimg{0.17\textwidth}{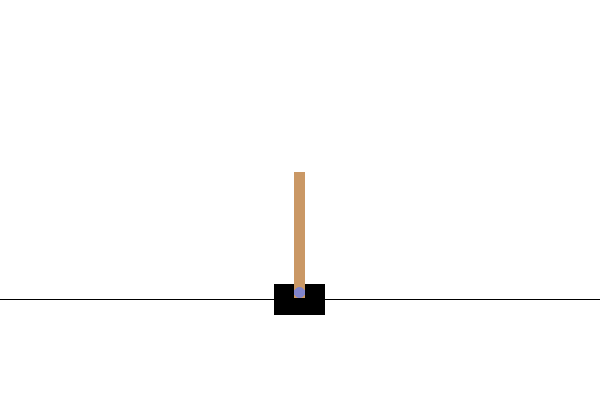} &
\frameimg{0.17\textwidth}{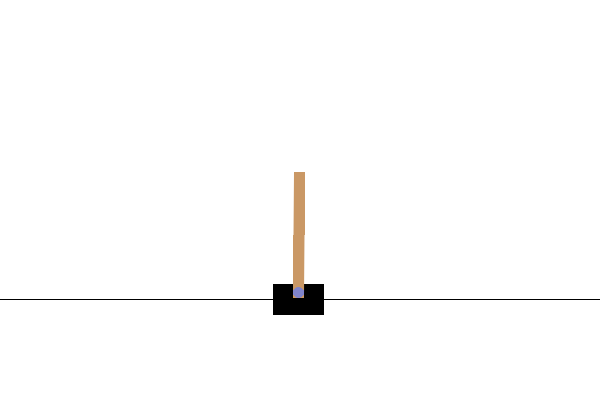} \\[4pt]
\end{tabular}
\caption{BPPO-BC with \(\text{entropy} = 0.01\) and \(\beta = 2.0\) \textsc{CartPole-v1} inference.}
\label{fig:cart-pole_inference}
\end{figure*}

\subsection{IQM Scores}
\noindent\textbf{Robust summary.}
Table \ref{tab:iqm_cartpole} confirms ceiling effects with tight IQM clustering.
The best IQM is achieved by BPPO-BC with entropy $0.01$ (IQM $492.09$), with several other methods close behind (e.g., PPO-$\chi$ at $491.66$, BTRPO at $490.21$).
Thus, Cartpole is best treated as a stability sanity check rather than a strong discriminator between trust-region geometries.

\begin{table}[t]
\centering
\caption{IQM scores across algorithms on \textsc{CartPole-v1} (4 seeds). Best result in bold.}
\label{tab:iqm_cartpole}
\begin{tabular}{lcccc} 
\toprule
Algorithm & Entropy & $\beta$ & IQM Score & Seeds \\
\midrule
BPPO & 0.0  & 0.0 & 383.85 & 4 \\
BPPO & 0.01 & 0.0 & 448.21 & 4 \\

BPPO-BC & 0.0  & 2.0 & 458.66 & 4 \\
BPPO-BC & 0.01 & 2.0 & \textbf{492.09} & 4 \\

BPPO-KL & 0.0  & 0.1 & 480.47 & 4 \\
BPPO-KL & 0.01 & 0.1 & 482.86 & 4 \\

BTRPO & 0.0  & 2.0 & 490.21 & 4 \\
BTRPO & 0.01 & 2.0 & 483.43 & 4 \\

PPO & 0.0  & 0.0 & 487.00 & 4 \\
PPO & 0.01 & 0.0 & 477.38 & 4 \\

PPO-$\chi$ & 0.0  & 1.0 & 491.66 & 4 \\
PPO-$\chi$ & 0.01 & 1.0 & 482.37 & 4 \\

PPO-BC & 0.0  & 2.0 & 489.57 & 4 \\
PPO-BC & 0.01 & 2.0 & 489.09 & 4 \\

PPO-JS & 0.0  & 1.0 & 465.42 & 4 \\
PPO-JS & 0.01 & 1.0 & 462.86 & 4 \\

PPO-Jeffreys & 0.0  & 1.0 & 462.69 & 4 \\
PPO-Jeffreys & 0.01 & 1.0 & 482.52 & 4 \\

PPO-KL & 0.0  & 1.0 & 485.52 & 4 \\
PPO-KL & 0.01 & 1.0 & 483.01 & 4 \\

PPO-Reverse KL & 0.0  & 1.0 & 460.91 & 4 \\
PPO-Reverse KL & 0.01 & 1.0 & 469.60 & 4 \\

TRPO & 0.0  & 0.1 & 488.66 & 4 \\
TRPO & 0.01 & 0.1 & 480.65 & 4 \\

WPO & 0.0  & 1.0 & 462.32 & 4 \\
WPO & 0.01 & 1.0 & 440.34 & 4 \\
\bottomrule
\end{tabular}
\end{table}

\newpage

\section{Procgen Training and Simulation}
\begin{figure*}[ht]
    \centering
    \includegraphics[width=\textwidth]{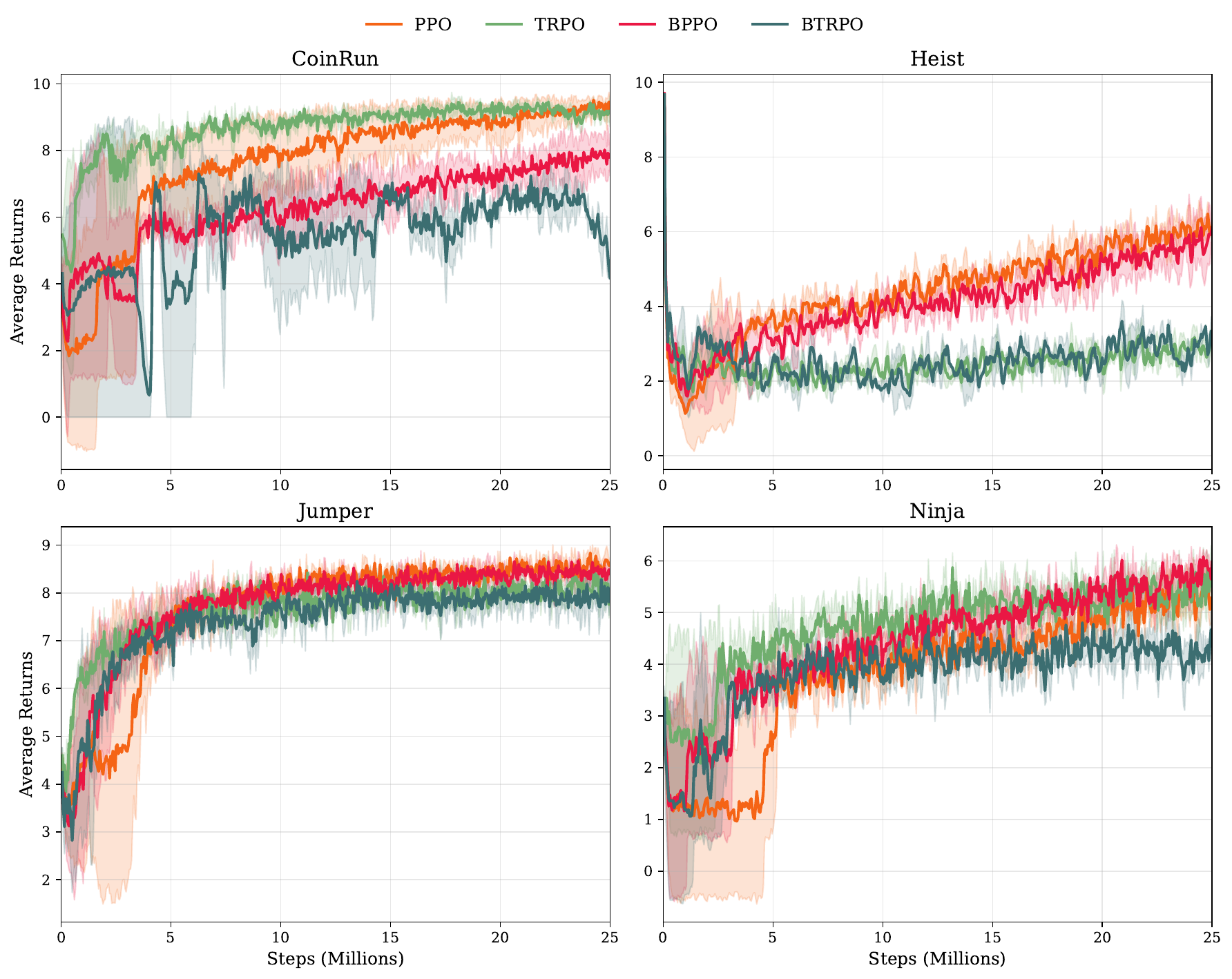}
    \caption{Procgen training on games over 3 seeds.}
\label{fig:procgen_training}
\end{figure*}

\noindent\textbf{Overview.}
Figure \ref{fig:procgen_training} evaluates Procgen training across four games under procedural stochasticity.
Unlike classical control, performance is more game-dependent due to diversity of levels, exploration demands, and sparse/structured rewards.
Consequently, we observe a mixed ranking across algorithms rather than a single consistent winner.

\paragraph{CoinRun.}
CoinRun shows strong performance for PPO/TRPO, with BPPO close behind.
This suggests that in relatively dense-reward settings where exploration is straightforward, overlap geometry remains competitive but does not necessarily dominate.

\paragraph{Heist.}
Heist appears more exploration-limited: methods that sustain effective updates and exploration tend to perform better.
Square-root clipping can help preserve a useful update signal under stochasticity while damping extreme ratio excursions, making BPPO competitive.

\paragraph{Jumper.}
Jumper shows smaller method separation, with several algorithms reaching similar plateaus.
This indicates the task is less sensitive to tail behavior and more influenced by general optimization variance.

\paragraph{Ninja.}
Ninja highlights a regime where continued improvement matters late in training.
BPPO can remain competitive due to reduced tail sensitivity and a smoother effective update signal, which helps avoid premature stagnation.

\newpage

\section{Likelihood Ratio on MuJoCo Envs}
\subsection{\(99^{\text{th}}\) Percentile Loglikelihood Ratio}
\noindent\textbf{Upper-tail summary.}
The $p_{99}$ ratio provides a robust view of the upper tail without being dominated by a single extreme sample.
Across MuJoCo tasks, TRPO typically shows larger $p_{99}$ ratios than clipping-based methods, indicating more frequent large-ratio events.
BPPO tends to keep $p_{99}$ closer to a moderate range while still allowing improvement, consistent with the motivation that overlap-based updates damp tail sensitivity.

\begin{figure*}[ht]
    \centering
    \includegraphics[width=\textwidth]{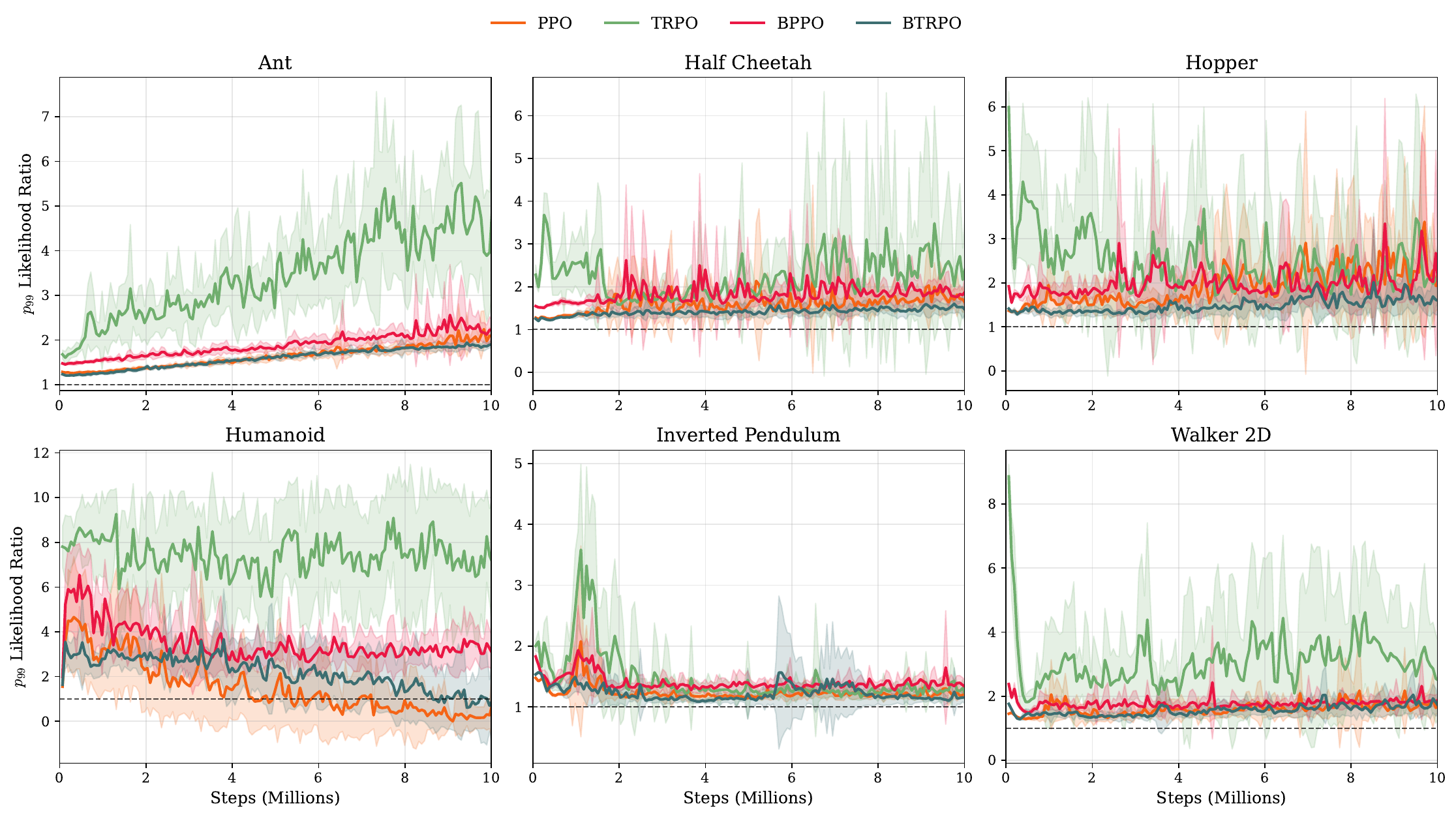}
    \caption{\(99^{\text{th}}\) percentile loglikelihood ratio on MuJoCo envs}
\label{fig:mujoco_99_training}
\end{figure*}

\subsection{Max Loglikelihood Ratio}
\noindent\textbf{Extreme-event behavior.}
The max ratio highlights rare but potentially destabilizing events.
TRPO often exhibits the largest max spikes (sometimes orders of magnitude), reflecting heavy-tail behavior.
PPO and BPPO substantially reduce these extremes; BPPO’s square-root geometry further reduces sensitivity to such spikes in the surrogate itself.

\begin{figure*}[ht]
    \centering
    \includegraphics[width=\textwidth]{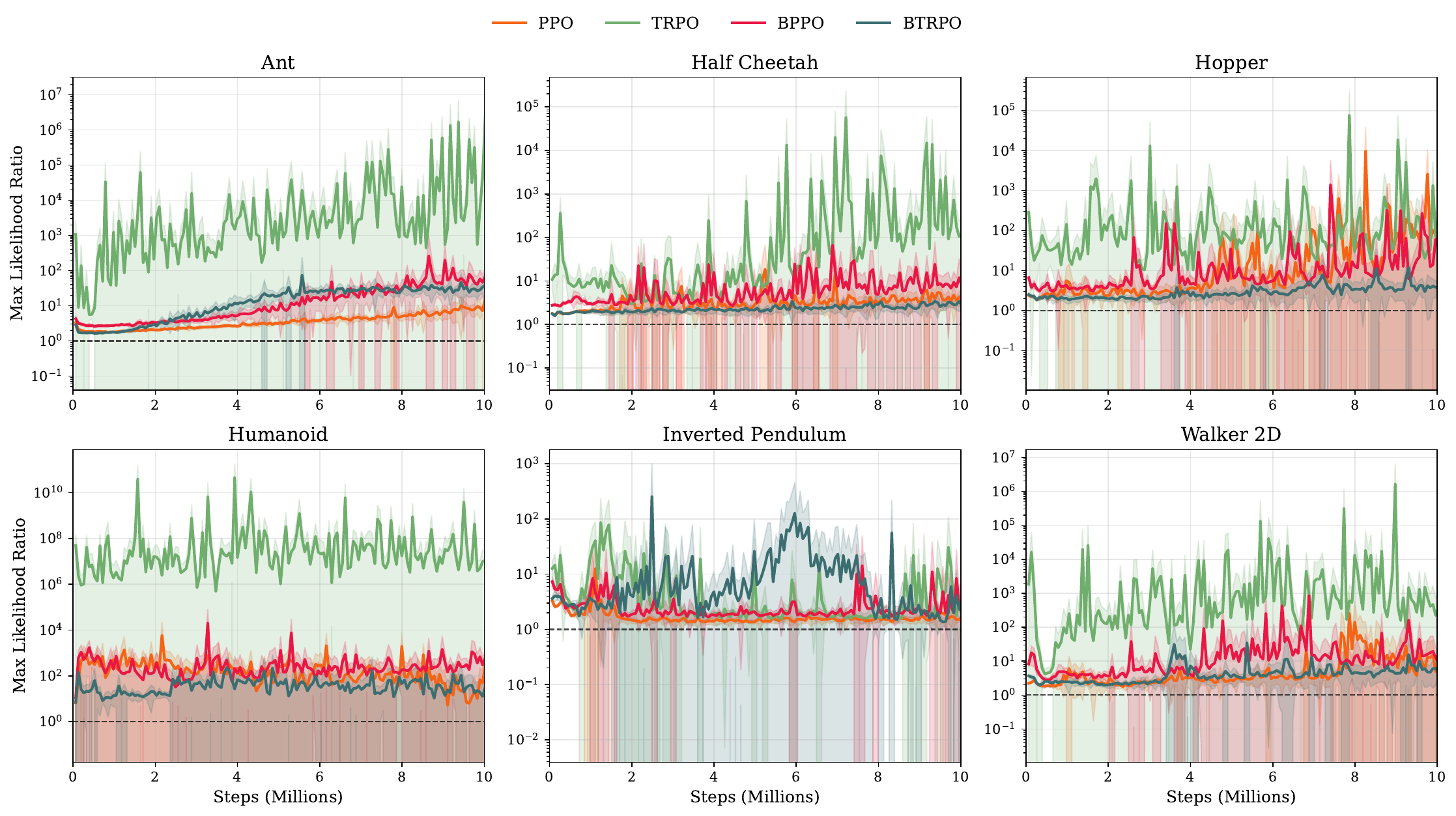}
    \caption{Max oglikelihood ratio on MuJoCo envs}
\label{fig:mujoco_max_training}
\end{figure*}

\subsection{Mean Loglikelihood Ratio}
\noindent\textbf{Drift and effective step size.}
The mean ratio should stay near 1 in expectation; sustained deviation indicates drift or biased updates.
Most methods remain close to 1 on easier tasks, but on hard locomotion domains the mean can drift and/or contract toward 1, consistent with increasingly conservative effective updates.
BPPO is designed to preserve near-unity mean ratio while keeping a non-trivial tail, enabling continued learning where updates otherwise “shrink.”

\begin{figure*}[ht]
    \centering
    \includegraphics[width=\textwidth]{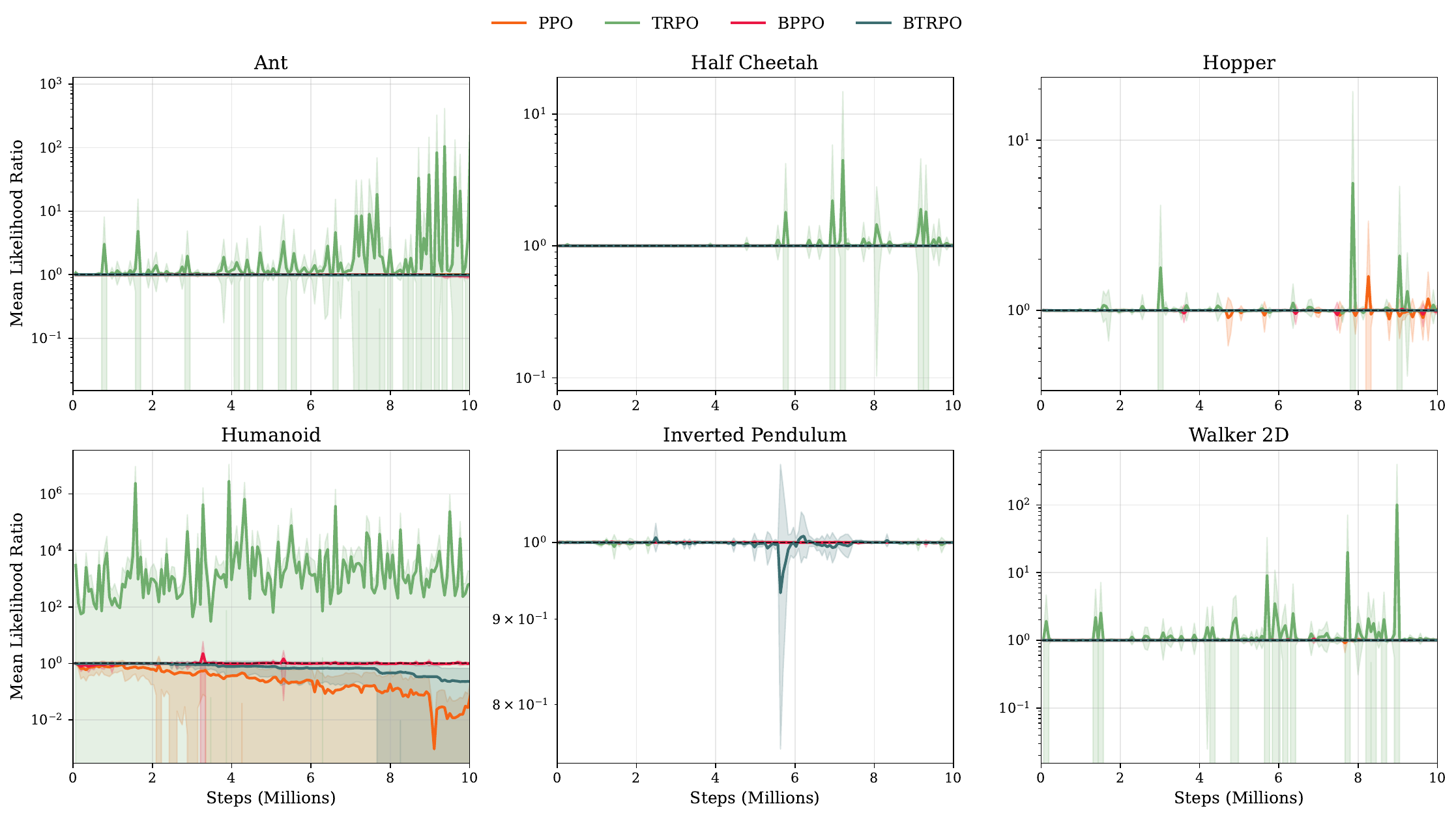}
    \caption{Mean oglikelihood ratio on MuJoCo envs}
\label{fig:mujoco_mean_training}
\end{figure*}

\subsection{Min Loglikelihood Ratio}
\noindent\textbf{Worst-case downweighting.}
The min ratio captures the strongest downweighting events (samples that become much less likely under the updated policy).
Very small minima indicate aggressive changes on some samples; clipping-based methods tend to avoid the most extreme collapses.
BPPO often sits between PPO and TRPO: more flexible than hard clipping on $r$, but less exposed to catastrophic extremes than raw ratio updates.

\begin{figure*}[ht]
    \centering
    \includegraphics[width=\textwidth]{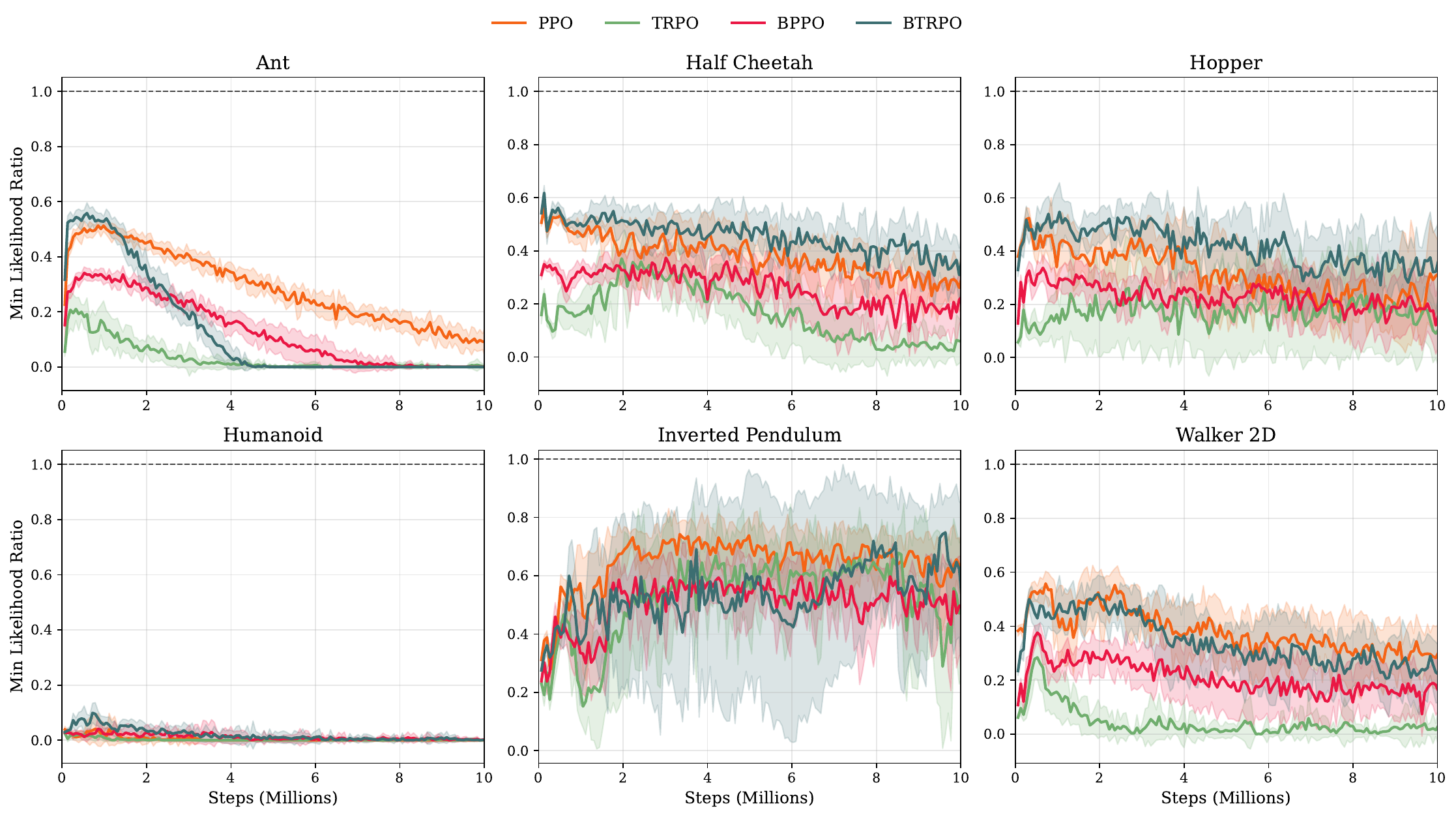}
    \caption{Min oglikelihood ratio on MuJoCo envs}
\label{fig:mujoco_min_training}
\end{figure*}

\newpage

\section{Likelihood Ratio on DM Envs}
\subsection{99th Percentile Loglikelihood Ratio}
\noindent\textbf{Upper-tail summary.}
On DM Control tasks, the $p_{99}$ ratio again separates KL-based TRPO from clipping/overlap-based methods.
TRPO generally shows larger $p_{99}$ ratios, while BPPO and PPO keep the upper tail closer to 1--2 in most tasks, consistent with improved robustness to rare large-ratio updates.

\begin{figure*}[ht]
    \centering
    \includegraphics[width=\textwidth]{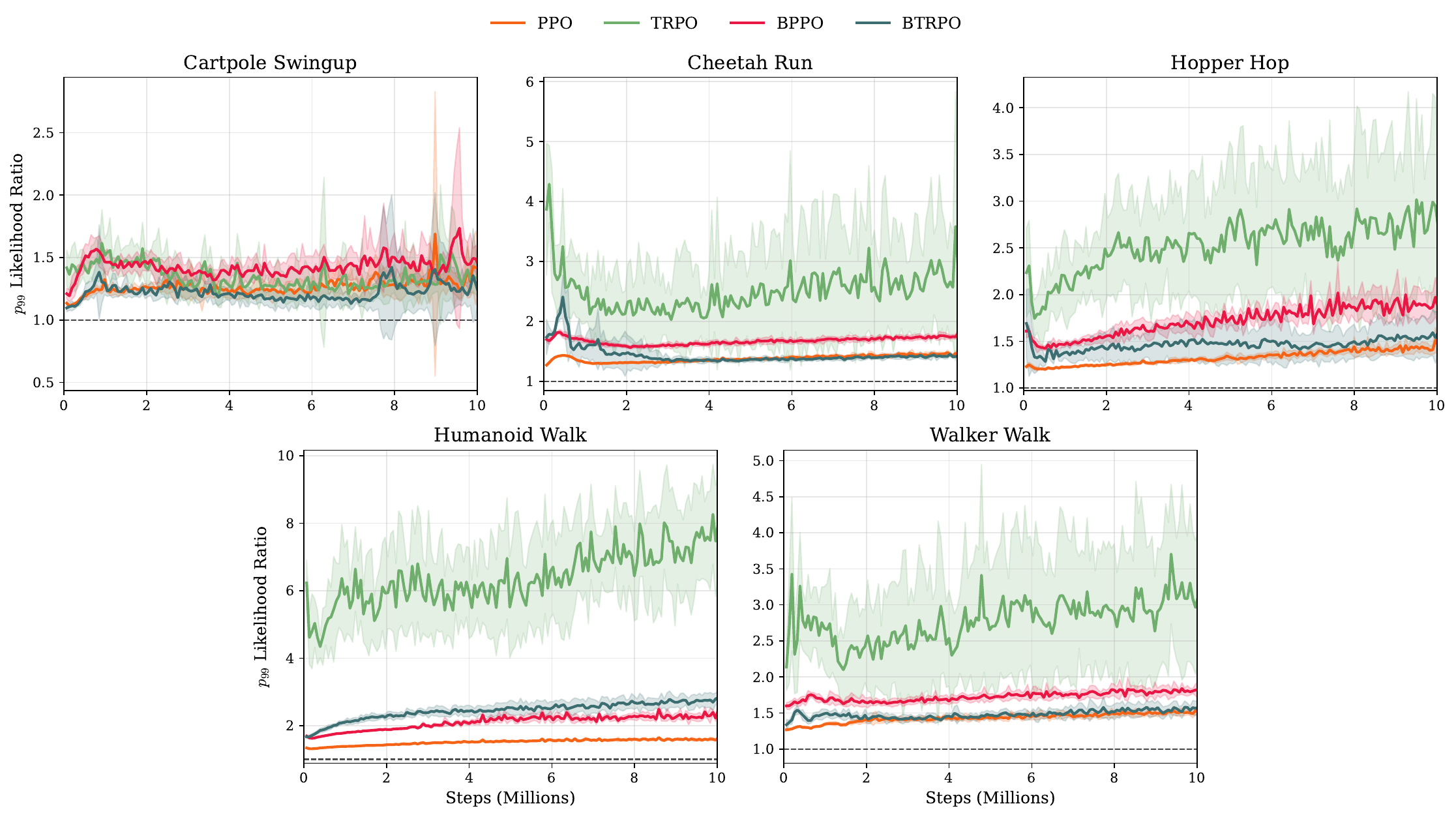}
    \caption{\(99^{\text{th}}\) percentile loglikelihood ratio on DM envs}
\label{fig:mujoco_99_training}
\end{figure*}

\subsection{Max Loglikelihood Ratio}
\noindent\textbf{Extreme spikes.}
Max ratios highlight the most extreme events, and TRPO often exhibits the largest spikes.
BPPO and PPO reduce these spikes substantially; BTRPO may sit between the two depending on the penalty strength and task difficulty.

\begin{figure*}[ht]
    \centering
    \includegraphics[width=\textwidth]{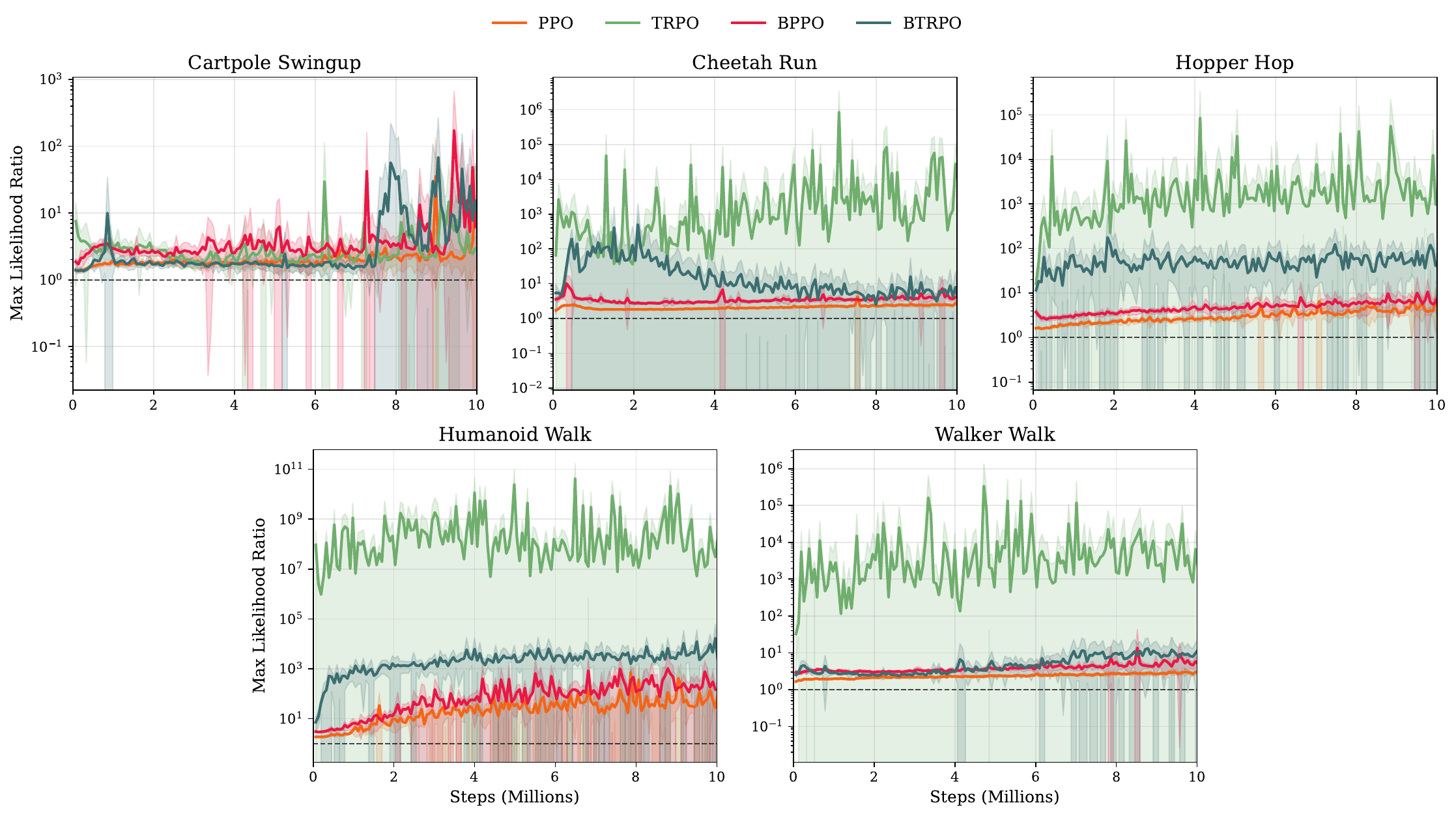}
    \caption{Max loglikelihood ratio on DM envs}
\label{fig:mujoco_max_training}
\end{figure*}

\subsection{Mean Loglikelihood Ratio}
\noindent\textbf{Step-size behavior.}
The mean ratio typically stays near 1, but harder domains can show drift or contraction.
A contraction toward 1 (or below) can correlate with stagnation, consistent with “update shrinkage” where the effective step becomes increasingly conservative.
BPPO aims to preserve near-unity mean while avoiding catastrophic tail behavior.

\begin{figure*}[ht]
    \centering
    \includegraphics[width=\textwidth]{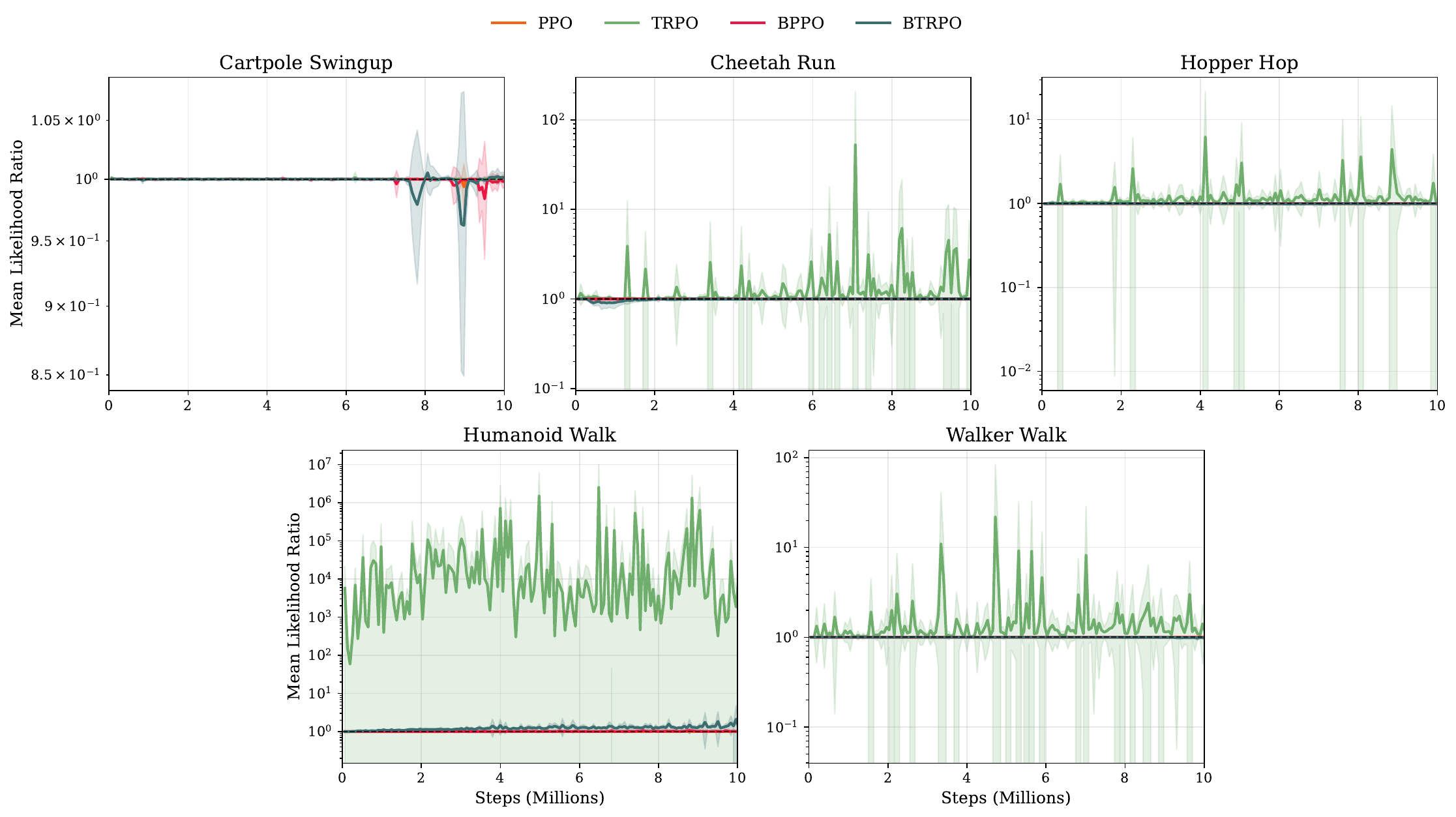}
    \caption{Mean loglikelihood ratio on DM envs}
\label{fig:mujoco_mean_training}
\end{figure*}

\subsection{Min Loglikelihood Ratio}
\noindent\textbf{Lower tail.}
The min ratio reflects the strongest downweighting events.
PPO often maintains larger minima due to clipping; TRPO tends to have smaller minima (more extreme changes on some samples).
BPPO typically provides an intermediate trade-off: reduced tail sensitivity while still allowing non-trivial policy movement.

\begin{figure*}[ht]
    \centering
    \includegraphics[width=\textwidth]{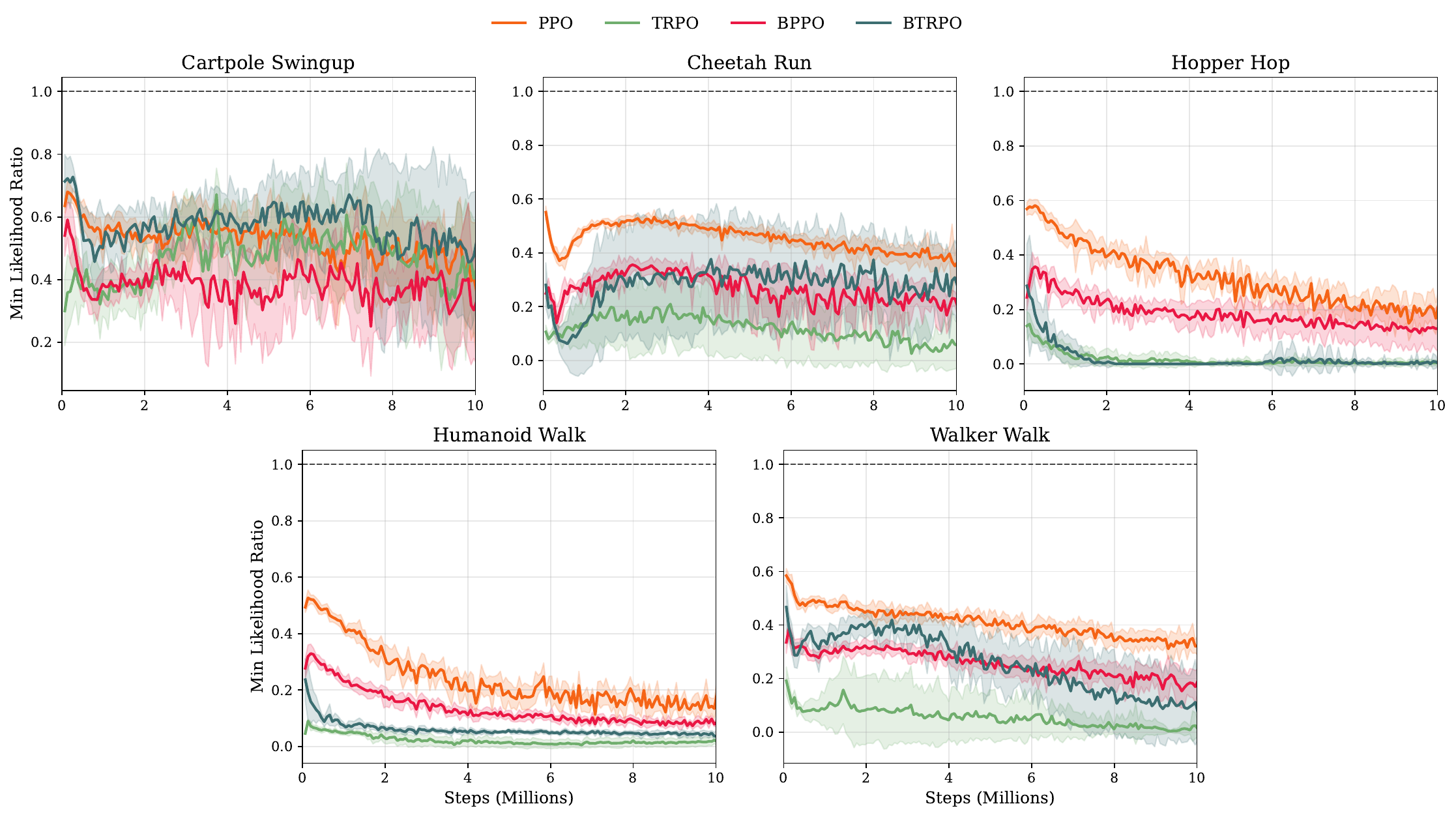}
    \caption{Min loglikelihood ratio on DM envs}
\label{fig:mujoco_min_training}
\end{figure*}

\newpage

\section{Ablation on Mountain Car}
\subsection{Ablation on clipping $\epsilon$, batch size, and learning rate (BPPO vs PPO)}
\noindent\textbf{Key pattern.}
Table \ref{tab:cartpole_ablation} shows that BPPO is more sensitive to the clipping range $\epsilon$ than PPO, and that the interaction with batch size and learning rate can flip outcomes.

\noindent\textbf{Moderate clipping can strongly favor BPPO.}
At $\epsilon=0.2$, BPPO achieves positive final return in regimes where PPO fails:
for example, at (BS $=8192$, LR $=3\cdot10^{-4}$) BPPO reaches $21.1$ while PPO is $-15.1$, and at (BS $=32768$, LR $=3\cdot10^{-4}$) BPPO reaches $91.9$ while PPO is $-22.0$.
This supports the intuition that square-root clipping can preserve an effective update signal where raw-ratio clipping becomes overly conservative.

\noindent\textbf{Overly large clipping destabilizes BPPO.}
At $\epsilon=0.5$, BPPO can collapse catastrophically in some regimes where PPO succeeds:
e.g., (BS $=8192$, LR $=3\cdot10^{-4}$) yields BPPO $-99.6$ while PPO reaches $92.6$.
Large $\epsilon$ effectively weakens the intended saturation, exposing BPPO to high-variance tail events under aggressive optimizer settings.

\begin{table}[t]
\centering
\caption{Ablation on clipping ($\epsilon$), batch size (BS), and learning rate (LR) on \textsc{CartPole} with 1M environment steps. We report final returns for BPPO and PPO (4 seeds).}
\label{tab:cartpole_ablation}
\begin{tabular}{cccccc}
\toprule
Clip & BS & LR & Final BPPO & Final PPO \\
\midrule
0.1 & 8192  & 1e-4 & -20.6 & -20.3 \\
0.1 & 8192  & 3e-4 & -15.7 & -15.0 \\
0.1 & 8192  & 1e-3 & -6.2  & -7.5  \\
0.1 & 16384 & 1e-4 & -23.1 & -22.7 \\
0.1 & 16384 & 3e-4 & -19.3 & -18.9 \\
0.1 & 16384 & 1e-3 & -11.8 & -12.4 \\
0.1 & 32768 & 1e-4 & -22.7 & -22.8 \\
0.1 & 32768 & 3e-4 & -22.0 & -21.6 \\
0.1 & 32768 & 1e-3 & -16.5 & -16.7 \\
\midrule
0.2 & 8192  & 1e-4 & -20.9 & -20.5 \\
0.2 & 8192  & 3e-4 & 21.1  & -15.1 \\
0.2 & 8192  & 1e-3 & -6.1  & -5.9  \\
0.2 & 16384 & 1e-4 & -22.6 & -23.0 \\
0.2 & 16384 & 3e-4 & -19.4 & -19.3 \\
0.2 & 16384 & 1e-3 & -12.2 & -11.8 \\
0.2 & 32768 & 1e-4 & -23.2 & -23.3 \\
0.2 & 32768 & 3e-4 & 91.9  & -22.0 \\
0.2 & 32768 & 1e-3 & -16.7 & -16.7 \\
\midrule
0.5 & 8192  & 1e-4 & -20.8 & -20.5 \\
0.5 & 8192  & 3e-4 & -99.6 & 92.6  \\
0.5 & 8192  & 1e-3 & -99.6 & 38.5  \\
0.5 & 16384 & 1e-4 & -22.9 & -21.5 \\
0.5 & 16384 & 3e-4 & -19.7 & -19.2 \\
0.5 & 16384 & 1e-3 & -99.6 & -11.3 \\
0.5 & 32768 & 1e-4 & -22.7 & -23.2 \\
0.5 & 32768 & 3e-4 & -22.0 & -22.5 \\
0.5 & 32768 & 1e-3 & -97.2 & -16.4 \\
\bottomrule
\end{tabular}
\end{table}

\subsection{Ablation on penalty weight $\beta$, batch size, and learning rate (TRPO vs BTRPO)}
\noindent\textbf{Key pattern (Table~10).}
Table~10 demonstrates that the Hellinger/BC penalty weight $\beta$ is a genuine trust-region knob: too small can destabilize, too large can over-constrain, and intermediate values can stabilize training depending on optimizer scale.

\noindent\textbf{$\beta=0$ can be unstable for BTRPO.}
When $\beta=0$, BTRPO can collapse even when TRPO succeeds:
e.g., (BS $=8192$, LR $=3\cdot10^{-4}$) TRPO reaches $91.3$ while BTRPO collapses to $-99.9$.
This indicates that the smooth overlap penalty is necessary to prevent destabilizing updates in some regimes.

\noindent\textbf{Intermediate $\beta$ can flip the outcome.}
At $\beta=0.1$, BTRPO can succeed where TRPO fails:
e.g., (BS $=8192$, LR $=10^{-4}$) TRPO is $-20.8$ while BTRPO reaches $92.4$, and at (BS $=16384$, LR $=3\cdot10^{-4}$) both are strong with BTRPO slightly higher ($93.2$ vs $92.5$).
At higher learning rates, the optimal $\beta$ can change sharply, reinforcing that fixed penalties trade off stability vs progress and motivating adaptive/dual tuning as future work.

\begin{table}[t]
\centering
\caption{Ablation on penalty weight $\beta$, batch size (BS), and learning rate (LR) for \textsc{TRPO} and \textsc{BTRPO} on \textsc{CartPole} with 1M environment steps (4 seeds).}
\label{tab:cartpole_trpo_btrpo_ablation}
\begin{tabular}{ccccc}
\toprule
$\beta$ & BS & LR & Final TRPO & Final BTRPO \\
\midrule
0.0 & 8192  & 1e-4 & -20.70 & -20.80 \\
0.0 & 8192  & 3e-4 & 91.30  & -99.90 \\
0.0 & 8192  & 1e-3 & -6.50  & -99.90 \\
0.0 & 16384 & 1e-4 & -22.50 & -22.90 \\
0.0 & 16384 & 3e-4 & 93.90  & -19.30 \\
0.0 & 16384 & 1e-3 & 93.50  & -99.90 \\
0.0 & 32768 & 1e-4 & -22.70 & -23.20 \\
0.0 & 32768 & 3e-4 & -34.20 & -22.40 \\
0.0 & 32768 & 1e-3 & -16.50 & -17.80 \\
\midrule
0.1 & 8192  & 1e-4 & -20.80 & 92.40 \\
0.1 & 8192  & 3e-4 & 93.20  & -15.80 \\
0.1 & 8192  & 1e-3 & 94.50  & -73.90 \\
0.1 & 16384 & 1e-4 & -22.40 & -23.00 \\
0.1 & 16384 & 3e-4 & 92.50  & 93.20 \\
0.1 & 16384 & 1e-3 & -12.10 & -12.00 \\
0.1 & 32768 & 1e-4 & -22.20 & -23.20 \\
0.1 & 32768 & 3e-4 & -21.60 & -21.80 \\
0.1 & 32768 & 1e-3 & -16.80 & 93.00 \\
\midrule
1.0 & 8192  & 1e-4 & -20.60 & -20.60 \\
1.0 & 8192  & 3e-4 & -15.60 & -15.80 \\
1.0 & 8192  & 1e-3 & -6.70  & 93.90 \\
1.0 & 16384 & 1e-4 & -22.40 & -21.90 \\
1.0 & 16384 & 3e-4 & -19.00 & -19.30 \\
1.0 & 16384 & 1e-3 & -12.40 & -12.20 \\
1.0 & 32768 & 1e-4 & -22.70 & -22.60 \\
1.0 & 32768 & 3e-4 & -21.80 & -21.80 \\
1.0 & 32768 & 1e-3 & -16.50 & -16.50 \\
\midrule
2.0 & 8192  & 1e-4 & -20.30 & -20.40 \\
2.0 & 8192  & 3e-4 & -15.60 & -15.90 \\
2.0 & 8192  & 1e-3 & 29.50  & -6.40 \\
2.0 & 16384 & 1e-4 & -22.40 & -22.70 \\
2.0 & 16384 & 3e-4 & -19.20 & -19.30 \\
2.0 & 16384 & 1e-3 & -13.00 & -12.40 \\
2.0 & 32768 & 1e-4 & -22.60 & -22.80 \\
2.0 & 32768 & 3e-4 & -21.60 & -20.90 \\
2.0 & 32768 & 1e-3 & -16.90 & -16.50 \\
\midrule
5.0 & 8192  & 1e-4 & -20.20 & -20.20 \\
5.0 & 8192  & 3e-4 & -16.80 & -16.10 \\
5.0 & 8192  & 1e-3 & -15.00 & -9.30  \\
5.0 & 16384 & 1e-4 & -21.70 & -22.30 \\
5.0 & 16384 & 3e-4 & -19.20 & -19.00 \\
5.0 & 16384 & 1e-3 & -15.20 & -12.10 \\
5.0 & 32768 & 1e-4 & -22.20 & -22.80 \\
5.0 & 32768 & 3e-4 & -21.40 & -21.40 \\
5.0 & 32768 & 1e-3 & -17.10 & -16.80 \\
\bottomrule
\end{tabular}
\end{table}

\newpage

\part*{Frequently Asked Questions (FAQs)}
\textbf{Q1: Why use the Hellinger distance (or Bhattacharyya coefficient) instead of the standard KL divergence?}

\textbf{A:} The KL divergence ($\mathbb{E}[\log r]$) measures average discrepancy and is famously insensitive to "tail excursions"—rare events where the probability ratio $r$ is very large. In RL, these excursions correspond to high-variance updates that can destabilize training. 
In contrast, the Hellinger distance operates on the square-root manifold ($q = \sqrt{r}$). Geometrically, this metric inherently penalizes large ratios more severely in the tails (dominance) while remaining locally equivalent to the Fisher Information metric near $r=1$. This allows BTRPO to control risk without needing the heuristic clipping used in PPO.

\vspace{1em}

\textbf{Q2: Is BPPO simply PPO with a different clipping function?}

\textbf{A:} No, the mechanism is fundamentally different. 
PPO employs \textit{censorship}: it hard-clips the probability ratio $r$ when it exceeds a threshold, effectively zeroing out the gradient signal for high-advantage updates. 
BPPO employs \textit{geometric dampening}. By optimizing the square-root ratio $q = \sqrt{r}$, the effective gradient signal scales as $O(\sqrt{r})$ rather than $O(r)$. This compresses heavy tails without removing them, allowing the agent to learn from "eureka moments" (high-advantage outliers) without suffering from variance explosion. 

\vspace{1em}

\textbf{Q3: Does BTRPO require second-order optimization like TRPO?}

\textbf{A:} No. BTRPO is a first-order method. 
Standard TRPO requires computing the Fisher-vector product (often via Conjugate Gradient) to enforce the KL constraint. BTRPO enforces the Hellinger constraint via a scalar penalty term $\beta(1-q_\theta)^2$. This penalty is computationally negligible (element-wise operations) and allows BTRPO to match the wall-clock speed of PPO while retaining the theoretical stability benefits of a Trust Region method.

\vspace{1em}

\textbf{Q4: Why does BPPO perform significantly better on high-dimensional tasks like \texttt{humanoid\_walk}?}

\textbf{A:} In high-dimensional spaces, the joint probability ratio $r = \prod \pi(a_i|s)$ can easily become large even for small per-dimension changes, triggering PPO's clipping mechanism frequently. This "censors" coherent, multi-joint policy updates. 
BPPO's square-root geometry naturally dampens this product structure ($ \sqrt{\prod r_i} = \prod \sqrt{r_i} $), preserving the directional information of the update while preventing magnitude explosion. This allows for more aggressive, yet stable, policy improvement in complex coordination tasks.

\vspace{1em}

\textbf{Q5: How sensitive is the method to the regularization coefficient $\beta$?}

\textbf{A:} We find BTRPO to be robust to the choice of $\beta$. Because the Hellinger penalty and the performance objective are geometrically consistent (both operating on the square-root manifold), they scale similarly across different environments. We found a standard range of $\beta \in [0.01, 0.1]$ to work well across all DM Control tasks, with $\beta=0.05$ serving as a robust default. This contrasts with KL penalties, which often require adaptive tuning (e.g., Adaptive KL in PPO) to prevent the penalty from vanishing or exploding.

\vspace{1em}

\textbf{Q6: How should we set the BPPO clipping parameter $\epsilon$? How does it compare to PPO's clipping?}

\textbf{A:} BPPO clips the square-root likelihood ratio
$q \triangleq \sqrt{r} = \sqrt{\pi_\theta(a\mid s)/\pi_{\mathrm{old}}(a\mid s)}$
rather than $r$ itself. Concretely, BPPO uses the same ``min/clip'' template but with
$q$ and $\mathrm{clip}(q,1-\epsilon,1+\epsilon)$.
This implies an \emph{effective} constraint on the usual ratio,
\[
q \in [1-\epsilon,\,1+\epsilon]
\quad\Longleftrightarrow\quad
r \in \big[(1-\epsilon)^2,\,(1+\epsilon)^2\big].
\]
Thus, $\epsilon$ is directly interpretable as a trust-region radius \emph{in overlap geometry}.
In practice we found $\epsilon$ in the same ballpark as PPO (e.g., $0.1$--$0.2$) to be a robust
default, with the key difference that clipping is applied in $q$-space (geometric dampening)
rather than $r$-space (censoring).

\vspace{1em}

\textbf{Q7: Does computing the Bhattacharyya / Hellinger terms require extra rollouts or expensive second-order machinery?}

\textbf{A:} No. All quantities are estimated on the same on-policy batch collected under
$\pi_{\mathrm{old}}$. In particular, $q=\sqrt{\pi_\theta/\pi_{\mathrm{old}}}$ is computed per-sample,
and the BTRPO penalty uses simple minibatch averages of $(1-q)^2$.
There are no Fisher-vector products, conjugate-gradient solves, or additional environment
interaction beyond standard PPO/TRPO-style data collection.

\vspace{1em}

\textbf{Q8: How does BPPO behave when advantages are negative? Does it retain PPO's sign-dependent clipping behavior?}

\textbf{A:} Yes. BPPO preserves the same sign-dependent structure as PPO because it uses the same
$\min(\cdot,\cdot)$ surrogate template, but applied to $q=\sqrt{r}$.
For $A_{\mathrm{old}}(s,a)>0$, the objective prevents overly large update weights by clipping $q$ above
$1+\epsilon$; for $A_{\mathrm{old}}(s,a)<0$, it prevents overly small weights by clipping $q$ below
$1-\epsilon$. The key difference from PPO is that extreme likelihood-ratio excursions are
\emph{geometrically damped} via $q=e^{\Delta/2}$ rather than weighted directly by $r=e^{\Delta}$,
which reduces sensitivity to rare tail events without changing the core monotonicity logic of
the clipped surrogate.

\vspace{1em}

\textbf{Q9: Where should we expect BPPO/BTRPO to help most, and where might gains be limited?}

\textbf{A:} Overlap geometry is designed to control \emph{tail excursions} of the likelihood ratio,
so gains are most pronounced in regimes where training instability is driven by rare but large
ratio spikes (e.g., high-dimensional continuous control with expressive policies and imperfect
advantage estimates). In settings where performance is dominated by other factors---for example,
very high intrinsic stochasticity, heavy exploration noise, or sparse-reward discovery---the
ratio tails may not be the primary bottleneck, and improvements can be smaller or mixed.
Accordingly, we view BPPO as targeting stability failures arising from ratio tails, rather than
claiming uniform dominance across all RL regimes.

\vspace{1em}

\textbf{Q10: Is BPPO merely a reparameterization of PPO, or does it represent a genuinely different trust region?}

\textbf{A:} BPPO is not a cosmetic reparameterization: it optimizes in a different geometric
coordinate, $q=\sqrt{r}$, which corresponds to the square-root density (overlap) manifold induced
by the Bhattacharyya/Hellinger geometry. This changes the update weighting from $r=e^{\Delta}$
to $q=e^{\Delta/2}$, thereby compressing the influence of tail events in the likelihood-ratio
distribution. Moreover, BTRPO's penalty is directly tied to overlap via
\[
\mathbb{E}_{\mathrm{old}}\big[(1-q)^2\big]
\;\propto\;
1 - \mathrm{BC}(\pi_{\mathrm{old}},\pi_\theta),
\]
so the regularizer is geometrically consistent with the underlying overlap notion of ``distance.''
Taken together, the method implements a different trust-region control objective: it constrains
separation in the \emph{overlap} geometry rather than in KL (an average log-separation) and thus
targets the specific failure mode of rare, large likelihood-ratio excursions.

\end{document}